\pdfoutput=1
\PassOptionsToPackage{numbers, compress}{natbib}
\documentclass[11pt]{article}

\usepackage[preprint]{acl}

\usepackage{times}
\usepackage{latexsym}

\usepackage[T1]{fontenc}

\usepackage[utf8]{inputenc}

\usepackage{microtype}

\usepackage{inconsolata}

\usepackage{graphicx}

\usepackage[utf8]{inputenc} 
\usepackage[T1]{fontenc}    
\usepackage{hyperref}       
\usepackage{url}            
\usepackage{booktabs}       
\usepackage{amsfonts}       
\usepackage{nicefrac}       
\usepackage{microtype}      
\usepackage{xcolor}         

\usepackage{graphicx}
\usepackage{multirow}
\usepackage{tabularx}
\usepackage{colortbl}
\usepackage{array} 
\usepackage{graphicx}
\usepackage{subcaption}

\hypersetup{
       colorlinks=true,
   linkcolor=black,
       citecolor=black,
       filecolor=black,
       menucolor=black,
       urlcolor=black,
       breaklinks=true
}

\usepackage{colortbl}
\definecolor{lightgray}{gray}{0.9}
\usepackage{booktabs}
\usepackage{color, soul}
\usepackage{listings}

\definecolor{vscodeBlue}{rgb}{0.2,0.6,1.0}
\definecolor{vscodeGreen}{rgb}{0.0,0.5,0.0}
\definecolor{vscodeGray}{rgb}{0.5,0.5,0.5}
\definecolor{vscodePurple}{rgb}{0.58,0.0,0.82}
\definecolor{vscodeBackground}{rgb}{0.95,0.95,0.95}
\definecolor{lightgreen}{rgb}{0.56, 0.93, 0.56}
\definecolor{lightorange}{rgb}{0.98, 0.81, 0.69}

\lstdefinestyle{mystyle}{
    backgroundcolor=\color{vscodeBackground},   
    commentstyle=\color{vscodeGreen},
    keywordstyle=\color{vscodeBlue},
    numberstyle=\tiny\color{vscodeGray},
    stringstyle=\color{vscodePurple},
    basicstyle=\ttfamily\scriptsize,
    breakatwhitespace=false,         
    breaklines=true,                 
    captionpos=b,                    
    keepspaces=true,                 
    numbers=none,                    
    numbersep=5pt,                  
    showspaces=false,                
    showstringspaces=false,
    showtabs=false,                  
    tabsize=2
}
\newcommand{\ignorebig}[1]{}

\newlength\savewidth

\newcommand{\model}{Multi-Stage-Multi-Try}
\newcommand{\smodel}{MSMT}
\newcommand{\ourbench}{SearchBench}

\definecolor{citecolor}{RGB}{34,139,34}
\definecolor{lightred}{RGB}{241,140,142}
\definecolor{amber(sae/ece)}{rgb}{1.0, 0.49, 0.0}
\definecolor{battleshipgrey}{rgb}{0.52, 0.52, 0.51}
\definecolor{cadmiumorange}{rgb}{0.93, 0.53, 0.18}
\definecolor{applegreen}{rgb}{0.55, 0.71, 0.0}
\definecolor{cadmiumgreen}{rgb}{0.0, 0.42, 0.24}
\definecolor{forestgreen}{rgb}{0.13, 0.55, 0.13}
\definecolor{red}{rgb}{0.89, 0.0, 0.13}

\usepackage{array}
\usepackage{tabularx}
\usepackage{geometry}
\usepackage{url}
\usepackage{booktabs}
\usepackage{makecell}
\usepackage{placeins}
\usepackage{float}

\hypersetup{ 
       colorlinks=true,
   linkcolor=darkblue, 
       citecolor=black,
       filecolor=black,
       menucolor=black,
       urlcolor=black,
       breaklinks=true
}

\usepackage[toc,page,header]{appendix}
\usepackage{minitoc}

\usepackage{hyperref}
\usepackage{balance}

\newcommand{\minisection}[1]{\subsection*{#1}\addcontentsline{toc}{subsection}{#1}}

\title{Navigating the Labyrinth: \\Evaluating LLMs’ Ability to Reason About Search Problems}

\author{%
  Nasim Borazjanizadeh\\
  {\small Berkeley AI Research, UC Berkeley}\\
  \And
  Roei Herzig \\
  {\small Berkeley AI Research, UC Berkeley}\\
  \And
  Trevor Darrell \\
  {\small Berkeley AI Research, UC Berkeley}\\
  \AND
  Rogerio Feris \\
  {\small MIT-IBM Watson AI Lab}\\
  \And
  Leonid Karlinsky \\
  {\small MIT-IBM Watson AI Lab}\\
}

\usepackage[moderate]{savetrees}

\begin{document}

\maketitle

\doparttoc 
\faketableofcontents 

\part{} 

\vspace{-1.1cm}
\begin{abstract}
Large Language Models (LLMs) have recently achieved impressive performance in math and reasoning benchmarks. However, they often struggle with logic problems and puzzles that are relatively easy for humans. To further investigate this, we introduce a new benchmark, \ourbench{}, which contains 11 unique search problems inspired by intuitive puzzles. Each \ourbench{} problem type is equipped with automated pipelines to generate an arbitrary number of instances and analyze the feasibility, correctness, and optimality of LLM-generated solutions. We show that using step-by-step, language-only reasoning, even the most advanced LLMs fail to solve \ourbench{}; for example, OpenAI's frontier models GPT-4 and o1-preview solve only 1.4\% and 18.6\% of problems, respectively. The reason is that SearchBench problems require considering multiple pathways and performing backtracking, posing a significant challenge to auto-regressive models. Interestingly, performance is significantly boosted when we prompt models to generate a complete A* search algorithm—a comparatively more cognitively difficult task. This approach effectively offloads the iterative search and backtracking process from the models, which they struggles with in text. This in-context learning baseline is further enhanced via a Multi-Stage-Multi-Try (MSMT) inference method, increasing GPT-4's rate of correct solutions to over 57\%.

\end{abstract}

\section{Introduction}
\vspace{-0.2cm}

The advent of Large Language Models (LLMs) has revolutionized the field of natural language processing, with models such as Llama3.1 ~\citep{llama3.1}, GPT-4~\citep{OpenAI2023GPT4TR}, and o1-preview~\citep{o1-preview} demonstrating unprecedented performance on math and science QA benchmarks, such as GSM8k~\citep{cobbe2021training} and GPQA~\citep{rein2023gpqa}. However, LLMs still exhibit surprising failures on some intuitive tasks~\citep{bian2023chatgpt, qin2023chatgpt, marcus2020next} and struggle with multi-step compositional reasoning, combinatorial problems, and planning~\citep{dziri2024faith, valmeekam2022large, wu2023reasoning}. Inspired by these observations and to further investigate LLMs' reasoning abilities, we offer a new benchmark of combinatorial search problems, \ourbench{}. The problems of \ourbench{} are inspired by popular puzzles and predominantly NP-hard combinatorial problems and necessitate exploring multiple action paths and backtracking to previous states.

\begin{figure*}[t]
    \includegraphics[width=16.4cm, height=9cm]{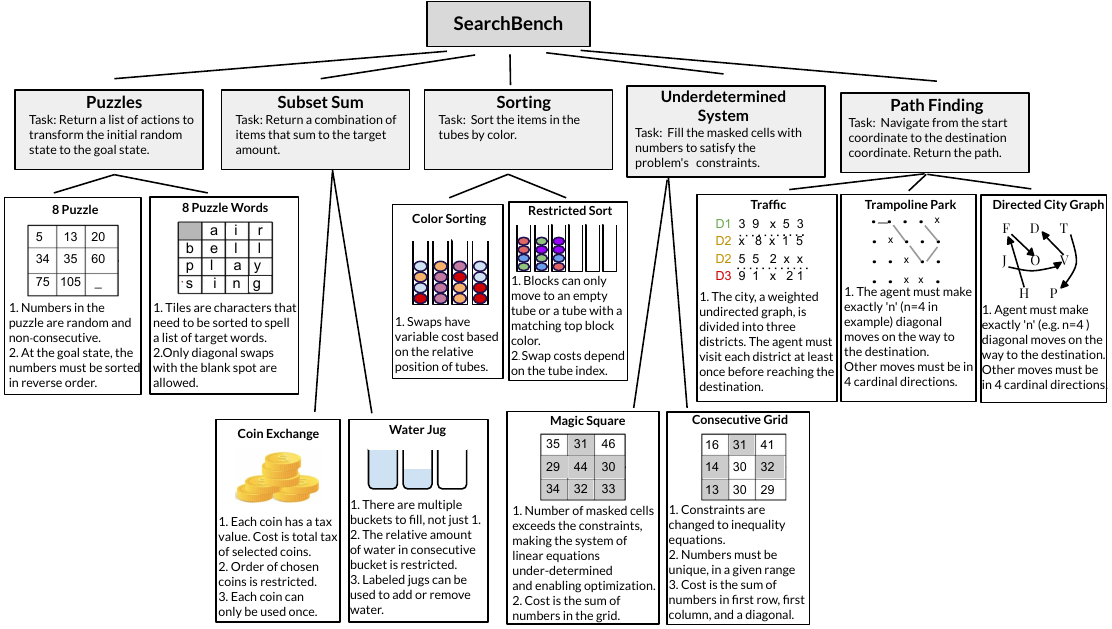}
    \caption{The taxonomy of SearchBench. The five nodes in level one represent the problem categories, and the 11 nodes in level two represent the problem types. We detail how the rules of known puzzles and combinatorial problems are modified in \ourbench{} to ensure LLMs haven't encountered solved instances of the problems during training.}
    \vspace{-0.01cm}\label{fig:searchbench-taxonomy}
\end{figure*}

\ourbench{} is challenging for LLMs due to several factors. The autoregressive architecture of current LLMs forces solving problems sequentially, which makes tasks requiring backtracking challenging \citep{dziri2024faith}. Additionally, natural language is not an ideal medium for accurately representing and updating intermediate states. Moreover, in combinatorial problems, the number of possible states increases exponentially with the number of actions, making naive exploration of the state space ineffective. Our empirical results show that even the most capable models solve less than 20\% of SearchBench problems end-to-end. To successfully solve SearchBench, a model must backtrack, consider multiple reasoning paths, and identify the most optimal outcome among many feasible options. These capabilities are essential for robust reasoning, making SearchBench a valuable benchmark for evaluating LLMs' reasoning capabilities as they continue to evolve.

\ourbench{} has five problem categories: (i) pathfinding, (ii) puzzles, (iii) subset sum, (iv) sorting, and (v) under-determined systems; further divided into 11 unique problem types. Each problem type is inspired by known puzzles and combinatorial problems, but includes modified rules to ensure they differ substantially from solved instances of the original problems that appear on the internet and are likely observed by LLMs during their pre-training. We generate about 100 instances of varying difficulty per problem type using an automatic pipeline, totaling 1,107 fixed instances. Each problem type in \ourbench{} also includes an automatic evaluation pipeline that assesses LLM-generated solutions on three dimensions: feasibility (choosing sequence of actions that adhere to the problem's rules), correctness (achieving the goal state), and optimality (finding the least cost solution).

Our analysis of LLMs' performance on our benchmark reveals a paradox. While models fail at the cognitively simple task of solving these puzzles step-by-step, we find their performance is significantly boosted when prompted to generate a complete A search algorithm. This approach succeeds because it leverages the model's strength in code generation, which is not an iterative task, offloading the exploration of a large action space from LLMs to code execution, resulting in improved performance (Fig. \ref{tab:all-results}). A* algorithm itself is a heuristic-based search algorithm known for its time efficiency and provable optimality,  offering advantages over BFS, which is computationally inefficient \citep{bfs}, and DFS, which does not guarantee an optimal solution \citep{dfs}. 

\begin{figure*}[t]
\centering
\includegraphics[width=\textwidth, height=7.8cm]
{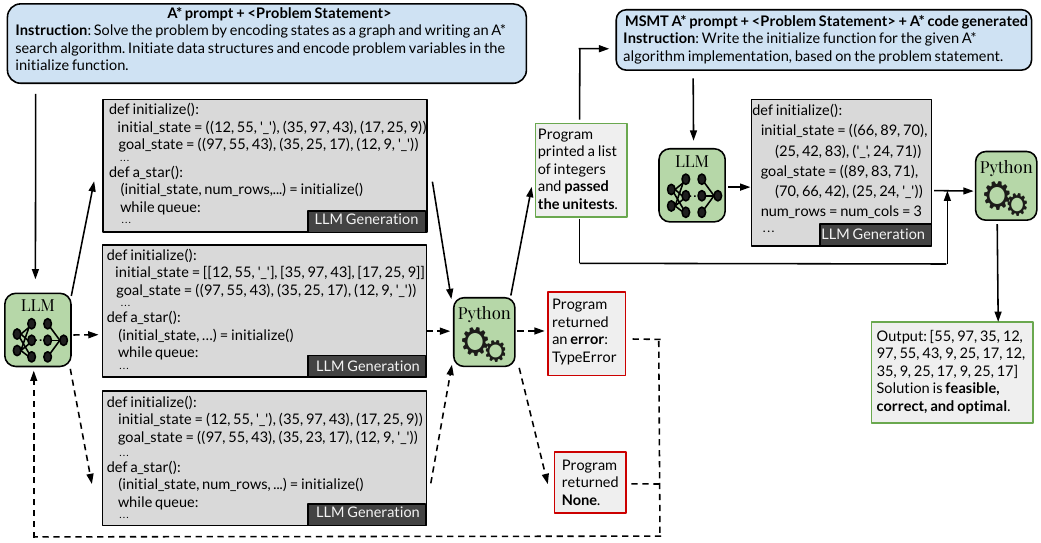}
\vspace{-0.1cm}\caption{Our \model{} (\smodel{}) A* prompting approach. }\label{fig:smt-astar}
\end{figure*}

Additionally, to improve the quality of the generated A* codes, motivated by recent work showing that multiple inferences and task decomposition improve LLM performance \citep{wang2022self, Yao2023TreeOT, long2023large}, we introduce the \model{} (\smodel{}) inference strategy. 
In this approach, we divide code generation into two stages. First, we prompt the model to write an instance-agnostic A* algorithm for the problem type. We then verify this implementation against a set of unit tests, without evaluating the solution, to check if (i) the code is executable, (ii) returns a list as output, and (iii) the data type of list elements is correct. Second, we instruct the model to implement the 'initialize' function, which encodes the variables specific to each problem instance. Our \smodel{} A* method (Fig. \ref{fig:smt-astar}) significantly improves LLMs' ability to solve search problems, outperforming other prompting strategies we used, including 0-shot text, 4-shot Chain-of-Thought (CoT)~\citep{Wei2022ChainOT} text, 0-shot code, and 4-shot A* prompting with naive greedy decoding. This method constitutes our strongest baseline on \ourbench{}; GPT-4 prompted with \smodel{} A* surpassing the o1-preview model. However, even using this approach, there remains considerable room for improvement on \ourbench{}, underscoring the challenge it presents in advancing LLMs' reasoning capabilities.

To summarize, our main contributions are as follows:
(i) We introduce the \ourbench{} benchmark, designed to evaluate LLMs' ability to solve combinatorial problems that require search and backtracking.
(ii) We demonstrate a key LLM bottleneck, showing that while models fail at iterative reasoning, they can successfully generate complex search algorithms; we then harness this capability with the \smodel{} A* framework to boost LLMs' performance on \ourbench{}. 
(iii)  We use \ourbench{} to thoroughly analyze the reasoning capabilities of several frontier LLMs, including GPT-4 and o1-preview, by employing various prompting and inference methods. This analysis uncovers the limitations of LLMs in tasks requiring iterative reasoning and highlights the potential for improving LLM performance on \ourbench{}.

\section{\ourbench{} Benchmark}

\ourbench{} includes five categories of problems: puzzles, subset sum, sorting, pathfinding, and under-determined systems. In theoretical computer science, combinatorial problems are classified into four types: existence, construction, enumeration, and optimization problems ~\citep{wilson2016combinatorics}. To ensure broad representation, we selected one problem category from each of these types for SearchBench. Particularly, subset sum problems represent the existence category, where the task is to determine if a subset of a given set sums to a specified value (refer to Tab. \ref{tab:example} for an example problem in this category). The 8-puzzle and 8-puzzle words fall under construction problems, which involve solving puzzles. Sorting problems, such as color sort and restricted sorting, are enumeration problems. Pathfinding problems are categorized as optimization problems.

Additionally, we introduce a new category of NP-hard combinatorial problmes in \ourbench{}, under-determined system problems. These problems consist of constraint satisfaction problems which are typically solved by defining a system of linear equations, and do not require search over states. We modified them to include fewer constraints than unknown variables, allowing for multiple correct solutions, and defined a unique cost function to enable search for a single optimal solution. This category was added in order to evaluate models' ability to generalize to novel combinatorial problems.

We selected 2-3 problem types for each category, resulting in 11 total problem types. Each type has a unique state space. For example, in 8-puzzle words, each state is an \(n \times m\) table of characters, while in coin exchange, each state is an ordered subset of given coins (See Appendix sec.~\ref{2_instances_prob_types} for more examples). Generally, our problems involve an initial state, a goal state, and a set of possible actions, and the task is to find a sequence of actions from the initial to the goal state with minimum cost. We modified the rules to ensure that solved instances of SearchBench were not encountered during the LLMs' massive internet-scale training. The \ourbench{} taxonomy and rule modifications are illustrated in Fig. \ref{fig:searchbench-taxonomy}.
\begin{table*}[ht]
\small
\centering
\begin{tabularx}{\textwidth}{|X|}
\hline
\textbf{Problem statement} \\
\hline
In the 'taxed coin exchange' problem, you are required to choose a subset of coins from this list \sethlcolor{lightgreen}\hl{[3, 6, 9, 10, 13, 15, 18, 5, 21, 19, 12, 15, 5, 9, 4, 16, 8, 4, 7, 7, 7, 2, 16, 14, 18, 3, 89, 21, 12, 10, 7, 14, 4, 11, 6, 20]}, such that the sum of the chosen coins adds up to \sethlcolor{lightgreen}\hl{229}. \sethlcolor{lightorange}\hl{Each coin in the list is unique and can only be used once}. Also \sethlcolor{lightorange}\hl{coins carry a tax value}. The tax values for each coin is \sethlcolor{lightgreen}\hl{14: 1, 89: 13, 2: 2, 5: 2, 4: 4, 6: 6, 8: 2, 16: 5, 21: 4, 20: 2, 18: 9, 11: 10, 10: 3, 12: 12, 15: 5, 13: 1, 3: 1, 19: 19, 7: 7, 9: 3}, where the tax for coins of the same value is the same. Also, \sethlcolor{lightorange}\hl{if the coin chosen is smaller than the previous one, it must have an even value, otherwise, if the coin is larger than or equal to the previous coin chosen, it must have an odd value}. The objective is to determine which subset of coins should be selected to \sethlcolor{lightorange}\hl{minimize the total tax paid}. The solution should be presented as a list of numbers, representing the value of the coins chosen in order, with the first coins chosen being in index 0, formatted in Python syntax.\\

\hline
\end{tabularx}
\caption{An instance of the 'Coin Exchange' problem: \sethlcolor{lightgreen}\hl{Green} indicated instance-specific components, and \sethlcolor{lightorange}\hl{orange} highlights the modified rules specific to \ourbench{}. GPT4 fails to generate a feasible solution for this instance using the three baseline prompting methods, but produces a correct, non-optimal solution using A* and \smodel{} A*.}
\vspace{-0.3cm} \label{tab:example}
\end{table*}

To construct \ourbench{}, we implemented an automatic generation pipeline for each problem type, ensuring each generated instance is solvable. We generated approximately 100 instances per type, resulting in a total of 1107 problem instances. The benchmark is then fixed. The generation pipelines can create instances with adjustable difficulty levels. Difficulty is defined by the state space size of the instance, with minimum difficulty requiring a few actions and maximum difficulty set such that problems could be solved correctly but not optimally by humans (See Appendix Sec. \ref{2_ss} for an analysis of the search space size). Hence, maximum human performance on SearchBench could be considered approximately 100\%. Moreover, studies like \citet{article, article-looks-easy} show that humans can solve the classic versions of SearchBench problems, but their performance declines as the state space size increases.



In contrast to other reasoning benchmarks \citep{saparov2022language, cobbe2021training, hendrycks2021measuring, patel2021nlp, clark2020transformers, tafjord2020proofwriter, sap-etal-2019-social, le-etal-2019-revisiting} that only measure correctness, to gain a more comprehensive understanding of LLM performance on \ourbench{}, our evaluation pipeline assesses LLM solutions across 3 dimensions: Feasibility, Correctness, and Optimality. Feasibility determines if any of the actions chosen violate the problem rules (e.g. passing through labyrinth walls). Correctness requires that the solution is both feasible and reaches the goal state from the given start state. Optimality indicates that the solution is both correct and has the minimum cost w.r.t. known optimum. For each \ourbench{} problem, we implemented a fast A* algorithm with a provably admissible and consistent heuristic, to produce the optimal solution. We ran this implementation for each instance in the benchmark to obtain its unique optimal solution.

We note that even though correctness implies feasibility, and optimally implies correctness, feasibility and correctness are valuable intermediate metrics in determining how close the models are to generating the fully correct solution. For example, in traffic problems, GPT4 often fails to record the first city visited, resulting in a feasible but incorrect solution. Defining feasibility helps distinguish this mostly correct implementation from more erroneous solutions. Correctness is stricter than feasibility and indicates that search-related tasks were implemented correctly, but the heuristic or recorded cost is incorrect, leading to non-optimal solutions.

\vspace{-0.05cm}

\section{Evaluated Methods}\label{sec:method}

\vspace{-0.05cm}

We use three baseline prompting methods to evaluate LLMs on \ourbench{}: 0-shot text, 4-shot CoT text, and 0-shot code. Additionally, we use two new code-based methods: 4-shot A* prompting and \smodel{} A*. Full prompts for each of the five approaches and GPT4's solution for an example problem are provided in Appendix Sec. \ref{full_prompts}.

To ensure the generality of our prompting methods, we selected one in-context example from each of the four \ourbench{} categories that are different from the category of the evaluated problem. This minimizes the similarity between the rules and context of the solved examples and the evaluated problem, and tests whether the model can solve unrelated combinatorial problems. Thus, if a model finds an optimal solution using these methods, it demonstrates true generalization rather than prompt-specific improvements. In Sec. \ref{section:ablation}, we further analyze the impact of including an example from the same top-level problem category. Additionally, 4-shot is the upper limit on the number of in-context examples due to the models' context length limit. For an analysis of the effect of fewer demonstrations (shots) on performance, see Appendix Sec. \ref{appendix:few-shot-astar}.

\noindent \textbf{0-shot text and 4-shot CoT text prompting methods:} In the text-based prompting methods, we instruct the model to solve the problem in an end-to-end manner, using text only. In 4-shot CoT prompts, the in-context examples include a representation of the intermediate states drawn using ASCII characters after each action to prevent hallucinations and illogical leaps in reasoning. 

\noindent \textbf{0-shot code prompting method:} This method instructs the LLM to produce a Python code that solves the given problem. The generated code is then executed to derive the final answer.

\noindent \textbf{A* Prompting:} In this approach, we prompt the LLM to implement an A* algorithm that solves $\mathcal{P}^C_i$ - a problem instance number $i$ of problem category $C$, providing four in-context examples of A* codes for four problems $\mathcal{P}^{\hat{C}}_j$, each selected from a different category $\hat{C} \ne C$. To implement A* for the target \ourbench{} problem, the LLM must perform abstract reasoning to devise a search strategy applicable to any state within the search space. This contrasts with solving problems end-to-end in text, where the model has access to the variables of each state, eliminating the need for abstract reasoning. However, end-to-end approaches require the model to perform every step of the iterative computations involved in searching the state space.

The in-context examples include detailed comments before each code segment, explaining the reasoning used to develop the strategy implemented within the code segment. These comments serve as CoT reasoning for devising the search strategy implemented in the code.

\noindent \textbf {\model{} (\smodel{}) A* Prompting:}
In this method, the model receives the same in-context examples as the `A* prompting', with different instructions. Here, inference is done in two stages as demonstrated in Fig. \ref{fig:smt-astar}. In the first stage, the model is instructed to implement the code as two functions: the 'a\_star' function includes an instance-agnostic A* algorithm for the target problem type, and the 'initialize' function encodes the variables given in the problem statement. We then verify if the generated code satisfies the following set of unit tests: (i) code is executable; (ii) code returns a list; (iii) and the list elements match the data type specified by the problem statement. If the code fails any unit test, \smodel{} re-generate the code. Next, in the second stage, the LLM is instructed to implement an `initialize' function, conditioned on the verified `a\_star' function from stage 1 for each instance of the problem type. The inclusion of simple unit tests, which can be expanded to more detailed tests if needed, offers a robust method for filtering out erroneous samples from the model's generations.

In our \smodel{} A* prompting approach, the model generates the full A* algorithm end-to-end without any external feedback, similar to how text-based prompting methods operate. Importantly, our \smodel{} A* does not rely on the majority vote of multiple solutions. Instead, the solution returned by the first model-generated code that passes the unit tests is taken as the final answer. This results in increased efficiency of \smodel{} A*, requiring only up to 1.5x number of inferences per problem on average compared to 5x-100x in majority vote approaches~\citep{wang2022self}.

\begin{figure*}[t]
    \centering
    \begin{subfigure}[b]{0.2305\textwidth}
        \centering
        \includegraphics[width=\textwidth, height = 2.45cm]
        {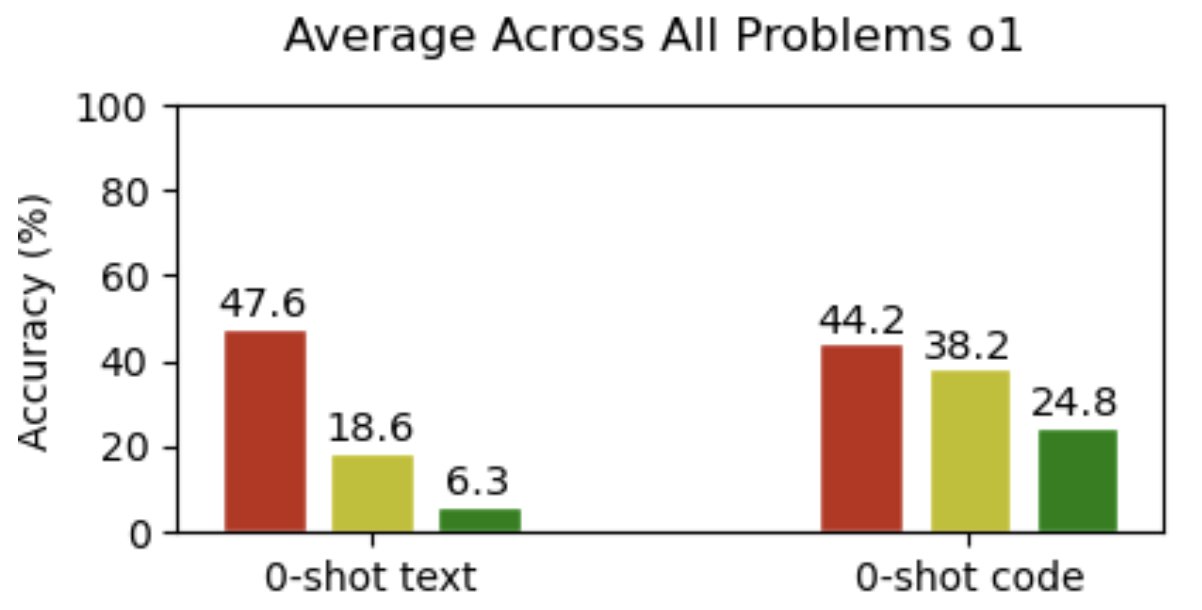}
    \end{subfigure}
        \vspace{0.05cm} 
    \hspace{-0.001\textwidth} 
    \begin{subfigure}[b]{0.35\textwidth}
        \centering
        \includegraphics[width=\textwidth, height = 2.6cm]{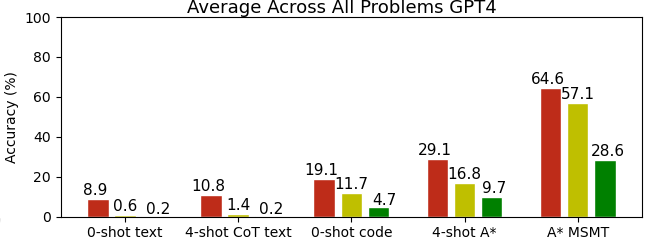}
    \end{subfigure}
            \vspace{0.08cm} 
    \hspace{-0.001\textwidth} 
    \begin{subfigure}[b]{0.35\textwidth}
        \centering
        \includegraphics[width=\textwidth, height = 2.6cm]{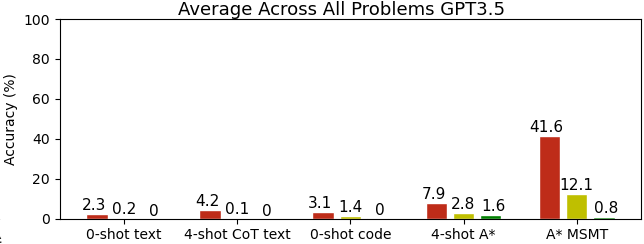}
    \end{subfigure}
            \vspace{0.08cm} 
    \hspace{-0.003\textwidth}
    \begin{subfigure}[b]{0.235\textwidth}
        \centering
        \includegraphics[width=\textwidth, height = 2.5cm]{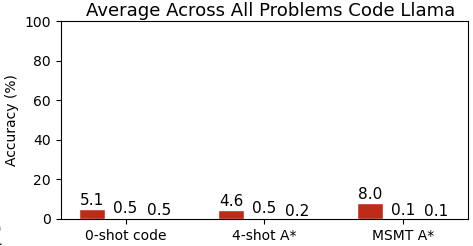}
    \end{subfigure}
                \vspace{0.08cm} 
    \hspace{-0.001\textwidth}
    \begin{subfigure}[b]{0.35\textwidth}
        \centering
        \includegraphics[width=\textwidth, height = 2.6cm]{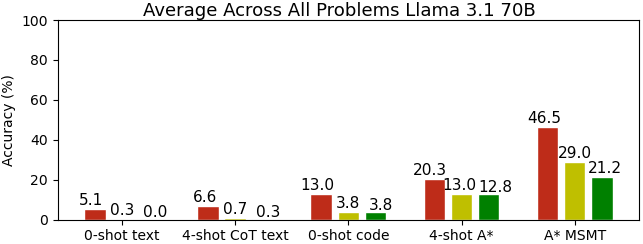}
    \end{subfigure}
                \vspace{0.08cm} 
    \hspace{-0.001\textwidth}
    \begin{subfigure}[b]{0.35\textwidth}
        \centering
        \includegraphics[width=\textwidth, height = 2.6cm]{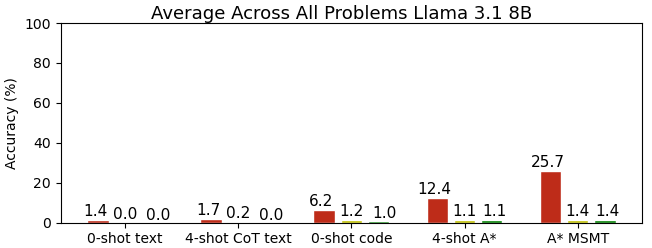}
    \end{subfigure}
                \vspace{0.08cm} 
    \hspace{0.02\textwidth}
     \begin{subfigure}[b]{0.2\textwidth}
        \centering
        \includegraphics[width=2cm, height =2.4cm ]{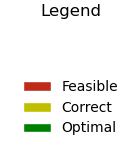}
    \end{subfigure}
    \hspace{0.001\textwidth} 
    \begin{subfigure}[b]{0.35\textwidth}
        \centering
        \includegraphics[width=\textwidth, height = 2.6cm]{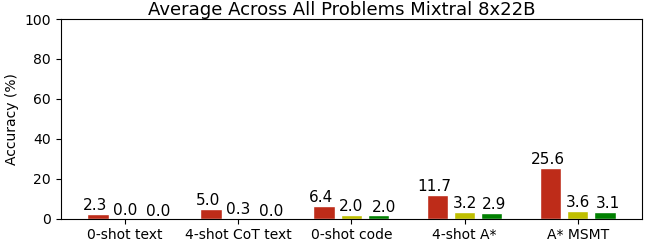}
    \end{subfigure}
    \begin{subfigure}[b]{0.35\textwidth}
        \centering
        \includegraphics[width=\textwidth, height = 2.6cm]{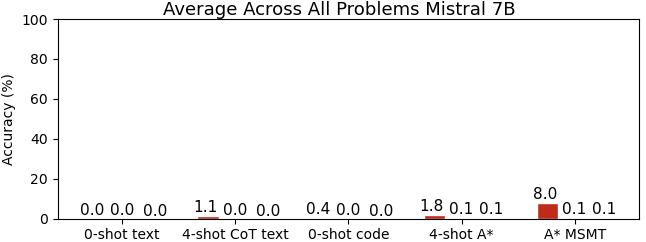}
    \end{subfigure}
    \vspace{-0.1cm}
    \caption{Average rate of feasible, correct, and optimal solutions for all problems using o1, GPT4, GPT3.5, Code Llama, Llama 3.1 70B, Llama 3.1 8B, Mixtral 8x22B, and Mistral 7B.}
\vspace{-0.3cm}\label{tab:all-results}
\end{figure*}

\vspace{-0.05cm}

\section{Related Work}

\noindent \textbf{Mathematical and Reasoning Benchmarks}: Evaluating LLMs~\citep{brown2020language, OpenAI2023GPT4TR, chatgpt, chung2024scaling, Chowdhery2022PaLMSL, Rae2021ScalingLM, taylor2022galactica,  thoppilan2022lamda} on mathematical and reasoning tasks has been a focus of recent research in natural language processing, leading to the development of benchmarks such as BIG-BENCH~\citep{srivastava2022beyond}, GSM8K~\citep{cobbe2021training}, AQUA~\citep{ling2017program}, SVAMP~\citep{patel2021nlp}, CommonsenseQA~\citep{talmor2018commonsenseqa}, StrategyQA~\citep{geva2021did}, and MATH~\citep{hendrycks2021measuring}. However, these benchmarks have limitations. GSM8K problems are relatively simple and often require a repetitive reasoning pattern to solve, and problems in BIG-BENCH are mostly single-step reasoning tasks. The MATH dataset, while more challenging, may not accurately assess a model's generalizable reasoning capabilities due to the advanced mathematical skills required. When prompted to solve problems using CoT prompting in text, LLMs perform well on these tasks; however, they fail on \ourbench{}'s problems, indicating that these benchmarks offer limited insight into LLMs’ ability to systematically explore a state space.

\noindent \textbf{Application of LLMs to Combinatorial Problems}:
Few studies such as~\citep{yang2023large, liu2024example, masoud2024exploring} have explored solving select combinatorial problems like the Traveling Salesman Problem with LLMs. \citet{mittal2024puzzlebench} introduced a dataset of combinatorial problems, "PuzzleBench". However, they only selected problems that can be represented in a symbolic solver (SMT2.0) and assumed there exists fixed pre-defined symbolic representations for input problems and outputs, limiting their datasets’ generalizability. Moreover, problems selected by \citet{mittal2024puzzlebench} and \citet{iklassov2024self} are instances of the classical combinatorial problems, raising issues of memorization, as algorithm implementations for instances of such problems are available online.

SearchBench stands out in several ways (i) Generalizability: Unlike PuzzleBench, SearchBench problems are described only in natural language, with no restrictions on rules or actions, and cover a wide verity of combinatorial problems, ensuring that a model capable of solving SearchBench can generalize to other combinatorial problems. (ii) Uniquely Modified Rules: This prevents memorization, as algorithms for classic versions of the problem are available online. (iii) Optimal Solution: Each problem type has a uniquely defined cost, ensuring a single optimal solution and avoiding multiple valid answers. (iv) Multi-Dimensional Evaluation: This provides deeper insights into how close models are to deriving the unique optimal solution. (v) Automated Instance Generation: This avoids data leakage or contamination, as new instances can be generated on demand.

\noindent \textbf{Prompting Strategies}:
Sophisticated prompting strategies have been developed to enhance models' reasoning abilities. One notable approach is Chain-of-Thought (CoT) prompting~\citep{Wei2022ChainOT}, which prompts LLMs to generate intermediate steps leading to the final output. This technique has led to advanced variations, including Tree-of-Thoughts~\citep{Yao2023TreeOT, long2023large}, and Graph-of-Thought~\citep{Yao2023BeyondCE, Lei2023BoostingLR, Besta2023GraphOT} methods that maintain a tree of intermediate generations. However, these methods rely on evaluating and rejecting intermediate steps, which does not integrate well with our problems. In search problems, intermediate states can't be easily classified as correct or incorrect, and all possible actions must be considered to find the optimal solution. Additionally, the state space of combinatorial problems grows exponentially, making it impractical for LLMs to navigate the frontier of the search tree without incorrectly disregarding most feasible states. 

Other prompting methods, such as Decomposition strategies~\citep{khot2022decomposed, zhou2022least, zhang2023natural}, simplify complex tasks into smaller, manageable subtasks to improve performance. Additionally, systems like LLM-Augmenter~\citep{peng2023check} rely on external databases to verify segments of the LLM's output. In this work, we propose the A* prompting strategy, where we prompt the model to solve problems by implementing a unique A* algorithm. Similarly, our A* MSMT approach decomposes the task of implementing the search algorithm into two stages and checks the model's generations against external validators; we use simple unit tests instead of external data sources or solved solution instances.

\section{Experiments}
 
We evaluated the performance of GPT4, GPT-3.5, and Code Llama Instruct 34B~\citep{roziere2023code} \footnote{Finetuned on the Phind dataset~\citep{phind}}, Llama 3.1 70B, Llama 3.1 8B, Mixtral 8x22B~\citep{mixtral}, and Mistral 7B~\citep{mistral7b} on \ourbench{}, using the following five prompting methods described in Sec. \ref{sec:method}: 0-shot text, 4-shot CoT text, 0-shot code, 4-shot A*, and 4-shot \smodel{} A*. Results are summarized in Fig. \ref{tab:all-results}.

\noindent \textbf{Implementation details:} We used GPT4, GPT-3.5 Turbo (GPT3.5 hereafter), and o1-preview (o1 hereafter) via OpenAI platform APIs. All code evaluations were conducted on a machine with 96 64-bit Intel Xeon Gold 5220R CPUs, a maximum speed of 4GHz, and a 71.5 MiB Level 3 cache.

\noindent \textbf{0-shot text and 4-shot CoT text prompting methods:} As shown in Fig. \ref{tab:all-results}, the correct solutions rate is below 1\% for all of the models using 0-shot text prompting, and less than 9\% of GPT4 solutions are feasible (follow the problem rules) using this method. This is expected as the exponentially growing state space size of \ourbench{} problems and the difficulty of backtracking during auto-regressive generation make it challenging to solve \ourbench{} problems using text-based prompting, even with the strongest LLMs. Moreover, 4-shot CoT text prompting only improves the rate of feasible solutions generated by less than 3\% for all models. This shows that the inherent complexity of search problems from \ourbench{} cannot be effectively addressed by text-based prompting alone.

\begin{figure*}[t]
    \centering
    \begin{subfigure}[b]{0.32\textwidth}
        \centering
        \includegraphics[width=2cm, height =2.4cm ]{Sample_Problem_vertical_legend.png}
    \end{subfigure}
    \hspace{0.01\textwidth}
        \vspace{0.05cm}
    \begin{subfigure}[b]{0.32\textwidth}
        \centering
        \includegraphics[width=\textwidth, height=2.5cm]{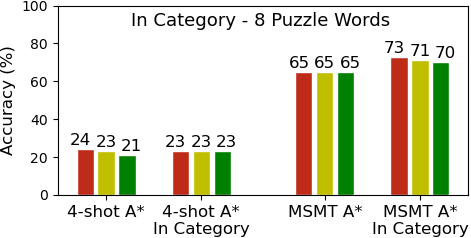}
    \end{subfigure}
    \hspace{-0.01\textwidth}
    \vspace{0.05cm}
    \begin{subfigure}[b]{0.32\textwidth}
        \centering
        \includegraphics[width=\textwidth, height=2.5cm]{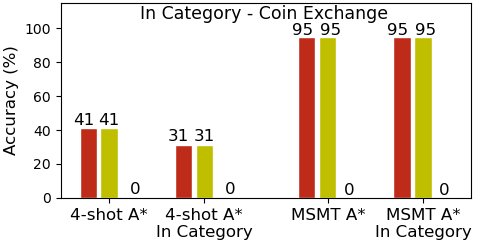}
    \end{subfigure}
    \vspace{0.05cm}
    \hspace{0.001\textwidth}
    \begin{subfigure}[b]{0.32\textwidth}
        \centering
        \includegraphics[width=\textwidth, height=2.5cm]{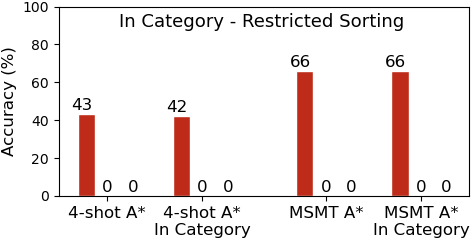}
    \end{subfigure}
    \hspace{0.01\textwidth}
    \begin{subfigure}[b]{0.32\textwidth}
        \centering
        \includegraphics[width=\textwidth, height=2.5cm]{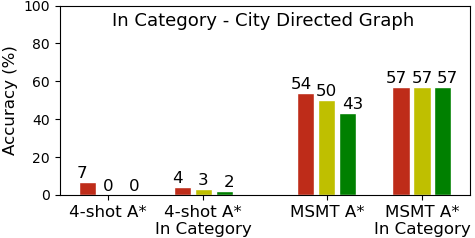}
    \end{subfigure}
    \hspace{0.01\textwidth}
    \begin{subfigure}[b]{0.31\textwidth}
        \centering
        \includegraphics[width=\textwidth, height=2.5cm]{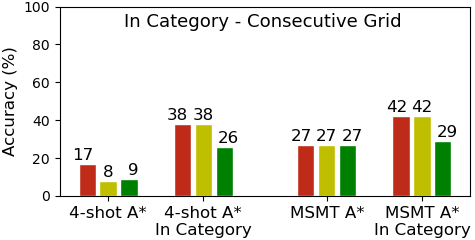}
    \end{subfigure}
    
    \caption{Comparing GPT4's performance, using A* prompting approaches, when one of the in-context examples is switched to a problem that shares the same category as the inference problem.}
    \vspace{-0.2cm}\label{fig:sim-problem-ablation}
\end{figure*}

Finally, we evaluate the recent o1 model~\citep{o1-preview}, which is designed for comprehensive reasoning. As shown, this model struggles with \ourbench{} problems, solving less than 19\% correctly with 0-shot text; however, it significantly outperformed other models in solving the problems end-to-end. We did not evaluate this model with A* prompting and \smodel{} A* due to its limited context length. Additionally, o1 solves problems with scaled compute at inference time, making \smodel{} unnecessary.

\noindent \textbf{0-shot code prompting method:} This prompting method improves performance over text-based prompting for all models except Mistral 7B, which remained close to 0\%. This is expected, as using Python to compute intermediate steps and execute the iterations of the algorithms devised by the LLMs reduces the load on the models. As seen in Fig. \ref{tab:all-results}, o1 solved 38.2\% of the problems correctly, 19.1\% of GPT4’s code generations result in a feasible solution, with only 11.7\% being correct. The next best performance was achieved by Llama 3.1 70B, which solved 13\% of the problems correctly. For an analysis of the computation time of programs generated by the LLMs, please refer to Appendix Sec. \ref{section:compute_time}.

\noindent \textbf{A* Prompting:} As shown in Fig. \ref{tab:all-results}, A* prompting improves the performance of all models on \ourbench{} except for Code Llama, which shows almost no improvement, indicating potential limitations of this model in in-context learning or following the given instructions. GPT-4's rates of feasible, correct, and optimal solutions increase by 10\%, 5\%, and 5\%, respectively, and Llama 3.1 70B's rates increase by 7\%, 9\%, and 9\%.

\noindent \textbf{\smodel{} A*:} As shown in Fig. \ref{tab:all-results}, \smodel{} A* prompting significantly boosts the performance of all models. Using this method, GPT4 correctly solved 57.1\% of \ourbench{} problems and achieved a 28.6\% rate of optimal solutions, outperforming o1. GPT4's performance improved consistently across all problem types compared to other prompting strategies (see Appendix Sec. \ref{appendix:detailed_gpt4_code} for for a detailed analysis of performance on each problem type). Other LLMs also showed strong improvements, except for Code Llama, which only improved in feasibility due as it still struggled to follow the instructions.
However, the 28.6\% optimal performance of GPT4 using \smodel{} A*, although inspiring, still leaves room for further improvements, underlining the importance of \ourbench{} for future research.



\section{Ablations and Analysis}
\label{section:ablation}

Here we provide a comprehensive analysis of \ourbench{} using GPT4. For further analysis, please refer to Appendix Sec.~\ref{appendix:few-shot-astar}, ~\ref{appendix:detailed_gpt4_code}, ~\ref{section:compute_time}, and ~\ref{2_ss}.

\noindent \textbf{Does including a more similar problem in prompt improve GPT4's performance?}  In our main experiments with A* and \smodel{} A* (Fig. \ref{tab:all-results}), we used four in-context examples, each from a different category than the target problem (Sec. \ref{sec:method}). This ensured that no exact segment of the target solution was included in the prompt,  hence better measuring LLM's reasoning generalization. Here, we evaluated GPT4's performance when a solved example from the same category but a different type as the evaluated problem, is included in the prompt. Results are summarized in Fig. \ref{fig:sim-problem-ablation}. We observed small improvements, with up to 15 additional instances solved. This indicates that \ourbench{} problems within the same category still differ significantly in rules, constraints, and target A* algorithm implementations.

The most significant improvement was observed for the Consecutive Grid problems from the under-determined systems category which involve filling in masked numbers according to constraints on the order of integers in a table. This category differs more significantly from other combinatorial problems, showing that including similar problems in the prompt leads to greater improvement for new tasks.

\begin{figure*}[th]
\centering
\includegraphics[height=5.3cm]
{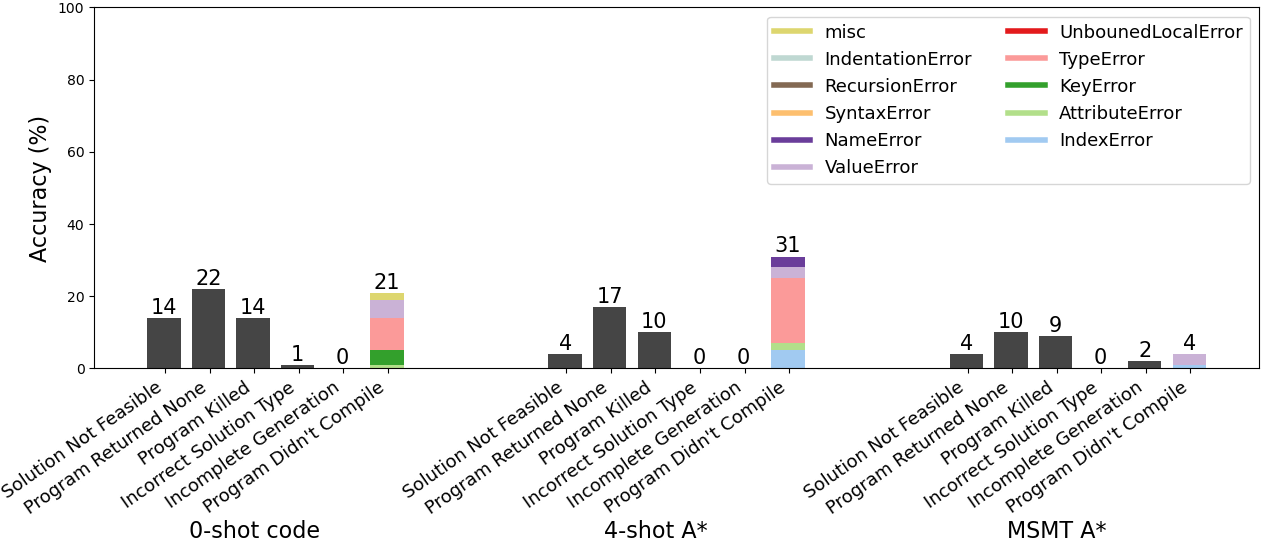}
\caption{Rate of errors returned by python programs generated by GPT4, categorized into 6 error types, calculated across all \ourbench{} problems with an infeasible solution.}
\vspace{-0.25cm}\label{fig:error-analysis-infeas}
\end{figure*}

\noindent \textbf{What types of runtime errors occur, and how often, when executing GPT4's implementations?} We analyzed the errors returned by GPT4's generated codes that resulted in infeasible solutions. The results are shown in Fig. \ref{fig:error-analysis-infeas}. We categorized errors into six types: (i) Solution Not Feasible: code executed but returned an infeasible solution; (ii) Program Returned None: program failed to find a solution; (iii) Program Killed: program did not finish within the allotted time; (iv) Incorrect Solution Type: returned solution had the wrong data type; (v) Incomplete Generation: model ran out of tokens; and (vi) Program Didn't Compile.

As shown in Fig. \ref{fig:error-analysis-infeas}, prompting the model with the A* method results in more non-compiling implementations compared to 0-shot code prompting. This is expected as A* is more complex and longer compared to the simpler algorithms typically implemented by the model using 0-shot code prompting, such as the greedy algorithm, BFS, or DFS. However, the number of infeasible solutions decreases with A* prompting, indicating that the model can better reason about the problem using this method. Comparing A* prompting to the \smodel{} A* method, we observe a significant decrease in errors that fail at least one unit test, such as 'Program Returned None', 'Program Killed', 'Incorrect Solution Type', 'Incomplete Generation', and 'Program Didn't Compile'.

\noindent \textbf{What are the most common reasoning errors made in GPT4's A* implementations?} We manually analyzed 50 A* codes generated by GPT4 that returned non-optimal solutions across five problems types: three pathfinding problems and two puzzle problems. These problems were chosen because GPT4 showed the least and greatest performance improvement, respectively, using A* prompting compared to 0-shot code (see Appendix Sec.  \ref{GPT4-code-based-table}). We identified seven distinct failure modes in the GPT4-generated A* implementations, each corresponding to a critical non-optimal within the overall search strategy, where failing any subtask results in a suboptimal solution. The results are summarized in Table \ref{table:comparison}, showing the percentage of 'correct reasoning' for each subtask, excluding coding errors. As shown, in pathfinding problems, the most common mistake was not recording the list of visited coordinates, a 13\% success rate, with the model often omitting the start coordinate in the path, leading to feasible but incorrect solutions. In puzzle problems, the frequent error was encoding the goal state, likely because our puzzles have unique goal states unlike the conventional 8-puzzle problem.


\begin{table}[ht]
\small
\centering
\renewcommand{\arraystretch}{1.2} 
\begin{tabular}{|m{3.3cm}|m{1.4cm}|m{1.3cm}|}
\hline
& \textbf{Pathfinding Problems} & \textbf{Puzzle Problems} \\
\hline
Encoding Initial State & 47\% & 100\% \\
\hline
Encoding Goal State & 74\% & 20\% \\
\hline
Recording the Path & 13\% & 70\% \\
\hline
Exit Condition & 70\% & 100\% \\
\hline
Iterating Successor States & 57\% & 100\% \\
\hline
Generating New State & 87\% & 100\% \\
\hline
Admissible \& Consistent Heuristic & 93\% & 60\% \\
\hline
\end{tabular}
\caption{The average accuracy of GPT4 on identified A* subtasks (failure modes) was analyzed using 50 code implementations for pathfinding and puzzle problems.}
\vspace{-0.4cm}\label{table:comparison}
\end{table}

\vspace{-0.07cm}
\section{Conclusion}
\vspace{-0.05cm}

In this work, we introduced \ourbench{}, a pioneering benchmark designed to evaluate LLMs' ability to solve new combinatorial search problems that require backtracking and considering multiple action sequences. We assessed LLMs using various text-based and code-based prompting methods and showed that while models fail at solving these puzzles step by step in natural language, their performance improves considerably when asked to implement an A* search algorithm—a cognitively more challenging task for humans but one that shifts the state space exploration burden from the model to code execution. This contrast reveals a key bottleneck in iterative reasoning and backtracking for autoregressive models, while also highlighting how their strengths in structured code generation, validated by simple unit tests, can be harnessed to overcome these limitations. For broader impact, see Appendix Sec. \ref{section:limit-impact}.

\vspace{-0.045cm}
\section{Limitations}
\vspace{-0.03cm}
The primary challenge in developing \ourbench{} was scaling the number of problem types. Designing unique search problems and creating pipelines to generate numerous instances with guaranteed solutions is both time-consuming and complex. Additionally, implementing a fast, instance-agnostic A* algorithm and developing evaluation pipelines to assess LLM-proposed solutions on multiple criteria further adds to the complexity.

\bibliography{main}

\begin{thebibliography}{54}
\providecommand{\natexlab}[1]{#1}

\bibitem[{Besta et~al.(2023)Besta, Blach, Kubicek, Gerstenberger, Gianinazzi, Gajda, Lehmann, Podstawski, Niewiadomski, Nyczyk, and Hoefler}]{Besta2023GraphOT}
Maciej Besta, Nils Blach, Ales Kubicek, Robert Gerstenberger, Lukas Gianinazzi, Joanna Gajda, Tomasz Lehmann, Michal Podstawski, Hubert Niewiadomski, Piotr Nyczyk, and Torsten Hoefler. 2023.
\newblock \href {https://api.semanticscholar.org/CorpusID:261030303} {Graph of thoughts: Solving elaborate problems with large language models}.
\newblock \emph{ArXiv}, abs/2308.09687.

\bibitem[{Bian et~al.(2023)Bian, Han, Sun, Lin, Lu, He, Jiang, and Dong}]{bian2023chatgpt}
Ning Bian, Xianpei Han, Le~Sun, Hongyu Lin, Yaojie Lu, Ben He, Shanshan Jiang, and Bin Dong. 2023.
\newblock Chatgpt is a knowledgeable but inexperienced solver: An investigation of commonsense problem in large language models.
\newblock \emph{arXiv preprint arXiv:2303.16421}.

\bibitem[{Brown et~al.(2020)Brown, Mann, Ryder, Subbiah, Kaplan, Dhariwal, Neelakantan, Shyam, Sastry, Askell et~al.}]{brown2020language}
Tom Brown, Benjamin Mann, Nick Ryder, Melanie Subbiah, Jared~D Kaplan, Prafulla Dhariwal, Arvind Neelakantan, Pranav Shyam, Girish Sastry, Amanda Askell, et~al. 2020.
\newblock Language models are few-shot learners.
\newblock \emph{Advances in neural information processing systems}, 33:1877--1901.

\bibitem[{Chowdhery et~al.(2022)Chowdhery, Narang, Devlin, Bosma, Mishra, Roberts, Barham, Chung, Sutton, Gehrmann, Schuh, Shi, Tsvyashchenko, Maynez, Rao, Barnes, Tay, Shazeer, Prabhakaran, Reif, Du, Hutchinson, Pope, Bradbury, Austin, Isard, Gur-Ari, Yin, Duke, Levskaya, Ghemawat, Dev, Michalewski, Garc{\'i}a, Misra, Robinson, Fedus, Zhou, Ippolito, Luan, Lim, Zoph, Spiridonov, Sepassi, Dohan, Agrawal, Omernick, Dai, Pillai, Pellat, Lewkowycz, Moreira, Child, Polozov, Lee, Zhou, Wang, Saeta, D{\'i}az, Firat, Catasta, Wei, Meier-Hellstern, Eck, Dean, Petrov, and Fiedel}]{Chowdhery2022PaLMSL}
Aakanksha Chowdhery, Sharan Narang, Jacob Devlin, Maarten Bosma, Gaurav Mishra, Adam Roberts, Paul Barham, Hyung~Won Chung, Charles Sutton, Sebastian Gehrmann, Parker Schuh, Kensen Shi, Sasha Tsvyashchenko, Joshua Maynez, Abhishek Rao, Parker Barnes, Yi~Tay, Noam~M. Shazeer, Vinodkumar Prabhakaran, Emily Reif, Nan Du, Benton~C. Hutchinson, Reiner Pope, James Bradbury, Jacob Austin, Michael Isard, Guy Gur-Ari, Pengcheng Yin, Toju Duke, Anselm Levskaya, Sanjay Ghemawat, Sunipa Dev, Henryk Michalewski, Xavier Garc{\'i}a, Vedant Misra, Kevin Robinson, Liam Fedus, Denny Zhou, Daphne Ippolito, David Luan, Hyeontaek Lim, Barret Zoph, Alexander Spiridonov, Ryan Sepassi, David Dohan, Shivani Agrawal, Mark Omernick, Andrew~M. Dai, Thanumalayan~Sankaranarayana Pillai, Marie Pellat, Aitor Lewkowycz, Erica Moreira, Rewon Child, Oleksandr Polozov, Katherine Lee, Zongwei Zhou, Xuezhi Wang, Brennan Saeta, Mark D{\'i}az, Orhan Firat, Michele Catasta, Jason Wei, Kathleen~S. Meier-Hellstern, Douglas Eck, Jeff Dean, Slav Petrov,
  and Noah Fiedel. 2022.
\newblock \href {https://api.semanticscholar.org/CorpusID:247951931} {Palm: Scaling language modeling with pathways}.
\newblock \emph{J. Mach. Learn. Res.}, 24:240:1--240:113.

\bibitem[{Chronicle et~al.(2006)Chronicle, Macgregor, Ormerod, and Burr}]{article-looks-easy}
Edward Chronicle, James Macgregor, Thomas Ormerod, and Alistair Burr. 2006.
\newblock \href {https://doi.org/10.1080/02724980543000033} {It looks easy! heuristics for combinatorial optimization problems}.
\newblock \emph{Quarterly journal of experimental psychology (2006)}, 59:783--800.

\bibitem[{Chung et~al.(2024)Chung, Hou, Longpre, Zoph, Tay, Fedus, Li, Wang, Dehghani, Brahma et~al.}]{chung2024scaling}
Hyung~Won Chung, Le~Hou, Shayne Longpre, Barret Zoph, Yi~Tay, William Fedus, Yunxuan Li, Xuezhi Wang, Mostafa Dehghani, Siddhartha Brahma, et~al. 2024.
\newblock Scaling instruction-finetuned language models.
\newblock \emph{Journal of Machine Learning Research}, 25(70):1--53.

\bibitem[{Clark et~al.(2020)Clark, Tafjord, and Richardson}]{clark2020transformers}
Peter Clark, Oyvind Tafjord, and Kyle Richardson. 2020.
\newblock Transformers as soft reasoners over language.
\newblock \emph{arXiv preprint arXiv:2002.05867}.

\bibitem[{Cobbe et~al.(2021)Cobbe, Kosaraju, Bavarian, Chen, Jun, Kaiser, Plappert, Tworek, Hilton, Nakano et~al.}]{cobbe2021training}
Karl Cobbe, Vineet Kosaraju, Mohammad Bavarian, Mark Chen, Heewoo Jun, Lukasz Kaiser, Matthias Plappert, Jerry Tworek, Jacob Hilton, Reiichiro Nakano, et~al. 2021.
\newblock Training verifiers to solve math word problems.
\newblock \emph{arXiv preprint arXiv:2110.14168}.

\bibitem[{Dziri et~al.(2024)Dziri, Lu, Sclar, Li, Jiang, Lin, Welleck, West, Bhagavatula, Le~Bras et~al.}]{dziri2024faith}
Nouha Dziri, Ximing Lu, Melanie Sclar, Xiang~Lorraine Li, Liwei Jiang, Bill~Yuchen Lin, Sean Welleck, Peter West, Chandra Bhagavatula, Ronan Le~Bras, et~al. 2024.
\newblock Faith and fate: Limits of transformers on compositionality.
\newblock \emph{Advances in Neural Information Processing Systems}, 36.

\bibitem[{Geva et~al.(2021)Geva, Khashabi, Segal, Khot, Roth, and Berant}]{geva2021did}
Mor Geva, Daniel Khashabi, Elad Segal, Tushar Khot, Dan Roth, and Jonathan Berant. 2021.
\newblock Did aristotle use a laptop? a question answering benchmark with implicit reasoning strategies.
\newblock \emph{Transactions of the Association for Computational Linguistics}, 9:346--361.

\bibitem[{Hendrycks et~al.(2021)Hendrycks, Burns, Kadavath, Arora, Basart, Tang, Song, and Steinhardt}]{hendrycks2021measuring}
Dan Hendrycks, Collin Burns, Saurav Kadavath, Akul Arora, Steven Basart, Eric Tang, Dawn Song, and Jacob Steinhardt. 2021.
\newblock Measuring mathematical problem solving with the math dataset.
\newblock \emph{arXiv preprint arXiv:2103.03874}.

\bibitem[{Iklassov et~al.(2024)Iklassov, Du, Akimov, and Takac}]{iklassov2024self}
Zangir Iklassov, Yali Du, Farkhad Akimov, and Martin Takac. 2024.
\newblock Self-guiding exploration for combinatorial problems.
\newblock \emph{arXiv preprint arXiv:2405.17950}.

\bibitem[{Khot et~al.(2022)Khot, Trivedi, Finlayson, Fu, Richardson, Clark, and Sabharwal}]{khot2022decomposed}
Tushar Khot, Harsh Trivedi, Matthew Finlayson, Yao Fu, Kyle Richardson, Peter Clark, and Ashish Sabharwal. 2022.
\newblock Decomposed prompting: A modular approach for solving complex tasks.
\newblock \emph{arXiv preprint arXiv:2210.02406}.

\bibitem[{Le et~al.(2019)Le, Boureau, and Nickel}]{le-etal-2019-revisiting}
Matthew Le, Y-Lan Boureau, and Maximilian Nickel. 2019.
\newblock \href {https://doi.org/10.18653/v1/D19-1598} {Revisiting the evaluation of theory of mind through question answering}.
\newblock In \emph{Proceedings of the 2019 Conference on Empirical Methods in Natural Language Processing and the 9th International Joint Conference on Natural Language Processing (EMNLP-IJCNLP)}, pages 5872--5877, Hong Kong, China. Association for Computational Linguistics.

\bibitem[{Lei et~al.(2023)Lei, Lin, Liao, and Ding}]{Lei2023BoostingLR}
Bin Lei, Pei-Hung Lin, Chunhua Liao, and Caiwen Ding. 2023.
\newblock \href {https://api.semanticscholar.org/CorpusID:261030743} {Boosting logical reasoning in large language models through a new framework: The graph of thought}.
\newblock \emph{ArXiv}, abs/2308.08614.

\bibitem[{Ling et~al.(2017)Ling, Yogatama, Dyer, and Blunsom}]{ling2017program}
Wang Ling, Dani Yogatama, Chris Dyer, and Phil Blunsom. 2017.
\newblock Program induction by rationale generation: Learning to solve and explain algebraic word problems.
\newblock \emph{arXiv preprint arXiv:1705.04146}.

\bibitem[{Liu et~al.(2024)Liu, Tong, Yuan, Lin, Luo, Wang, Lu, and Zhang}]{liu2024example}
Fei Liu, Xialiang Tong, Mingxuan Yuan, Xi~Lin, Fu~Luo, Zhenkun Wang, Zhichao Lu, and Qingfu Zhang. 2024.
\newblock An example of evolutionary computation+ large language model beating human: Design of efficient guided local search.
\newblock \emph{arXiv preprint arXiv:2401.02051}.

\bibitem[{Long(2023)}]{long2023large}
Jieyi Long. 2023.
\newblock Large language model guided tree-of-thought.
\newblock \emph{arXiv preprint arXiv:2305.08291}.

\bibitem[{Marcus(2020)}]{marcus2020next}
Gary Marcus. 2020.
\newblock The next decade in ai: four steps towards robust artificial intelligence.
\newblock \emph{arXiv preprint arXiv:2002.06177}.

\bibitem[{Masoud et~al.(2024)Masoud, Abdelhay, and Elhenawy}]{masoud2024exploring}
Mahmoud Masoud, Ahmed Abdelhay, and Mohammed Elhenawy. 2024.
\newblock Exploring combinatorial problem solving with large language models: A case study on the travelling salesman problem using gpt-3.5 turbo.
\newblock \emph{arXiv preprint arXiv:2405.01997}.

\bibitem[{Meta(2024)}]{llama3.1}
Meta. 2024.
\newblock \href {https://ai.meta.com/blog/meta-llama-3-1/} {Introducing llama 3.1: Our most capable models to date}.

\bibitem[{Mistral(2023{\natexlab{a}})}]{mistral7b}
Mistral. 2023{\natexlab{a}}.
\newblock \href {https://mistral.ai/news/announcing-mistral-7b/} {Mistral 7b}.

\bibitem[{Mistral(2023{\natexlab{b}})}]{mixtral}
Mistral. 2023{\natexlab{b}}.
\newblock \href {https://mistral.ai/news/mixtral-of-experts/} {Mixtral of experts}.

\bibitem[{Mittal et~al.(2024)Mittal, Kartik, Singla et~al.}]{mittal2024puzzlebench}
Chinmay Mittal, Krishna Kartik, Parag Singla, et~al. 2024.
\newblock Puzzlebench: Can llms solve challenging first-order combinatorial reasoning problems?
\newblock \emph{arXiv preprint arXiv:2402.02611}.

\bibitem[{OpenAI(2022)}]{chatgpt}
OpenAI. 2022.
\newblock \href {https://openai.com/index/chatgpt/} {Chatgpt: Optimizing language models for dialogue}.

\bibitem[{OpenAI(2023)}]{OpenAI2023GPT4TR}
OpenAI. 2023.
\newblock \href {https://api.semanticscholar.org/CorpusID:257532815} {Gpt-4 technical report}.
\newblock \emph{ArXiv}, abs/2303.08774.

\bibitem[{OpenAI(2024)}]{o1-preview}
OpenAI. 2024.
\newblock \href {https://openai.com/index/learning-to-reason-with-llms/} {Learning to reason with llms}.

\bibitem[{Patel et~al.(2021)Patel, Bhattamishra, and Goyal}]{patel2021nlp}
Arkil Patel, Satwik Bhattamishra, and Navin Goyal. 2021.
\newblock Are nlp models really able to solve simple math word problems?
\newblock \emph{arXiv preprint arXiv:2103.07191}.

\bibitem[{Peng et~al.(2023)Peng, Galley, He, Cheng, Xie, Hu, Huang, Liden, Yu, Chen et~al.}]{peng2023check}
Baolin Peng, Michel Galley, Pengcheng He, Hao Cheng, Yujia Xie, Yu~Hu, Qiuyuan Huang, Lars Liden, Zhou Yu, Weizhu Chen, et~al. 2023.
\newblock Check your facts and try again: Improving large language models with external knowledge and automated feedback.
\newblock \emph{arXiv preprint arXiv:2302.12813}.

\bibitem[{Phind(2023)}]{phind}
Phind. 2023.
\newblock \href {https://www.phind.com/blog/code-llama-beats-gpt4} {Beating gpt-4 on humaneval with a fine-tuned codellama-34b}.

\bibitem[{Pizlo and Li(2005)}]{article}
Zygmunt Pizlo and Zheng Li. 2005.
\newblock \href {https://doi.org/10.3758/BF03193214} {Solving combinatorial problems: The 15-puzzle}.
\newblock \emph{Memory \& cognition}, 33:1069--84.

\bibitem[{Qin et~al.(2023)Qin, Zhang, Zhang, Chen, Yasunaga, and Yang}]{qin2023chatgpt}
Chengwei Qin, Aston Zhang, Zhuosheng Zhang, Jiaao Chen, Michihiro Yasunaga, and Diyi Yang. 2023.
\newblock Is chatgpt a general-purpose natural language processing task solver?
\newblock \emph{arXiv preprint arXiv:2302.06476}.

\bibitem[{Rae et~al.(2021)Rae, Borgeaud, Cai, Millican, Hoffmann, Song, Aslanides, Henderson, Ring, Young, Rutherford, Hennigan, Menick, Cassirer, Powell, van~den Driessche, Hendricks, Rauh, Huang, Glaese, Welbl, Dathathri, Huang, Uesato, Mellor, Higgins, Creswell, McAleese, Wu, Elsen, Jayakumar, Buchatskaya, Budden, Sutherland, Simonyan, Paganini, Sifre, Martens, Li, Kuncoro, Nematzadeh, Gribovskaya, Donato, Lazaridou, Mensch, Lespiau, Tsimpoukelli, Grigorev, Fritz, Sottiaux, Pajarskas, Pohlen, Gong, Toyama, de~Masson~d'Autume, Li, Terzi, Mikulik, Babuschkin, Clark, de~Las~Casas, Guy, Jones, Bradbury, Johnson, Hechtman, Weidinger, Gabriel, Isaac, Lockhart, Osindero, Rimell, Dyer, Vinyals, Ayoub, Stanway, Bennett, Hassabis, Kavukcuoglu, and Irving}]{Rae2021ScalingLM}
Jack~W. Rae, Sebastian Borgeaud, Trevor Cai, Katie Millican, Jordan Hoffmann, Francis Song, John Aslanides, Sarah Henderson, Roman Ring, Susannah Young, Eliza Rutherford, Tom Hennigan, Jacob Menick, Albin Cassirer, Richard Powell, George van~den Driessche, Lisa~Anne Hendricks, Maribeth Rauh, Po-Sen Huang, Amelia Glaese, Johannes Welbl, Sumanth Dathathri, Saffron Huang, Jonathan Uesato, John F.~J. Mellor, Irina Higgins, Antonia Creswell, Nathan McAleese, Amy Wu, Erich Elsen, Siddhant~M. Jayakumar, Elena Buchatskaya, David Budden, Esme Sutherland, Karen Simonyan, Michela Paganini, L.~Sifre, Lena Martens, Xiang~Lorraine Li, Adhiguna Kuncoro, Aida Nematzadeh, Elena Gribovskaya, Domenic Donato, Angeliki Lazaridou, Arthur Mensch, Jean-Baptiste Lespiau, Maria Tsimpoukelli, N.~K. Grigorev, Doug Fritz, Thibault Sottiaux, Mantas Pajarskas, Tobias Pohlen, Zhitao Gong, Daniel Toyama, Cyprien de~Masson~d'Autume, Yujia Li, Tayfun Terzi, Vladimir Mikulik, Igor Babuschkin, Aidan Clark, Diego de~Las~Casas, Aurelia Guy, Chris
  Jones, James Bradbury, Matthew~G. Johnson, Blake~A. Hechtman, Laura Weidinger, Iason Gabriel, William~S. Isaac, Edward Lockhart, Simon Osindero, Laura Rimell, Chris Dyer, Oriol Vinyals, Kareem~W. Ayoub, Jeff Stanway, L.~L. Bennett, Demis Hassabis, Koray Kavukcuoglu, and Geoffrey Irving. 2021.
\newblock \href {https://api.semanticscholar.org/CorpusID:245353475} {Scaling language models: Methods, analysis \& insights from training gopher}.
\newblock \emph{ArXiv}, abs/2112.11446.

\bibitem[{Rein et~al.(2023)Rein, Hou, Stickland, Petty, Pang, Dirani, Michael, and Bowman}]{rein2023gpqa}
David Rein, Betty~Li Hou, Asa~Cooper Stickland, Jackson Petty, Richard~Yuanzhe Pang, Julien Dirani, Julian Michael, and Samuel~R Bowman. 2023.
\newblock Gpqa: A graduate-level google-proof q\&a benchmark.
\newblock \emph{arXiv preprint arXiv:2311.12022}.

\bibitem[{Roziere et~al.(2023)Roziere, Gehring, Gloeckle, Sootla, Gat, Tan, Adi, Liu, Remez, Rapin et~al.}]{roziere2023code}
Baptiste Roziere, Jonas Gehring, Fabian Gloeckle, Sten Sootla, Itai Gat, Xiaoqing~Ellen Tan, Yossi Adi, Jingyu Liu, Tal Remez, J{\'e}r{\'e}my Rapin, et~al. 2023.
\newblock Code llama: Open foundation models for code.
\newblock \emph{arXiv preprint arXiv:2308.12950}.

\bibitem[{Sap et~al.(2019)Sap, Rashkin, Chen, Le~Bras, and Choi}]{sap-etal-2019-social}
Maarten Sap, Hannah Rashkin, Derek Chen, Ronan Le~Bras, and Yejin Choi. 2019.
\newblock \href {https://doi.org/10.18653/v1/D19-1454} {Social {IQ}a: Commonsense reasoning about social interactions}.
\newblock In \emph{Proceedings of the 2019 Conference on Empirical Methods in Natural Language Processing and the 9th International Joint Conference on Natural Language Processing (EMNLP-IJCNLP)}, pages 4463--4473, Hong Kong, China. Association for Computational Linguistics.

\bibitem[{Saparov and He(2022)}]{saparov2022language}
Abulhair Saparov and He~He. 2022.
\newblock Language models are greedy reasoners: A systematic formal analysis of chain-of-thought.
\newblock \emph{arXiv preprint arXiv:2210.01240}.

\bibitem[{Srivastava et~al.(2022)Srivastava, Rastogi, Rao, Shoeb, Abid, Fisch, Brown, Santoro, Gupta, Garriga-Alonso et~al.}]{srivastava2022beyond}
Aarohi Srivastava, Abhinav Rastogi, Abhishek Rao, Abu Awal~Md Shoeb, Abubakar Abid, Adam Fisch, Adam~R Brown, Adam Santoro, Aditya Gupta, Adri{\`a} Garriga-Alonso, et~al. 2022.
\newblock Beyond the imitation game: Quantifying and extrapolating the capabilities of language models.
\newblock \emph{arXiv preprint arXiv:2206.04615}.

\bibitem[{Tafjord et~al.(2020)Tafjord, Mishra, and Clark}]{tafjord2020proofwriter}
Oyvind Tafjord, Bhavana~Dalvi Mishra, and Peter Clark. 2020.
\newblock Proofwriter: Generating implications, proofs, and abductive statements over natural language.
\newblock \emph{arXiv preprint arXiv:2012.13048}.

\bibitem[{Talmor et~al.(2018)Talmor, Herzig, Lourie, and Berant}]{talmor2018commonsenseqa}
Alon Talmor, Jonathan Herzig, Nicholas Lourie, and Jonathan Berant. 2018.
\newblock Commonsenseqa: A question answering challenge targeting commonsense knowledge.
\newblock \emph{arXiv preprint arXiv:1811.00937}.

\bibitem[{Taylor et~al.(2022)Taylor, Kardas, Cucurull, Scialom, Hartshorn, Saravia, Poulton, Kerkez, and Stojnic}]{taylor2022galactica}
Ross Taylor, Marcin Kardas, Guillem Cucurull, Thomas Scialom, Anthony Hartshorn, Elvis Saravia, Andrew Poulton, Viktor Kerkez, and Robert Stojnic. 2022.
\newblock Galactica: A large language model for science.
\newblock \emph{arXiv preprint arXiv:2211.09085}.

\bibitem[{Thoppilan et~al.(2022)Thoppilan, De~Freitas, Hall, Shazeer, Kulshreshtha, Cheng, Jin, Bos, Baker, Du et~al.}]{thoppilan2022lamda}
Romal Thoppilan, Daniel De~Freitas, Jamie Hall, Noam Shazeer, Apoorv Kulshreshtha, Heng-Tze Cheng, Alicia Jin, Taylor Bos, Leslie Baker, Yu~Du, et~al. 2022.
\newblock Lamda: Language models for dialog applications.
\newblock \emph{arXiv preprint arXiv:2201.08239}.

\bibitem[{Valmeekam et~al.(2022)Valmeekam, Olmo, Sreedharan, and Kambhampati}]{valmeekam2022large}
Karthik Valmeekam, Alberto Olmo, Sarath Sreedharan, and Subbarao Kambhampati. 2022.
\newblock Large language models still can't plan (a benchmark for llms on planning and reasoning about change).
\newblock \emph{arXiv preprint arXiv:2206.10498}.

\bibitem[{Wang et~al.(2022)Wang, Wei, Schuurmans, Le, Chi, Narang, Chowdhery, and Zhou}]{wang2022self}
Xuezhi Wang, Jason Wei, Dale Schuurmans, Quoc Le, Ed~Chi, Sharan Narang, Aakanksha Chowdhery, and Denny Zhou. 2022.
\newblock Self-consistency improves chain of thought reasoning in language models.
\newblock \emph{arXiv preprint arXiv:2203.11171}.

\bibitem[{Wei et~al.(2022)Wei, Wang, Schuurmans, Bosma, hsin Chi, Xia, Le, and Zhou}]{Wei2022ChainOT}
Jason Wei, Xuezhi Wang, Dale Schuurmans, Maarten Bosma, Ed~Huai hsin Chi, F.~Xia, Quoc Le, and Denny Zhou. 2022.
\newblock \href {https://api.semanticscholar.org/CorpusID:246411621} {Chain of thought prompting elicits reasoning in large language models}.
\newblock \emph{ArXiv}, abs/2201.11903.

\bibitem[{Wikipedia(2025{\natexlab{a}})}]{bfs}
Wikipedia. 2025{\natexlab{a}}.
\newblock \href {https://en.wikipedia.org/wiki/Breadth-first_search} {Breadth-first search}.

\bibitem[{Wikipedia(2025{\natexlab{b}})}]{dfs}
Wikipedia. 2025{\natexlab{b}}.
\newblock \href {https://en.wikipedia.org/wiki/Depth-first_search} {Depth-first search}.

\bibitem[{Wilson(2016)}]{wilson2016combinatorics}
R.J. Wilson. 2016.
\newblock \href {https://books.google.com/books?id=r2hdAQAACAAJ} {\emph{Combinatorics: A Very Short Introduction}}.
\newblock Very short introductions. Oxford University Press.

\bibitem[{Wu et~al.(2023)Wu, Qiu, Ross, Aky{\"u}rek, Chen, Wang, Kim, Andreas, and Kim}]{wu2023reasoning}
Zhaofeng Wu, Linlu Qiu, Alexis Ross, Ekin Aky{\"u}rek, Boyuan Chen, Bailin Wang, Najoung Kim, Jacob Andreas, and Yoon Kim. 2023.
\newblock Reasoning or reciting? exploring the capabilities and limitations of language models through counterfactual tasks.
\newblock \emph{arXiv preprint arXiv:2307.02477}.

\bibitem[{Yang et~al.(2023)Yang, Wang, Lu, Liu, Le, Zhou, and Chen}]{yang2023large}
Chengrun Yang, Xuezhi Wang, Yifeng Lu, Hanxiao Liu, Quoc~V Le, Denny Zhou, and Xinyun Chen. 2023.
\newblock Large language models as optimizers.
\newblock \emph{arXiv preprint arXiv:2309.03409}.

\bibitem[{Yao et~al.(2023{\natexlab{a}})Yao, Yu, Zhao, Shafran, Griffiths, Cao, and Narasimhan}]{Yao2023TreeOT}
Shunyu Yao, Dian Yu, Jeffrey Zhao, Izhak Shafran, Thomas~L. Griffiths, Yuan Cao, and Karthik Narasimhan. 2023{\natexlab{a}}.
\newblock \href {https://api.semanticscholar.org/CorpusID:258762525} {Tree of thoughts: Deliberate problem solving with large language models}.
\newblock \emph{ArXiv}, abs/2305.10601.

\bibitem[{Yao et~al.(2023{\natexlab{b}})Yao, Li, and Zhao}]{Yao2023BeyondCE}
Yao Yao, Z.~Li, and Hai Zhao. 2023{\natexlab{b}}.
\newblock \href {https://api.semanticscholar.org/CorpusID:258947684} {Beyond chain-of-thought, effective graph-of-thought reasoning in large language models}.
\newblock \emph{ArXiv}, abs/2305.16582.

\bibitem[{Zhang et~al.(2023)Zhang, Ge, Luo, Chuang, Gao, Gong, Wu, Kim, Meng, and Glass}]{zhang2023natural}
Tianhua Zhang, Jiaxin Ge, Hongyin Luo, Yung-Sung Chuang, Mingye Gao, Yuan Gong, Xixin Wu, Yoon Kim, Helen Meng, and James Glass. 2023.
\newblock Natural language embedded programs for hybrid language symbolic reasoning.
\newblock \emph{arXiv preprint arXiv:2309.10814}.

\bibitem[{Zhou et~al.(2022)Zhou, Sch{\"a}rli, Hou, Wei, Scales, Wang, Schuurmans, Cui, Bousquet, Le et~al.}]{zhou2022least}
Denny Zhou, Nathanael Sch{\"a}rli, Le~Hou, Jason Wei, Nathan Scales, Xuezhi Wang, Dale Schuurmans, Claire Cui, Olivier Bousquet, Quoc Le, et~al. 2022.
\newblock Least-to-most prompting enables complex reasoning in large language models.
\newblock \emph{arXiv preprint arXiv:2205.10625}.

\end{thebibliography}

\clearpage

\newpage
\appendix
\onecolumn
\addcontentsline{toc}{section}{Appendix} 
\part{Appendix} 
\parttoc 
\clearpage

\section{n-shot Ablation Experiments}
\label{appendix:few-shot-astar}
\begin{figure*}[t]
    \centering
    \begin{subfigure}[b]{0.2\textwidth}
        \centering
        \includegraphics[width=\textwidth, height = 2.25cm]{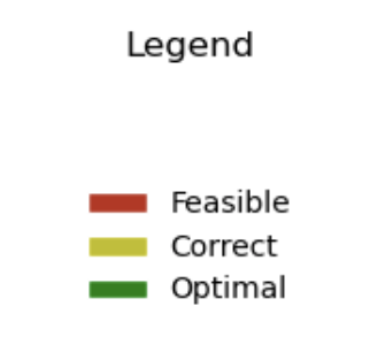}
    \end{subfigure}
    \hspace{-0.001\textwidth} 
    \begin{subfigure}[b]{0.32\textwidth}
        \centering
        \includegraphics[width=\textwidth, height = 2.84cm]
        {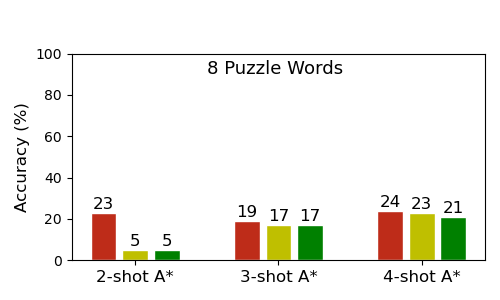}
    \end{subfigure}
    \hspace{-0.001\textwidth} 
    \begin{subfigure}[b]{0.32\textwidth}
        \centering
        \includegraphics[width=\textwidth, height = 2.684cm]{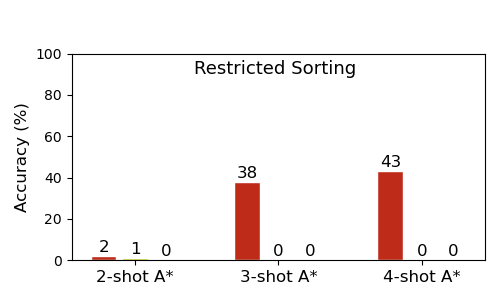}
    \end{subfigure}
    \hspace{-0.001\textwidth} 
    \begin{subfigure}[b]{0.32\textwidth}
        \centering
        \includegraphics[width=\textwidth, height = 2.684cm]{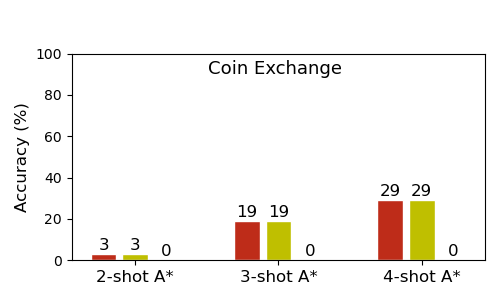}
    \end{subfigure}
    \hspace{-0.001\textwidth}
    \begin{subfigure}[b]{0.32\textwidth}
        \centering
        \includegraphics[width=\textwidth, height = 2.684cm]{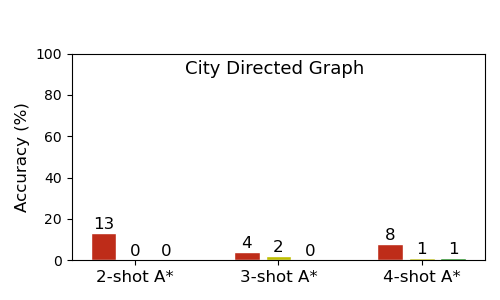}
    \end{subfigure}
    \hspace{-0.001\textwidth}
    \begin{subfigure}[b]{0.32\textwidth}
        \centering
        \includegraphics[width=\textwidth, height = 2.84cm]{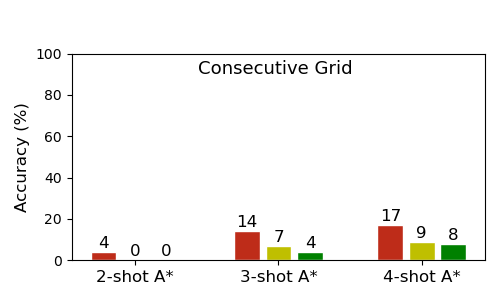}
    \end{subfigure}
    \caption{Comparing 2-shot, 3-shot and 4-shot performance of GPT4 between A*-prompting approaches.}
\label{fig:few-shot-astar}
\end{figure*}

To examine the effect of different numbers of demonstrations on GPT4's performance using A* and \smodel{} A* prompting methods, we performed ablation experiments with 2-shot and 3-shot A* prompts.  4-shot is upper limit on the number of in-context examples due to the context length constraints of the models, including GPT4.  In all few-shot experiments, the examples used in the prompts were not from the evaluated problem category. The results, summarized in Fig.  \ref{fig:few-shot-astar}, show a consistent trend of performance improvement with the addition of more examples, as expected.

\section{Detailed analysis of GPT4's performance on \ourbench{}}\label{appendix:detailed_gpt4_code}
Tab. \ref{GPT4-code-based-table} details GPT4 code-based method performance for each of \ourbench{}'s 11 problems. Consistently 4-shot A* prompting outperforms 0-shot code for most problems. Interestingly for problems in the pathfinding category, prompting GPT4 with 0-shot code outperforms A* prompting.

Examining closer, GPT4 mainly uses DFS for pathfinding in 0-shot code. While simpler than A*, DFS doesn't guarantee optimal solutions, as reflected in GPT4's high feasible and correct rates but lower optimal rates. Implementing A* with an admissible and consistent heuristic requires the model to implement a more complex strategy in the code involving additional constraints and more sophisticated data structures. This increases the likelihood of reasoning or coding errors, which could explain the dip in GPT4's performance using A* prompting compared to 0-shot code when solving these problems.

\begin{figure*}[t]
\centering
\includegraphics[width=14.3cm, height=3.8cm]{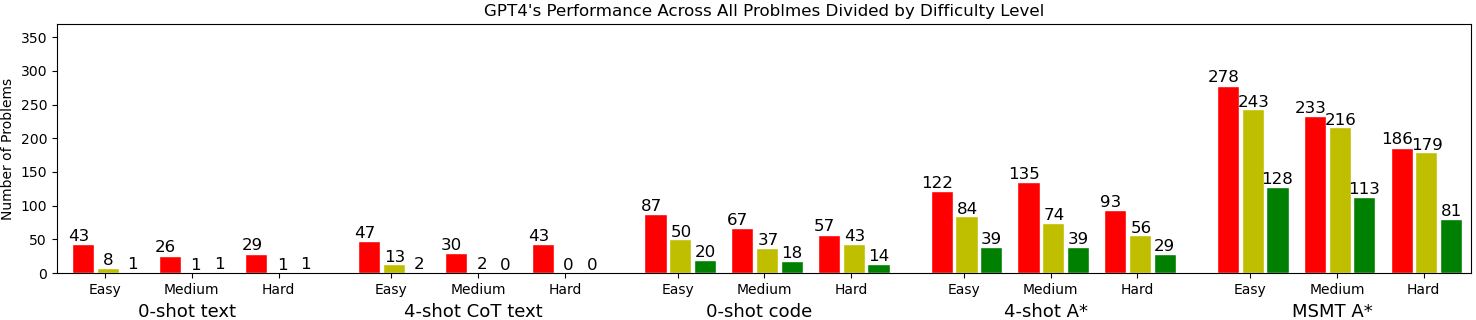}
\caption{Count of feasible, correct, and optimal solutions generated by GPT4 via code-based methods for 3 levels of problem difficulty.}
\label{gpt4:difficulty}
\end{figure*}

Figure \ref{gpt4:difficulty} further analyzes the relationship between problem difficulty (quantified by state space size of the problem) and the performance of GPT4. As observed, the model's performance is generally higher on easier problems, particularly in terms of the rate of correct solutions. This is expected, as easier problems have a smaller state space to explore. However, the performance of the model does not change drastically across different difficulty levels. This indicates that the combinatorial problems in \ourbench{} are intrinsically hard for LLMs to solve in text due to the requirement for backtracking. Moreover, the difference in implementing an A* search algorithm for a difficult or easy instance of \ourbench{} is limited to encoding the initial and goal states. The rest of the algorithm implementation task remains the same. This is the reason why the model's performance is comparable across different difficulty levels, both using text-based and code-based methods.

\begin{table*}[htbp]
\centering
\begin{tabularx}{\textwidth}{|l|>{\centering\arraybackslash}X|>{\centering\arraybackslash}X|>{\centering\arraybackslash}X|}
\hline
\small{\textbf{Problem}} & \small{\textbf{0-shot code}} & \small{\textbf{4-shot A*}} & \small{\textbf{MSMT A*}} \\
\hline
\small{8 Puzzle} & \small{\makebox[1cm][l]{F:  3} \makebox[1cm][l]{C: 0} \makebox[1cm][l]{O: 0}} & \small{\makebox[1cm][l]{F:  63} \makebox[1cm][l]{C: 60} \makebox[1cm][l]{O: 60}} & \small{\makebox[1cm][l]{F: 76} \makebox[1cm][l]{C: 68} \makebox[1cm][l]{O: 68}} \\

\small{8 Puzzle Words} & \small{\makebox[1cm][l]{F:  5} \makebox[1cm][l]{C:  5} \makebox[1cm][l]{O:  5}} & \small{\makebox[1cm][l]{F: 24} \makebox[1cm][l]{C: 23} \makebox[1cm][l]{O: 21}} & \small{\makebox[1cm][l]{F: 66} \makebox[1cm][l]{C: 65} \makebox[1cm][l]{O: 65}} \\
\hline
\small{Color Sorting} & \small{\makebox[1cm][l]{F: 17} \makebox[1cm][l]{C:  1} \makebox[1cm][l]{O:  1}} & \small{\makebox[1cm][l]{F: 41} \makebox[1cm][l]{C: 35} \makebox[1cm][l]{O:  6}} & \small{\makebox[1cm][l]{F: 91} \makebox[1cm][l]{C: 91} \makebox[1cm][l]{O:  0}} \\

\small{Restricted Sorting} & \small{\makebox[1cm][l]{F: 32} \makebox[1cm][l]{C:  0} \makebox[1cm][l]{O:  0}} & \small{\makebox[1cm][l]{F: 43} \makebox[1cm][l]{C:  0} \makebox[1cm][l]{O:  0}} & \small{\makebox[1cm][l]{F: 66} \makebox[1cm][l]{C:  0} \makebox[1cm][l]{O:  0}} \\
\hline
\small{Water Jug} & \small{\makebox[1cm][l]{F:  7} \makebox[1cm][l]{C:  7} \makebox[1cm][l]{O:  6}} & \small{\makebox[1cm][l]{F:  8} \makebox[1cm][l]{C:  8} \makebox[1cm][l]{O:  0}} & \small{\makebox[1cm][l]{F: 95} \makebox[1cm][l]{C: 95} 
\makebox[1cm][l]{O:  0}} \\

\small{Coin Exchange} & \small{\makebox[1cm][l]{F:  2} \makebox[1cm][l]{C:  1} \makebox[1cm][l]{O:  0}} & \small{\makebox[1cm][l]{F: 31} \makebox[1cm][l]{C: 31} \makebox[1cm][l]{O:  0}} & \small{\makebox[1cm][l]{F: 95} \makebox[1cm][l]{C: 95} \makebox[1cm][l]{O:  0}} \\
\hline
\small{Traffic} & \small{\makebox[1cm][l]{F: 65} \makebox[1cm][l]{C: 50} \makebox[1cm][l]{O: 13}} & \small{\makebox[1cm][l]{F: 24} \makebox[1cm][l]{C:  5} \makebox[1cm][l]{O:  5}} & \small{\makebox[1cm][l]{F: 65} \makebox[1cm][l]{C: 60} \makebox[1cm][l]{O: 60}} \\

\small{Trampoline Matrix} & \small{\makebox[1cm][l]{F: 27} \makebox[1cm][l]{C: 27} \makebox[1cm][l]{O: 22}} & \small{\makebox[1cm][l]{F: 51} \makebox[1cm][l]{C:  4} \makebox[1cm][l]{O:  4}} & \small{\makebox[1cm][l]{F: 57} \makebox[1cm][l]{C: 53} \makebox[1cm][l]{O: 46}} \\

\small{City Directed Graph} & \small{\makebox[1cm][l]{F: 29} \makebox[1cm][l]{C: 28} \makebox[1cm][l]{O:  1}} & \small{\makebox[1cm][l]{F:  7} \makebox[1cm][l]{C:  0} \makebox[1cm][l]{O:  0}} & \small{\makebox[1cm][l]{F: 55} \makebox[1cm][l]{C: 51} \makebox[1cm][l]{O: 45}} \\
\hline
\small{Magic Square} & \small{\makebox[1cm][l]{F:  3} \makebox[1cm][l]{C:  1} \makebox[1cm][l]{O:  0}} & \small{\makebox[1cm][l]{F:  8} \makebox[1cm][l]{C:  5} \makebox[1cm][l]{O:  0}} & \small{\makebox[1cm][l]{F: 14} \makebox[1cm][l]{C: 14} \makebox[1cm][l]{O:  0}} \\

\small{Consecutive Grid} & \small{\makebox[1cm][l]{F: 15} \makebox[1cm][l]{C:  2} \makebox[1cm][l]{O:  0}} & \small{\makebox[1cm][l]{F: 17} \makebox[1cm][l]{C:  9} \makebox[1cm][l]{O:  8}} & \small{\makebox[1cm][l]{F: 27} \makebox[1cm][l]{C: 27} \makebox[1cm][l]{O: 27}} \\
\hline
\end{tabularx}
\caption{GPT4's performance when prompted with our code-based approaches, on each problem type. The values are percentages of the feasible (F), correct (C), and optimal (O) solutions.}
\label{GPT4-code-based-table}
\end{table*}

\section{Compute Time of LLM-Generated Codes
}\label{section:compute_time}

\begin{figure*}[th]
    \centering
    \begin{subfigure}[b]{0.32\textwidth}
        \centering
        \includegraphics[width=\textwidth, height =2.3cm ]{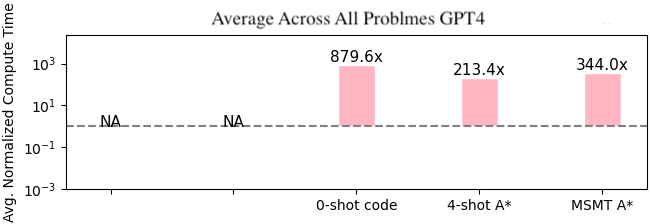}
    \end{subfigure}
    \hspace{0.001\textwidth}
    \begin{subfigure}[b]{0.315\textwidth}
        \centering
        \includegraphics[width=\textwidth, height =2.3cm]{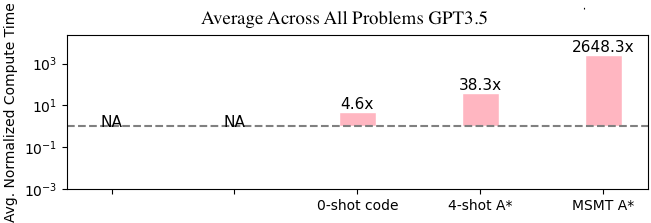}
    \end{subfigure}
    \hspace{0.001\textwidth}
    \begin{subfigure}[b]{0.315\textwidth}
        \centering
        \includegraphics[width=\textwidth,height =2.3cm]{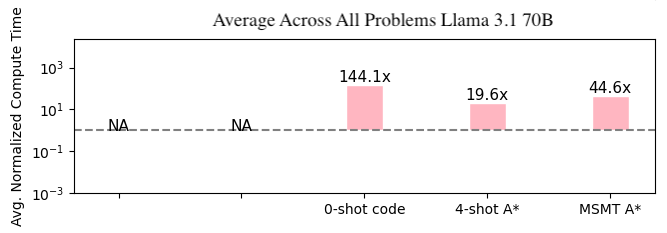}
    \end{subfigure}
    
    \caption{Average compute time of codes returning a correct solution normalized against the compute time of out A* implementation for all problems using GPT4, GPT3.5, Llama 3.1 70B.}\label{fig:runtime}
\end{figure*}

In this section, we analyze the computation time of programs generated by LLMs that produce correct solutions. We compare this time to the duration required to calculate the optimal solution for the problem instance using our fast A* implementation. This comparison provides insights into the efficiency of the algorithms generated by the LLMs. The average compute time of LLM-generated codes, normalized against the compute time of our A* implementation for the given instance, is reported in Fig. ~\ref{fig:runtime}.

Our findings indicate that LLM-generated implementations are significantly slower than our A* implementation. Specifically, GPT4's A* implementations were 213 times slower than the optimal A* solution, suggesting that GPT4's heuristics are still less efficient. Additionally, on average, GPT4’s 0-shot code generations that return a correct solution run 900 times slower than the optimal A* implementation. These results underscore the intrinsic difficulty of \ourbench{} problems, even when addressed through code generation.

\section{Broader Impact} \label{section:limit-impact}


Our research, which aims to assist the development of models capable of general reasoning and reliable problem-solving, has the potential to yield significant societal benefits. Combinatorial problems, like those in our dataset, are fundamental in fields such as robotics, logistics, network design, and industrial optimization. Developing models that can tackle unique versions of these problems by designing efficient algorithms or performing systematic searches end-to-end could greatly enhance AI's applicability across various domains.  However, this improvement in the reasoning capabilities of language models could also lead to job displacement, as these models could increasingly automate complex tasks traditionally performed by humans.

\clearpage

\section{SearchBench Variables}


\setlength{\fboxrule}{2pt}

\begin{table}[H]
\caption*{Table continued in the next page.}

\fbox{%
\begin{minipage}{\textwidth}
\begin{tabularx}{\textwidth}{>{\raggedright\arraybackslash}X}
\textbf{\large Variables} \\
\Xhline{3\arrayrulewidth}
\end{tabularx}

\begin{tabularx}{\textwidth}{>{\raggedright\arraybackslash}p{0.35\textwidth}X}
\textbf{diff\_sorted\_id} & A unique numeric identifier assigned to each problem instance within a specific problem type. These identifiers are ordered by difficulty level, that is the problem instance with diff\_sorted\_id of 1 is  easier than the instance with diff\_sorted\_id of 50. \\
\midrule
\textbf{problem\_statement} & A natural language description that outlines the problem to be solved. The problem statement is the sole piece of information given to language models when they are instructed to solve \ourbench{} problems. \\
\midrule
\textbf{problem\_type} & Indicates the problem type, out of 11 problem types in \ourbench{}, that this particular problem is an instance of. \\
\midrule
\textbf{problem\_category} & The specific category, out of the five predefined problem categories in \ourbench{}, to which this problem belongs. \\
\midrule
\textbf{relative\_diff\_score} & A numeric score that indicates the difficulty of this problem instance relative to other instances within the same problem type. This value is not comparable across different problem types. \\
\midrule
\textbf{opt\_solution} & A list of actions that, starting from the given initial state, lead to the goal state  with the minimum cost as defined by the problem's criteria. \\
\midrule
\textbf{opt\_solution\_cost} & The cost of the optimal solution for this problem instance.\\
\midrule
\textbf{opt\_solution\_compute\_t} & The time, in seconds, that our instance-agnostic A* implementation for the problem type took to solve this specific problem instance.\\
\midrule
\textbf{solution\_depth} & The number of actions required to reach the goal state from the given initial state with the minimum cost. This metric can be used to calculate an upper bound on the size of the search tree, represented as $b^{d}$, for this instance, where, b is an upper bound on the branching factor of the tree, which indicates the maximum number of actions leading to successor states from any given state, and d is the solution depth, representing the number of actions in the optimal solution. \\
\midrule
\textbf{max\_successor\_states} & The maximum number of successor states that can be reached from any given state in this problem. This value is an upper bound on the branching factor of the state search tree for this problem.\\
\end{tabularx}
\end{minipage}
}
\end{table}

\begin{table}[H]
\caption{This table provides a description of each column in SearchBench. Each row in SeacrhBench is an specific problem instance, and columns are fields of each instance.}
\fbox{%
\begin{minipage}{\textwidth}
\begin{tabularx}{\textwidth}{>{\raggedright\arraybackslash}X}
\textbf{\large Variables} \\
\Xhline{3\arrayrulewidth}
\end{tabularx}

\begin{tabularx}{\textwidth}{>{\raggedright\arraybackslash}p{0.35\textwidth}X}
\textbf{num\_vars\_per\_state} & An upper bound on the number of variables in each state of the problem. Given that the number of states grows exponentially for SearchBench problems, this value provides an estimate of the memory required to traverse the search tree of the problem.\\
\midrule
\textbf{is\_feasible\_args} & A list of variables of the problem instance that must be passed to the ‘is\_feasible’ function of the evaluation pipeline to determine whether a suggested solution adheres to the rules and constraints of the problem. \\
\midrule
\textbf{is\_correct\_args} & A list of variables in the problem statement of this instance that must be passed as arguments to the 'is\_correct' function in the evaluation pipeline, in order to evaluate the correctness of a suggested solution. \\
\midrule
\textbf{A*\_args} & Variables of this problem instance that must be passed to our A* implementation for the problem type to obtain the optimal solution for the instance. \\
\end{tabularx}
\end{minipage}
}
\end{table}

\section{Search Tree Size Analysis}\label{2_ss}

\begin{table}[H]
\caption{Statistics of metrics pertaining to the search-tree-size of a specific instance, compared across all instances within SearchBench.}
\noindent \fbox{%
\begin{minipage}{\textwidth}
\begin{tabularx}{\textwidth}{>{\raggedright\arraybackslash}X}
\textbf{\large Statistics} \\
\Xhline{3\arrayrulewidth}
\end{tabularx}

\begin{tabularx}{\textwidth}{>{\raggedright\arraybackslash}p{0.23\textwidth}p{0.1\textwidth}p{0.05\textwidth}p{0.07\textwidth}p{0.07\textwidth}p{0.08\textwidth}p{0.1\textwidth}p{0.05\textwidth}}
\textbf{name} & \textbf{type} & \textbf{min} & \textbf{median} & \textbf{max} & \textbf{mean} & \textbf{standard deviation} & \textbf{missing} \\
\midrule
opt\_solution\_compute\_t & float (seconds) & 0.018 & 0.068 & 599.044 & 17.363 & 67.513 & 0\% \\
\midrule
solution\_depth & int & 4 & 14 & 46 & 15.516 & 7.89 & 0\% \\
\midrule
max\_successor\_states & int & 4 & 12 & 132 & 24.633 & 24.622 & 0\% \\
\midrule
num\_vars\_per\_state & int & 2 & 13 & 60 & 14.785 & 12.05 & 0\% \\
\end{tabularx}
\end{minipage}
}

\end{table}

\begin{figure*}[th]
    \centering
    \begin{subfigure}[b]{0.32\textwidth}
        \centering
        \includegraphics[width=\textwidth, height =4cm ]{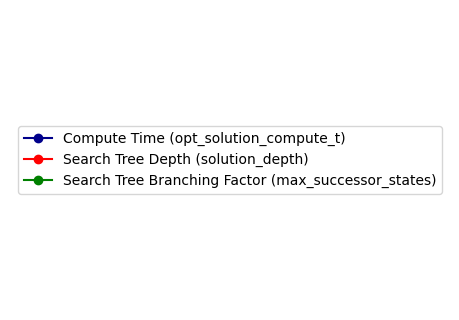}
    \end{subfigure}
    \hspace{-0.001\textwidth}
    \begin{subfigure}[b]{0.315\textwidth}
        \centering
        \includegraphics[width=\textwidth]{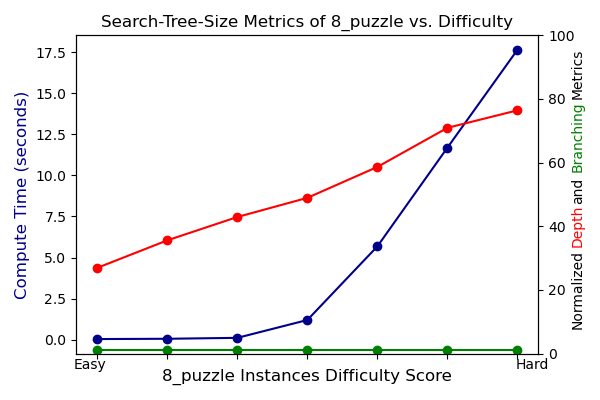}
    \end{subfigure}
    \hspace{-0.001\textwidth}
    \begin{subfigure}[b]{0.315\textwidth}
        \centering
        \includegraphics[width=\textwidth]{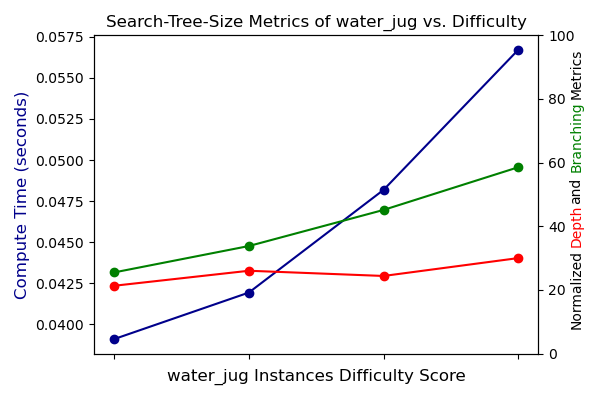}
    \end{subfigure}
    \hspace{-0.001\textwidth}
    \begin{subfigure}[b]{0.315\textwidth}
        \centering
        \includegraphics[width=\textwidth]{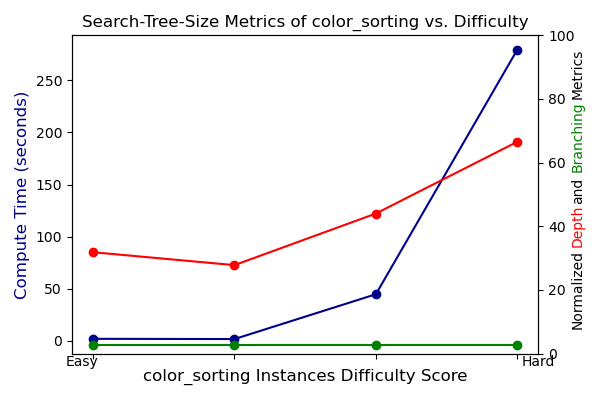}
    \end{subfigure}
    \hspace{-0.001\textwidth}
    \begin{subfigure}[b]{0.315\textwidth}
        \centering
        \includegraphics[width=\textwidth]{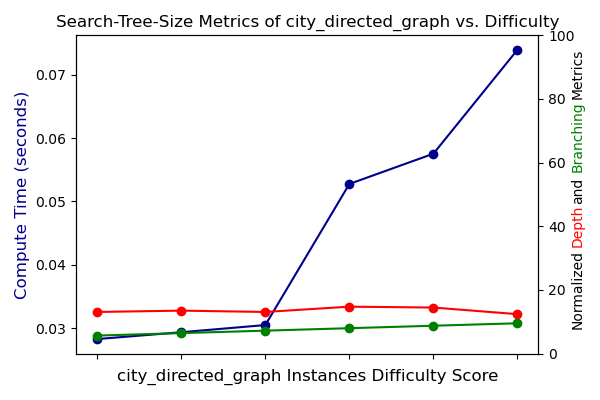}
    \end{subfigure}
    \hspace{-0.001\textwidth}
    \begin{subfigure}[b]{0.315\textwidth}
        \centering
        \includegraphics[width=\textwidth]{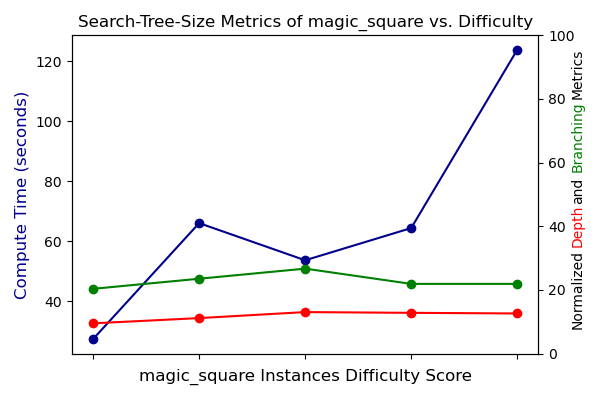}
    \end{subfigure}
    
    \caption{The plots depict the correlation between the increasing difficulty level and the corresponding increase in three metrics: the average depth of the solutions, the branching factor of the state search tree, and the exponential growth of the time required by our A* algorithm to solve the instances, demonstrated across five problem types in SearchBench.}\label{fig:tree-size}
\end{figure*}

Figure \ref{fig:tree-size} presents the relationship between the size of the state search tree and the difficulty levels of instances in SearchBench. It displays the average solution-depth and max\_successor\_state (normalized against the maximum and minimum solution\_depth and max\_successor\_state across all instances in SearchBench)  for one problem type from each of the five categories in SearchBench. Additionally, it shows the time our A* algorithm took to navigate the search tree for instances of variable difficulty (compute time is averaged across instances with the same difficulty). We used a machine with 96 64-bit Intel Xeon Gold 5220R CPUs with a maximum speed of 4GHz, and 71.5 MiB Level 3 cache to run the A* implementations.

The figure shows that the solution depth increases linearly with the difficulty scores of problem instances. However, for the city graph, it remains relatively constant, suggesting that the optimal number of hops to reach a destination node from a start node is consistent for our chosen range of directed graph connectivity and sizes (10 to 15 nodes).The max\_successor\_states, which represents the upper bound on the number of actions leading to successor states from each state, either remains constant or grows linearly with increasing difficulty level. This metric indicates the branching factor of the search tree size.

However, the compute time required to navigate this search tree grows much faster, exponentially, for most problems, as expected, given the search tree size is $b^{d}$, where b is the branching factor, and d is the solution depth. It's worth noting that we used a fast heuristic A* algorithm, which doesn't navigate the full search tree. An exhaustive algorithm like BFS, which explores every node, would result in a much faster exponential growth of compute times. In our experiments, a BFS implementation didn't finish executing even for some of the easiest instances within a 12-hour window.

\clearpage

\section{GPT4's MSMT A* Implementations for Two Instances of Each Problem Type}\label{2_instances_prob_types}

In this section, we present the A* algorithm generated by GPT4 using the MSMT A* prompting method, which successfully passed the unit tests. Additionally, we include GPT4's implementation of the 'initialize' function for a more challenging instance, generated in the second stage of the MSMT.

To facilitate the model's reasoning process when mapping the problem to a graph, designing steps of the A* algorithm, and reasoning about the admissibility or consistency of the heuristic, we employ a Chain of Thought (CoT) reasoning in text as comments in our in-context examples. We provide additional information before each code segment in our prompts that walk through the reasoning steps required to arrive at the strategy implemented in the code. Below, you can observe GPT4's comments that outline the intermediate reasoning steps the model generated to implement various code sections.

The implementations begin on the next page.

\clearpage

\minisection{8 Puzzle}

\begin{table}[H]
\centering
\begin{tabular}{|p{14cm}|}
\hline
\footnotesize \textbf{8\_puzzle instance (diff\_sorted\_id: 1)}: In the 8-puzzle game, you are given a grid with numbered square tiles arranged randomly and one tile missing. The goal is to arrange the tiles in descending order by sliding them into the empty space. The tiles can move in 4 directions: left, right, up, and down.  Given the initial state of the puzzle below, where the empty spot is represented as “\_”, provide the shortest list of tiles that need to be swapped with the empty spot to achieve the goal state. The goal state is when all tiles are in descending order, with the largest number in the top left corner, and the empty spot is in the bottom right corner. The solution should be a list of numbers in Python format, where each number represents the number on the tile that the empty spot is swapped with at each turn. Initial state of the puzzle: [[12, 55, '\_'], [35, 97, 43], [17, 25, 9]]\\
\hline
{\footnotesize A* algorithm generated by GPT4 for the problem above using A* \smodel{} approach that passed the unit tests:}\begin{lstlisting}[language=Python]
import heapq

def initialize():
   # Define the initial state of the puzzle as a 2d tuple
   initial_state = ((12, 55, '_'), (35, 97, 43), (17, 25, 9))
   num_rows = 3
   num_cols = 3
   goal_state = ((97, 55, 43), (35, 25, 17), (12, 9, '_'))
   visited_costs = {}
   visited_costs[initial_state] = 0
   queue = [(0, 0, [], initial_state)]
   return initial_state, num_rows, num_cols, goal_state, visited_costs, queue
  
def a_star():
   initial_state, num_rows, num_cols, goal_state, visited_costs, queue = initialize()
   while queue:
       _, g, actions, state = heapq.heappop(queue)
       # If the current state is the goal state, return the actions taken
       if state == goal_state:
           return actions
       # Generate all possible actions from the current state, which includes swapping the empty spot with any of the 4 neighboring tiles
       empty_spot_coord = [(i, j) for i, row in enumerate(state) for j, element in enumerate(row) if element == '_'][0]
       for d_row, d_col in [(0, -1), (0, 1), (1, 0), (-1, 0)]:
           new_row, new_col = empty_spot_coord[0] + d_row, empty_spot_coord[1] + d_col
           # Check if the new coordinate is valid, ie if the new tile is within the bounds of the puzzle
           if 0 <= new_row < num_rows and 0 <= new_col < num_cols:
               # Generate the new state
               new_state = [list(row[:]) for row in state]
               # Swap the empty spot with the new tile
               new_state[empty_spot_coord[0]][empty_spot_coord[1]], new_state[new_row][new_col] = new_state[new_row][new_col], new_state[empty_spot_coord[0]][empty_spot_coord[1]]
               new_state = tuple(tuple(row) for row in new_state)
               # The cost so far is the number of swaps made, as the task is to minimize the number of swaps required
               new_cost = g + 1
               if new_state not in visited_costs or new_cost < visited_costs[new_state]:
                   visited_costs[new_state] = new_cost
                   # The action taken is the number on the tile that was swapped with the empty spot
                   heapq.heappush(queue, (g + heuristic(new_state, goal_state), new_cost, actions + [int(new_state[empty_spot_coord[0]][empty_spot_coord[1]])], new_state))
   return None
\end{lstlisting} \\
\hline
\end{tabular}
\caption*{\footnotesize The GPT4's implementation of the A* algorithm continues on the next page.}
\end{table}

\begin{table}[H]
\centering
\begin{tabular}{|p{14cm}|}
\hline
{\footnotesize Continuation of GPT4's implementation: }\begin{lstlisting}[language=Python]
def heuristic(state, goal):
   # An admissible and consistent heuristic for this problem is the sum of the Manhattan distances from each tile in its current position to its position in the goal state
   # This heuristic relaxes the constraint that only the empty spot can be moved, and assumes that each tile can be moved to its goal position independently of the others
   # Thus the heuristic reports a lower estimate on the cost of reaching the goal state and is admissible
   # The heuristic is consistent because the estimated cost from the current state to the goal can never be greater than the sum of cost from the current node to a successor node plus the estimated cost from the successor node to the goal because the cost of moving the empty spot to an adjacent tile is 1, which is equal to the decrease in the Manhattan distance of the tile that was swapped with the empty spot
   # Also the cost of the goal state is 0, as all tiles are in their goal positions
   h = 0
   for i in range(len(state)):
       for j in range(len(state[0])):
           if state[i][j] != '_':
               goal_coord = [(x, y) for x, row in enumerate(goal) for y, element in enumerate(row) if element == state[i][j]][0]
               h += abs(i - goal_coord[0]) + abs(j - goal_coord[1])
   return h

print(a_star())

\end{lstlisting} \\
\hline
\end{tabular}
\caption{\footnotesize The A* algorithm was generated by GPT4 using the MSMT A* approach. For 8\_puzzle problem type, GPT4's A* implementation for the first instance passed the unit tests.}
\end{table}

\begin{table}[H]
\centering
\begin{tabular}{|p{14cm}|}
\hline
\footnotesize \textbf{8\_puzzle instance (diff\_sorted\_id: 94)}: In the 8-puzzle game, you are given a grid with numbered square tiles arranged randomly and one tile missing. The goal is to arrange the tiles in descending order by sliding them into the empty space. The tiles can move in 4 directions: left, right, up, and down.  Given the initial state of the puzzle below, where the empty spot is represented as “\_”, provide the shortest list of tiles that need to be swapped with the empty spot to achieve the goal state. The goal state is when all tiles are in descending order, with the largest number in the top left corner, and the empty spot is in the bottom right corner. The solution should be a list of numbers in Python format, where each number represents the number on the tile that the empty spot is swapped with at each turn. Initial state of the puzzle: [[98, 30, 85, 44, 59, 40], [56, 68, 50, 61, 47, 15], [37, 48, 24, 81, 22, '\_']]\\
\hline
{\footnotesize 'initialize' function generated by GPT4 for the problem above in the second stage of the MSMT approach}\begin{lstlisting}[language=Python]
def initialize():
   # Define the initial state of the puzzle as a 2d tuple
   initial_state = (('98', '30', '85', '44', '59', '40'),
                    ('56', '68', '50', '61', '47', '15'),
                    ('37', '48', '24', '81', '22', '_'))
   num_rows = 3
   num_cols = 6
   # Define the goal state of the puzzle as a 2d tuple
   goal_state = (('98', '85', '68', '61', '59', '56'),
                 ('50', '48', '47', '44', '40', '37'),
                 ('30', '24', '22', '15', '81', '_'))

   visited_costs = {}
   visited_costs[initial_state] = 0
   queue = [(0, 0, [], initial_state)]
   return initial_state, num_rows, num_cols, goal_state, visited_costs, queue
\end{lstlisting} \\
\hline
\end{tabular}
\caption{\footnotesize GPT4 was only successful in generating a feasible solution for this instance using the MSMT A* approach. The other four code and text-based prompting methods used in our experiments resulted in infeasible solutions.}
\end{table}

\minisection{8 Puzzle Words}

\begin{table}[H]
\centering
\begin{tabular}{|p{14.5cm}|}
\hline
\footnotesize \textbf{8\_puzzle\_words instance (diff\_sorted\_id: 1)}: In the game 'Sort the Chars', we are given a table of n by m dimensions. This table contains n words, each with m characters, except for the first word which has m - 1 characters. Each character is written on a separate tile. The objective of the game is to rearrange the characters such that row i spells the i-th word in the list, with the blank tile ('\_') placed in the top left corner of the board in the end. We can rearrange the tiles by swapping the blank space with any of its 4 diagonal neighboring tiles. Given the list of words and initial state of the board below, where the black space is represented as '\_', what is the shortest list of swap actions (reported in python syntax) that can sort the board into the given list of target words? The list must only include the 4 diagonal swap directions: up-right, down-right, up-left, or down-left, representing the direction in which the blank space was swpped in. Target words: cam, hill, pray, doer   The initial board: [['i', 'c', 'a', 'm'], ['h', 'p', 'l', 'o'], ['\_', 'r', 'a', 'y'], ['d', 'l', 'e', 'r']]\\
\hline
{\footnotesize A* algorithm generated by GPT4 for the problem above using A* \smodel{} approach that passed the unit tests:}\begin{lstlisting}[language=Python]
import heapq

def initialize():
   # Define the initial state of the board as a 2d tuple
   initial_state = (('i', 'c', 'a', 'm'), ('h', 'p', 'l', 'o'), ('_', 'r', 'a', 'y'), ('d', 'l', 'e', 'r'))
   num_rows = 4
   num_cols = 4
   # Define the goal state of the board as a 2d tuple
   goal_state = (('_', 'c', 'a', 'm'), ('h', 'i', 'l', 'l'), ('p', 'r', 'a', 'y'), ('d', 'o', 'e', 'r'))
   visited_costs = {}
   visited_costs[initial_state] = 0
   queue = [(0, 0, [], initial_state)]
   return initial_state, num_rows, num_cols, goal_state, visited_costs, queue
  
def a_star():
   initial_state, num_rows, num_cols, goal_state, visited_costs, queue = initialize()
   while queue:
       _, g, actions, state = heapq.heappop(queue)
       # If the current state is the goal state, return the actions taken
       if state == goal_state:
           return actions
       # Generate all possible actions from the current state, which includes swapping the blank space with any of its 4 diagonal neighbors
       for d_row, d_col in [(-1, 1), (1, 1), (-1, -1), (1, -1)]:
           # Get the coordinate of the blank space
           blank_coord = [(i, j) for i, row in enumerate(state) for j, element in enumerate(row) if element == '_'][0]
           new_row, new_col = blank_coord[0] + d_row, blank_coord[1] + d_col
           # Check if the new coordinate is valid, ie if the new position is within the bounds of the board
           if 0 <= new_row < num_rows and 0 <= new_col < num_cols:
               # Generate the new state
               new_state = [list(row[:]) for row in state]
               new_state[blank_coord[0]][blank_coord[1]], new_state[new_row][new_col] = new_state[new_row][new_col], new_state[blank_coord[0]][blank_coord[1]]
               new_state = tuple(tuple(row) for row in new_state)
               # The cost so far is the number of actions taken, as the task is to minimize the number of actions required to sort the board
               new_cost = g + 1
               if new_state not in visited_costs or new_cost < visited_costs[new_state]:
                   visited_costs[new_state] = new_cost
                   # The direction of the swap action is the direction in which the blank space was moved
                   if d_row == -1 and d_col == 1:
                       action = 'up-right'
                   elif d_row == 1 and d_col == 1:
                       action = 'down-right'
                   elif d_row == -1 and d_col == -1:
                       action = 'up-left'
                   elif d_row == 1 and d_col == -1:
                       action = 'down-left'
                   heapq.heappush(queue, (g + heuristic(new_state, goal_state), new_cost, actions + [action], new_state))
   return None
\end{lstlisting} \\
\hline
\end{tabular}
\caption*{\footnotesize The GPT4's implementation of the A* algorithm continues on the next page.}
\end{table}

\begin{table}[H]
\centering
\begin{tabular}{|p{14cm}|}
\hline
{\footnotesize Continuation of GPT4's implementation: }\begin{lstlisting}[language=Python]
def heuristic(state, goal):
   # An admissible and consistent heuristic for this problem is the count of tiles that are not in their goal position
   # This heuristic relaxes the constraint that only the blank space can be moved, and only to a diagonal position
   # It is admissible because it never overestimates the cost to reach the goal, as each misplaced tile must be moved at least once
   # It's consistent because moving a tile reduces the heuristic cost of the successor node by a max of 1 (if the moved tile is placed in its goal position), which is equal to the cost of reaching the successor node
   # Thus h(n) is always less than or equal to c(n, n')(equal to 1) + h(n')
   # And the cost of the goal state is 0, as all tiles will be in their goal position
   h = 0
   for i in range(len(state)):
       for j in range(len(state[0])):
           if state[i][j] != goal[i][j]:
               h += 1
   return h

print(a_star())

\end{lstlisting} \\
\hline
\end{tabular}
\caption{\footnotesize The A* algorithm was generated by GPT4 using the MSMT A* approach. For 8\_puzzle\_words problem type, GPT4's A* implementation for the first instance passed the unit tests.}
\end{table}

\begin{table}[H]
\centering
\begin{tabular}{|p{14cm}|}
\hline
\footnotesize \textbf{8\_puzzle\_words instance (diff\_sorted\_id: 93)}: In the game 'Sort the Chars', we are given a table of n by m dimensions. This table contains n words, each with m characters, except for the first word which has m - 1 characters. Each character is written on a separate tile. The objective of the game is to rearrange the characters such that row i spells the i-th word in the list, with the blank tile ('\_') placed in the top left corner of the board in the end. We can rearrange the tiles by swapping the blank space with any of its 4 diagonal neighboring tiles. Given the list of words and initial state of the board below, where the black space is represented as '\_', what is the shortest list of swap actions (reported in python syntax) that can sort the board into the given list of target words? The list must only include the 4 diagonal swap directions: up-right, down-right, up-left, or down-left, representing the direction in ehich the blank space was swpped in. Target words: helen, adelea, bumble, iodism, trullo   The initial board: [['d', 'h', '\_', 'l', 'e', 'n'], ['a', 'a', 'e', 'b', 'e', 'u'], ['m', 'u', 'i', 'b', 'e', 'e'], ['i', 'o', 'd', 'm', 's', 'l'], ['t', 'r', 'l', 'l', 'l', 'o']]\\
\hline
{\footnotesize 'initialize' function generated by GPT4 for the problem above in the second stage of the MSMT approach}\begin{lstlisting}[language=Python]
def initialize():
   # Define the initial state of the board as a 2d tuple
   initial_state = (('d', 'h', '_', 'l', 'e', 'n'),
                    ('a', 'a', 'e', 'b', 'e', 'u'),
                    ('m', 'u', 'i', 'b', 'e', 'e'),
                    ('i', 'o', 'd', 'm', 's', 'l'),
                    ('t', 'r', 'l', 'l', 'l', 'o'))
   num_rows = 5
   num_cols = 6
   # Define the goal state of the board as a 2d tuple
   goal_state = (('_', 'h', 'e', 'l', 'e', 'n'),
                 ('a', 'd', 'e', 'l', 'e', 'a'),
                 ('b', 'u', 'm', 'b', 'l', 'e'),
                 ('i', 'o', 'd', 'i', 's', 'm'),
                 ('t', 'r', 'u', 'l', 'l', 'o'))
   visited_costs = {}
   visited_costs[initial_state] = 0
   queue = [(0, 0, [], initial_state)]
   return initial_state, num_rows, num_cols, goal_state, visited_costs, queue
\end{lstlisting} \\
\hline
\end{tabular}
\caption{\footnotesize GPT4 was only successful in generating an optimal solution for this instance using the MSMT A* approach. The other four code and text-based prompting methods used in our experiments resulted in infeasible solutions.}
\end{table}

\clearpage

\minisection{Coin Exchange}

\begin{table}[H]
\centering
\begin{tabular}{|p{14.5cm}|}
\hline
\footnotesize \textbf{coin\_exchange instance (diff\_sorted\_id: 1)}: In the 'taxed coin exchange' problem, you are required to choose a subset of coins from this list [14, 2, 5, 8, 65, 5, 16, 2, 5, 2, 12, 17, 16, 15, 20, 4, 19, 9, 10, 64, 2, 10, 14, 2, 17], such that the sum of the chosen coins adds up to 208. Each coin in the list is unique and can only be used once. Also coins carry a tax value. The tax values for each coin is {20: 13, 4: 2, 17: 16, 19: 8, 64: 4, 5: 3, 14: 10, 10: 9, 65: 11, 16: 2, 12: 7, 8: 5, 2: 2, 15: 14, 9: 8}, where the tax for coins of the same value is the same. Also, if the coin chosen is smaller than the previous one, it must have an even value, otherwise, if the coin is larger than or equal to the previous coin chosen, it must have an odd value. The objective is to determine which subset of coins should be selected to minimize the total tax paid. The solution should be presented as a list of numbers, representing the value of the coins chosen in order, with the first coins chosen being in index 0, formatted in Python syntax.\\
\hline
{\footnotesize A* algorithm generated by GPT4 for the problem above using A* \smodel{} approach that passed the unit tests:}\begin{lstlisting}[language=Python]
import heapq

def initialize():
   # Define the initial state of the problem, which includes the coins available, the sum of the coins chosen so far, the tax paid so far, and the value of the last coin chosen
   initial_state = ((14, 2, 5, 8, 65, 5, 16, 2, 5, 2, 12, 17, 16, 15, 20, 4, 19, 9, 10, 64, 2, 10, 14, 2, 17), 0, 0, 0)
   goal_sum = 208
   tax_values = {20: 13, 4: 2, 17: 16, 19: 8, 64: 4, 5: 3, 14: 10, 10: 9, 65: 11, 16: 2, 12: 7, 8: 5, 2: 2, 15: 14, 9: 8}
   visited_costs = {}
   visited_costs[initial_state] = 0
   queue = [(0, 0, [], initial_state)]
   return initial_state, goal_sum, tax_values, visited_costs, queue
  
def a_star():
   initial_state, goal_sum, tax_values, visited_costs, queue = initialize()
   while queue:
       _, g, actions, state = heapq.heappop(queue)
       # Unpack the information encoded in each state
       coins, sum_coins, tax_paid, last_coin = state
       # If the sum of the coins chosen so far equals the goal sum, return the coins chosen
       if sum_coins == goal_sum:
           return actions
       # If the sum of the coins chosen so far is less than the goal sum, generate all possible actions from the current state, which includes choosing any of the remaining coins
       elif sum_coins < goal_sum:
           for coin in coins:
               # Check if the new state, containing the new coin, would be valid; ie the coin must be even if it is smaller than the last coin chosen, and odd if it is larger than or equal to the last coin chosen
               if (coin < last_coin and coin % 2 == 0) or (coin >= last_coin and coin % 2 == 1):
                   # Generate the new state
                   new_state = (tuple(c for c in coins if c != coin), sum_coins + coin, tax_paid + tax_values[coin], coin)
                   # The additional cost of this state is the tax paid for the new coin as we are trying to minimize the total tax paid
                   new_cost = g + tax_values[coin]
                   if new_state not in visited_costs or new_cost < visited_costs[new_state]:
                       visited_costs[new_state] = new_cost
                       # The new coin must be added to the actions
                       heapq.heappush(queue, (g + heuristic(new_state, goal_sum), new_cost, actions + [coin], new_state))
   return None
\end{lstlisting} \\
\hline
\end{tabular}
\caption*{\footnotesize The GPT4's implementation of the A* algorithm continues on the next page.}
\end{table}

\begin{table}[H]
\centering
\begin{tabular}{|p{14cm}|}
\hline
{\footnotesize Continuation of GPT4's implementation: }\begin{lstlisting}[language=Python]
def heuristic(state, goal_sum):
   # An admissible and consistent heuristic for this problem is the difference between the goal sum and the sum of the coins chosen so far
   # The heuristic assumes we can reach the goal sum without paying any additional tax, thus reporting a lower cost estimate to reach the goal state, making it admissible
   # The heuristic is consistent, ie non-decreasing along the path to the goal state, because the cost of choosing a coin is the tax paid for that coin, which is always greater than or equal to 0, the decrease in the difference between the goal sum and the sum of the coins chosen so far
   # Also the cost of the goal state is 0, as the sum of the coins chosen equals the goal sum and there are no more coins to choose

   return goal_sum - state[1]


print(a_star())

\end{lstlisting} \\
\hline
\end{tabular}
\caption{\footnotesize The A* algorithm was generated by GPT4 using the MSMT A* approach. For coin\_exchange problem type, GPT4's A* implementation for the first instance passed the unit tests.}
\end{table}

\begin{table}[H]
\centering
\begin{tabular}{|p{14cm}|}
\hline
\footnotesize \textbf{coin\_exchange instance (diff\_sorted\_id: 104)}: In the 'taxed coin exchange' problem, you are required to choose a subset of coins from this list [18, 23, 23, 36, 27, 10, 25, 14, 6, 43, 6, 31, 5, 43, 3, 13, 16, 8, 11, 6, 39, 44, 41, 28, 12, 8, 1, 14, 22, 23, 11, 24, 37, 9, 2, 31, 27, 10, 9, 38, 37, 4, 4, 17, 18, 36, 18, 32, 3, 12, 4, 42, 8, 4, 4, 16, 30, 40, 13, 41], such that the sum of the chosen coins adds up to 441. Each coin in the list is unique and can only be used once. Also coins carry a tax value. The tax values for each coin is {6: 4, 18: 3, 38: 1, 23: 2, 16: 8, 12: 2, 4: 2, 24: 13, 28: 17, 30: 17, 1: 1, 9: 9, 27: 8, 11: 4, 37: 16, 41: 19, 13: 3, 31: 8, 44: 8, 22: 7, 43: 13, 10: 3, 17: 8, 36: 7, 39: 3, 25: 17, 40: 10, 8: 6, 32: 16, 14: 4, 2: 1, 42: 3, 3: 2, 5: 2}, where the tax for coins of the same value is the same. Also, if the coin chosen is smaller than the previous one, it must have an even value, otherwise, if the coin is larger than or equal to the previous coin chosen, it must have an odd value. The objective is to determine which subset of coins should be selected to minimize the total tax paid. The solution should be presented as a list of numbers, representing the value of the coins chosen in order, with the first coins chosen being in index 0, formatted in Python syntax.\\
\hline
{\footnotesize 'initialize' function generated by GPT4 for the problem above in the second stage of the MSMT approach}\begin{lstlisting}[language=Python]
def initialize():
   def initialize():
   # Define the initial state of the problem, which includes the coins available, the sum of the coins chosen so far, the tax paid so far, and the value of the last coin chosen
   initial_state = ((18, 23, 23, 36, 27, 10, 25, 14, 6, 43, 6, 31, 5, 43, 3, 13, 16, 8, 11, 6, 39, 44, 41, 28, 12, 8, 1, 14, 22, 23, 11, 24, 37, 9, 2, 31, 27, 10, 9, 38, 37, 4, 4, 17, 18, 36, 18, 32, 3, 12, 4, 42, 8, 4, 4, 16, 30, 40, 13, 41), 0, 0, 0)
   goal_sum = 441
   tax_values = {6: 4, 18: 3, 38: 1, 23: 2, 16: 8, 12: 2, 4: 2, 24: 13, 28: 17, 30: 17, 1: 1, 9: 9, 27: 8, 11: 4, 37: 16, 41: 19, 13: 3, 31: 8, 44: 8, 22: 7, 43: 13, 10: 3, 17: 8, 36: 7, 39: 3, 25: 17, 40: 10, 8: 6, 32: 16, 14: 4, 2: 1, 42: 3, 3: 2, 5: 2}
   visited_costs = {}
   visited_costs[initial_state] = 0
   queue = [(0, 0, [], initial_state)]
   return initial_state, goal_sum, tax_values, visited_costs, queue
\end{lstlisting} \\
\hline
\end{tabular}
\caption{\footnotesize GPT4 was only successful in generating a correct (but non-optimal) solution for this instance of the coin\_exchange problem type using the MSMT A* approach. The other four code and text-based prompting methods used in our experiments resulted in infeasible solutions.}
\end{table}

\minisection{Water Jug}

\begin{table}[H]
\centering
\begin{tabular}{|p{14.5cm}|}
\hline
\footnotesize \textbf{water\_jug instance (diff\_sorted\_id: 1)}: In the 'taxed coin exchange' problem, you are required to choose a subset of coins from this list [14, 2, 5, 8, 65, 5, 16, 2, 5, 2, 12, 17, 16, 15, 20, 4, 19, 9, 10, 64, 2, 10, 14, 2, 17], such that the sum of the chosen coins adds up to 208. Each coin in the list is unique and can only be used once. Also coins carry a tax value. The tax values for each coin is {20: 13, 4: 2, 17: 16, 19: 8, 64: 4, 5: 3, 14: 10, 10: 9, 65: 11, 16: 2, 12: 7, 8: 5, 2: 2, 15: 14, 9: 8}, where the tax for coins of the same value is the same. Also, if the coin chosen is smaller than the previous one, it must have an even value, otherwise, if the coin is larger than or equal to the previous coin chosen, it must have an odd value. The objective is to determine which subset of coins should be selected to minimize the total tax paid. The solution should be presented as a list of numbers, representing the value of the coins chosen in order, with the first coins chosen being in index 0, formatted in Python syntax.\\
\hline
{\footnotesize A* algorithm generated by GPT4 for the problem above using A* \smodel{} approach that passed the unit tests:}\begin{lstlisting}[language=Python]
import heapq

def initialize():
   # Define the initial state of the buckets as a tuple
   initial_state = (0, 0, 0)
   # Define the goal state of the buckets as a tuple
   goal_state = (274, 297, 343)
   # Define the capacities of the jugs as a list
   jug_capacities = [41, 26, 44, 75, 40, 136]
   # Initialize a dictionary to store the cost of reaching each visited state
   visited_costs = {}
   visited_costs[initial_state] = 0
   # Initialize a priority queue of states not yet visited, with the initial state as the first element. The priority of each element is the cost to reach that state (g) + the estimate remaining cost (h) to reach the goal
   # Record the actions required to get to each state in a list; no actions performed to reach the initial state
   queue = [(0, 0, [], initial_state)]
   return initial_state, goal_state, jug_capacities, visited_costs, queue
  
def a_star():
   # The initialize function initializes and returns the visited_costs dictionary and the priority queue and encodes all of the variables given in the problem (ie the initial and goal state of the buckets and the capacities of the jugs)
   initial_state, goal_state, jug_capacities, visited_costs, queue = initialize()
   # While there are un-visited states
   while queue:
       # Pop the state with the lowest sum of the cost so far and estimated cost to the goal from the queue
       _, g, actions, state = heapq.heappop(queue)
       # We can check if the current state is the goal state with a simple equality check, as the goal state is predefined
       if state == goal_state:
           return actions
       # Generate all possible actions from the current state, which includes filling each of the 3 buckets with each of the 6 jugs, and emptying each of the 3 buckets with each of the 6 jugs
       for i in range(len(jug_capacities)):
           for j in range(len(state)):
               # Check if the new state would be valid, ie if the bucket can be filled with the jug without overflowing and if the amount of water in each bucket does not exceed the amount of water in the bucket placed after it
               if state[j] + jug_capacities[i] <= goal_state[j] and (j == len(state) - 1 or state[j] + jug_capacities[i] <= state[j + 1]):
                   # The action is valid, generate the new state
                   new_state = list(state)
                   new_state[j] += jug_capacities[i]
                   new_state = tuple(new_state)
                   # The cost so far is the number of actions performed, as our objective is to minimize the number of actions required to reach the goal state 
                   new_cost = g + 1
                   # If the new state is unvisited or we found a new path with a lower cost to reach this state, add it to the queue of not-yet-visited states
                   if new_state not in visited_costs or new_cost < visited_costs[new_state]:
                       visited_costs[new_state] = new_cost
                       heapq.heappush(queue, (g + heuristic(new_state, goal_state), new_cost, actions + [('+', jug_capacities[i], j + 1)], new_state))

\end{lstlisting} \\
\hline
\end{tabular}
\caption*{\footnotesize The GPT4's implementation of the A* algorithm continues on the next page.}
\end{table}

\begin{table}[H]
\centering
\begin{tabular}{|p{14cm}|}
\hline
{\fontsize{7.5pt}{8.5pt}\selectfont \textbf{Continuation of GPT4's implementation:}
\begin{lstlisting}[language=Python]
               # Similarly, check if the new state would be valid, ie if the bucket can be emptied with the jug and if the amount of water in each bucket does not exceed the amount of water in the bucket placed after it
               if state[j] - jug_capacities[i] >= 0 and (j == 0 or state[j] - jug_capacities[i] >= state[j - 1]):
                   # The action is valid, generate the new state
                   new_state = list(state)
                   new_state[j] -= jug_capacities[i]
                   new_state = tuple(new_state)
                   # The cost so far is the number of actions performed, as our objective is to minimize the number of actions required to reach the goal state 
                   new_cost = g + 1
                   # If the new state is unvisited or we found a new path with a lower cost to reach this state, add it to the queue of not-yet-visited states
                   if new_state not in visited_costs or new_cost < visited_costs[new_state]:
                       visited_costs[new_state] = new_cost
                       heapq.heappush(queue, (g + heuristic(new_state, goal_state), new_cost, actions + [('-', jug_capacities[i], j + 1)], new_state))
   return None
   
def heuristic(state, goal):
   # An admissible and consistent heuristic is the sum of the differences between the current and goal amount of water in each bucket
   # The heuristic relaxes the constraint that the amount of water in each bucket can not at any point in time exceed the amount of water in the bucket placed after it and presumes we can fill or empty the buckets directly to their goal amount
   # Thus the heuristic reports a lower estimate on the cost to reach goal state and is admissible
   # The heuristic is consistent because the cost of filling or emptying a bucket is always 1, which is exactly the decrease or increase in the difference between the current and goal amount of water in the bucket, if the bucket is filled or emptied, respectively, otherwise the estimated cost of the successor node is the same or higher, and the heuristic estimate for the goal state is 0, as the difference between the current and goal amount of water in each bucket would be 0 in the goal state.
   return sum(abs(a - b) for a, b in zip(state, goal))

print(a_star())

\end{lstlisting}} \\
\hline
\end{tabular}
\captionsetup{belowskip=0pt}
\caption{\fontsize{7.5pt}{9pt}\selectfont The A* algorithm was generated by GPT4 using the MSMT A* approach. GPT4's first A* implementation for the water\_jug problem type passed the unit tests.}
\end{table}

\begin{table}[H]
\centering
\begin{tabular}{|p{14cm}|}
\hline
\fontsize{8.2pt}{9.1pt}\selectfont \textbf{water\_jug instance (diff\_sorted\_id: 100)}: Given 9 labeled water jugs with capacities 126, 120, 127, 110, 54, 39, 38, 47, 139, 99 liters, we aim to fill 4 unlabeled buckets, numbered 1 to 4 and arranged in a line in ascending order, with 219, 240, 312, 411 liters of water respectively. The amount of water in each unlabeled bucket can not at any point in time exceed the amount of water in the bucket placed after it. Jugs can only be filled to the top and emptied completely, and the unlabeled buckets cannot be overfilled. An action, represented as a tuple ('+', X, Y) or ('-', X, Y), involves adding to or removing water from the unlabeled bucket numbered Y, using the jug with capacity X. Determine the shortest sequence of actions needed to fill the buckets as specified, and present the solution as a list of action tuples in Python syntax.\\
\hline
{\fontsize{8pt}{8.5pt}\selectfont 'initialize' function generated by GPT4 for the problem above in the second stage of the MSMT approach}\begin{lstlisting}[language=Python]
def initialize():
   # Define the initial state of the buckets as a tuple
   initial_state = (0, 0, 0, 0)
   # Define the goal state of the buckets as a tuple
   goal_state = (219, 240, 312, 411)
   # Define the capacities of the jugs as a list
   jug_capacities = [126, 120, 127, 110, 54, 39, 38, 47, 139, 99]
   # Initialize a dictionary to store the cost of reaching each visited state
   visited_costs = {}
   visited_costs[initial_state] = 0
   # Initialize a priority queue of states not yet visited, with the initial state as the first element. The priority of each element is the cost to reach that state (g) + the estimate remaining cost (h) to reach the goal
   # Record the actions required to get to each state in a list; no actions performed to reach the initial state
   queue = [(0, 0, [], initial_state)]
   return initial_state, goal_state, jug_capacities, visited_costs, queue
\end{lstlisting} \\
\hline
\end{tabular}
\captionsetup{belowskip=0pt}
\caption{\fontsize{7.5pt}{9pt}\selectfont GPT4 successfully generated a correct (but non-optimal) solution for this water\_jug problem instance using the MSMT A* and A* prompting approach. The other three baseline code and text-based prompting methods resulted in infeasible solutions.}
\end{table}

\minisection{Restricted Sorting}

\begin{table}[H]
\centering
\begin{tabular}{|p{14.5cm}|}
\hline
\footnotesize \textbf{restricted\_sorting instance (diff\_sorted\_id: 3)}: In 'Restricted Sorting', there are 6 stacks each with a capacity of 4 blocks, with 3 stacks filled with blocks of varying shades and the remaining are empty. The challenge is to sort the blocks by shade such that if a stack is not empty, it is stacked with 3 blocks of a single shade. The player can only transfer one block at a time from the top of a stack to an empty stack or to a stack that has only blocks of that shade, without exceeding the stacks’ capacity. Transferring blocks to certain stacks is more expensive than others. The cost of moving one block to the top of each stack is: {0: 3, 1: 6, 2: 5, 3: 6, 4: 1, 5: 1}, where the keys are the index of each stack, indexing from 0. The cost of moving a block is always at least 1. The solution should be a list of tuples, each containing, first, the index of the stack from which a block is picked up from and, second, the index of the stack to which it is transferred, indexing from 0. Given the initial state of the stacks, represented by the lists below (with the leftmost item being the shade of the topmost block in each stack)(and the first stack being the stack at index 0), what is the list of transfer pairs (reported in python syntax) with the least possible cost, that will result in all the blocks being correctly sorted? [[], ['Yellow', 'Green', 'Red', 'Red'], ['Blue', 'Green', 'Red', 'Green'], ['Yellow', 'Blue', 'Yellow', 'Blue'], [], []]\\
\hline
{\footnotesize A* algorithm generated by GPT4 for the problem above using A* \smodel{} approach that passed the unit tests:}\begin{lstlisting}[language=Python]
import heapq

def initialize():
   # Define the initial state of the stacks as a tuple of tuples
   initial_state = ((), ('Yellow', 'Green', 'Red', 'Red'), ('Blue', 'Green', 'Red', 'Green'), ('Yellow', 'Blue', 'Yellow', 'Blue'), (), ())
   num_stacks = 6
   stack_capacity = 4
   # Define the cost of moving a block to each stack
   stack_costs = {0: 3, 1: 6, 2: 5, 3: 6, 4: 1, 5: 1}
   visited_costs = {}
   visited_costs[initial_state] = 0
   queue = [(0, 0, [], initial_state)]
   return initial_state, num_stacks, stack_capacity, stack_costs, visited_costs, queue
  
def a_star():
   initial_state, num_stacks, stack_capacity, stack_costs, visited_costs, queue = initialize()
   while queue:
       _, g, actions, state = heapq.heappop(queue)
       # If all the stacks are empty or contain blocks of a single shade, return the actions taken
       if all(len(stack) == 0 or len(set(stack)) == 1 for stack in state):
           return actions
       # Generate all possible actions from the current state, which includes moving the top block from any stack to any other stack
       for from_stack_ind in range(num_stacks):
           # Check if the stack has any blocks to move
           if state[from_stack_ind]:
               for to_stack_ind in range(num_stacks):
                   # Check if the stack can accept more blocks and if the block can be moved to the stack
                   if (len(state[to_stack_ind]) < stack_capacity and
                       (not state[to_stack_ind] or state[to_stack_ind][-1] == state[from_stack_ind][-1])):
                       # Generate the new state
                       new_state = list(list(stack) for stack in state)
                       block = new_state[from_stack_ind].pop()
                       new_state[to_stack_ind].append(block)
                       new_state = tuple(tuple(stack) for stack in new_state)
                       # The cost of the new state is the cost of moving a block to the stack
                       new_cost = g + stack_costs[to_stack_ind]
                       if new_state not in visited_costs or new_cost < visited_costs[new_state]:
                           visited_costs[new_state] = new_cost
                           h = heuristic(new_state)
                           heapq.heappush(queue, (new_cost + h, new_cost, actions + [(from_stack_ind, to_stack_ind)], new_state))
   return None

\end{lstlisting} \\
\hline
\end{tabular}
\caption*{\footnotesize The GPT4's implementation of the A* algorithm continues on the next page.}
\end{table}

\begin{table}[H]
\centering
\begin{tabular}{|p{14cm}|}
\hline
{\footnotesize Continuation of GPT4's implementation: }\begin{lstlisting}[language=Python]
def heuristic(state):
   # The heuristic function can be the sum of the number of blocks in each stack that are not of the same shade as the top block
   # This heuristic is admissible because it presumes we can move the blocks to any stack, even if the stack is full or contains blocks of a different shade, thus reporting a lower cost estimate to reach the goal state
   # The heuristic is consistent, ie non-decreasing along the path to the goal state, because the cost of moving a block from one stack to another is always at least 1, which is exactly the decrease in the heuristic value, if the block is moved to a stack with blocks of the same shade, otherwise the heuristic value of the successor node is the same or higher, and the heuristic estimate for the goal state is 0, as all the blocks in each stack are of the same shade in the goal state
   h = 0
   for stack in state:
       if stack:
           h += len([block for block in stack if block != stack[-1]])
   return h

print(a_star())
\end{lstlisting} \\
\hline
\end{tabular}
\captionsetup{belowskip=0pt}
\caption{\footnotesize The A* algorithm was generated by GPT4 using the MSMT A* approach. GPT4's third A* implementation for the restricted\_sorting problem type passed the unit tests.}
\end{table}

\begin{table}[H]
\centering
\begin{tabular}{|p{14cm}|}
\hline
\footnotesize \textbf{restricted\_sorting instance (diff\_sorted\_id: 57)}: In 'Restricted Sorting', there are 8 stacks each with a capacity of 5 blocks, with 4 stacks filled with blocks of varying shades and the remaining are empty. The challenge is to sort the blocks by shade such that if a stack is not empty, it is stacked with 4 blocks of a single shade. The player can only transfer one block at a time from the top of a stack to an empty stack or to a stack that has only blocks of that shade, without exceeding the stacks’ capacity. Transferring blocks to certain stacks is more expensive than others. The cost of moving one block to the top of each stack is: {0: 2, 1: 3, 2: 9, 3: 8, 4: 3, 5: 7, 6: 9, 7: 5}, where the keys are the index of each stack, indexing from 0. The cost of moving a block is always at least 1. The solution should be a list of tuples, each containing, first, the index of the stack from which a block is picked up from and, second, the index of the stack to which it is transferred, indexing from 0. Given the initial state of the stacks, represented by the lists below (with the leftmost item being the shade of the topmost block in each stack)(and the first stack being the stack at index 0), what is the list of transfer pairs (reported in python syntax) with the least possible cost, that will result in all the blocks being correctly sorted? [[], [], ['Green', 'Red', 'Black', 'Blue', 'Blue'], ['Yellow', 'Blue', 'Blue', 'Green', 'Black'], ['Yellow', 'Yellow', 'Yellow', 'Red', 'Black'], [], [], ['Red', 'Green', 'Green', 'Red', 'Black']]\\
\hline
{\footnotesize 'initialize' function generated by GPT4 for the problem above in the second stage of the MSMT approach}\begin{lstlisting}[language=Python]
def initialize():
   # Define the initial state of the stacks as a tuple of tuples
   initial_state = ((), (), ('Green', 'Red', 'Black', 'Blue', 'Blue'), 
                    ('Yellow', 'Blue', 'Blue', 'Green', 'Black'), 
                    ('Yellow', 'Yellow', 'Yellow', 'Red', 'Black'), (), (), 
                    ('Red', 'Green', 'Green', 'Red', 'Black'))
   num_stacks = 8
   stack_capacity = 5
   # Define the cost of moving a block to each stack
   stack_costs = {0: 2, 1: 3, 2: 9, 3: 8, 4: 3, 5: 7, 6: 9, 7: 5}
   visited_costs = {}
   visited_costs[initial_state] = 0
   queue = [(0, 0, [], initial_state)]
   return initial_state, num_stacks, stack_capacity, stack_costs, visited_costs, queue
\end{lstlisting} \\
\hline
\end{tabular}
\captionsetup{belowskip=0pt}
\caption{\footnotesize GPT4 successfully generated a feasible (but not correct) solution for this restricted\_sorting problem instance using the MSMT A* and A* prompting approach. The other three baseline code and text-based prompting methods resulted in infeasible solutions.}
\end{table}

\minisection{Color Sorting}

\begin{table}[H]
\centering
\begin{tabular}{|p{14.5cm}|}
\hline
\footnotesize \textbf{color\_sorting instance (diff\_sorted\_id: 1)}: The game of 'Sort It' begins with 3 tubes, each filled with 4 balls of different colors. The goal is to sort the balls by color, with each tube containing balls of only one color. Only one ball can be moved at a time, taken from the top of one tube and placed on top of another. The capacity of each tube (maximum number of balls we can fit in each tube) is 6 balls. It is not allowed to place a ball in a tube that already has 6 balls. The solution should be a list of tuples, each containing, first, the index of the tube from which a ball is taken and, second, the index of the tube to which it is moved, indexing from 0. Given the initial state of the tubes, represented by the lists below (with the leftmost item being the color of the topmost ball in each tube), what is the shortest list of move tuples that will result in all the balls being correctly sorted? [['Green', 'Red', 'Green', 'Red'], ['Blue', 'Blue', 'Red', 'Green'], ['Red', 'Blue', 'Green', 'Blue']]\\
\hline
{\footnotesize A* algorithm generated by GPT4 for the problem above using A* \smodel{} approach that passed the unit tests:}\begin{lstlisting}[language=Python]
import heapq

def initialize():
   # Define the initial state and the goal state of the tubes, represented as 3x4 tuples
   initial_state = (('Green', 'Red', 'Green', 'Red'), ('Blue', 'Blue', 'Red', 'Green'), ('Red', 'Blue', 'Green', 'Blue'))
   goal_state = (('Red', 'Red', 'Red', 'Red'), ('Green', 'Green', 'Green', 'Green'), ('Blue', 'Blue', 'Blue', 'Blue'))   
   # Encoding other variables given in the problem statement
   num_tubes = 3
   tube_capacity = 6
   # Initialize a dictionary to store the cost of reaching each visited state
   visited_costs = {}
   visited_costs[initial_state] = 0
   # Initialize a priority queue of states not yet visited, with the initial state as the first element. The priority of each element is the cost to reach that state (g) + the estimate remaining cost (h) to reach the goal
   # Record the moves required to get to each state in a list; no moves performed to reach the initial state
   queue = [(0, 0, [], initial_state)]
   return initial_state, goal_state, num_tubes, tube_capacity, visited_costs, queue
  
def a_star():
   # The initialize function initializes and returns the visited_costs dictionary and the priority queue and encodes all of the variables given in the problem (ie the initial and goal state of the tubes and the tube capacity)
   initial_state, goal_state, num_tubes, tube_capacity, visited_costs, queue = initialize()
   # While there are un-visited states
   while queue:
       # Pop the state with the lowest sum of the cost so far and estimated cost to the goal from the queue
       _, g, actions, state = heapq.heappop(queue)
       # We can check if the current state is the goal state with a simple equality check, as the goal state is predefined
       if state == goal_state:
           return actions
       # Generate all valid actions from the current state, which includes moving the top ball from any tube to the top of any other tube
       for from_tube in range(num_tubes):
           for to_tube in range(num_tubes):
               # Check if the move is valid, ie if the from_tube is not empty and the to_tube is not full
               if state[from_tube] and len(state[to_tube]) < tube_capacity:
                   # The action is valid, generate the new state
                   new_state = [list(tube[:]) for tube in state]
                   ball_to_be_moved = new_state[from_tube].pop(0)
                   # Do the move
                   new_state[to_tube].insert(0, ball_to_be_moved)
                   new_state = tuple(tuple(tube) for tube in new_state)
                   # The cost so far is the number of moves made, as our objective is to minimize the number of moves required to reach the goal state 
                   new_cost = g + 1          
                   # If the new state is unvisited or we found a new path with a lower cost to reach this state, add it to the queue of not-yet-visited states
                   if new_state not in visited_costs or new_cost < visited_costs[new_state]:
                       visited_costs[new_state] = new_cost
                       heapq.heappush(queue, (g + heuristic(new_state, goal_state), new_cost, actions + [(from_tube, to_tube)], new_state))               
   return None

\end{lstlisting} \\
\hline
\end{tabular}
\caption*{\footnotesize The GPT4's implementation of the A* algorithm continues on the next page.}
\end{table}

\begin{table}[H]
\centering
\begin{tabular}{|p{14cm}|}
\hline
{\footnotesize Continuation of GPT4's implementation: }\begin{lstlisting}[language=Python]
def heuristic(state, goal):
   # An admissible and consistent heuristic is the sum of the number of misplaced balls in each tube
   # The heuristic relaxes the constraint that a ball can only be moved from the top of a tube and presumes we can move the balls to their goal position by taking them from any position in the tube
   # Thus the heuristic reports a lower estimate on the cost to reach goal state and is admissible
   # The heuristic is consistent because the cost of moving a ball to a neighboring tube is always 1, which is exactly the decrease in the number of misplaced balls, if the ball is moved toward its goal position, otherwise the estimated cost of the successor node is the same or higher, and the heuristic estimate for the goal state is 0, as the number of misplaced balls would be 0 in the goal state.
   h = 0
   for i in range(len(state)):
       for j in range(len(state[i])):
           # Can't compare integers with "_" when finding the goal position of each ball, thus ignore the "_" ball
           if state[i][j] != '_':
               # Add the the number of misplaced balls in each tube to the estimate
               h += sum(1 for ball in state[i] if ball != goal[i][0])
   return h

print(a_star())
\end{lstlisting} \\
\hline
\end{tabular}
\caption{\footnotesize The A* algorithm was generated by GPT4 using the MSMT A* approach. GPT4's A* implementation for the first instance of the color\_sorting problem type passed the unit tests.}
\end{table}

\begin{table}[H]
\centering
\begin{tabular}{|p{14cm}|}
\hline
\footnotesize \textbf{color\_sorting instance (diff\_sorted\_id: 97)}: The game of 'Sort It' begins with 3 tubes, each filled with 6 balls of different colors. The goal is to sort the balls by color, with each tube containing balls of only one color. Only one ball can be moved at a time, taken from the top of one tube and placed on top of another. The capacity of each tube (maximum number of balls we can fit in each tube) is 9 balls. It is not allowed to place a ball in a tube that already has 9 balls. The solution should be a list of tuples, each containing, first, the index of the tube from which a ball is taken and, second, the index of the tube to which it is moved, indexing from 0. Given the initial state of the tubes, represented by the lists below (with the leftmost item being the color of the topmost ball in each tube), what is the shortest list of move tuples that will result in all the balls being correctly sorted? [['Red', 'Red', 'Blue', 'Red', 'Blue', 'Green'], ['Red', 'Green', 'Blue', 'Blue', 'Green', 'Red'], ['Blue', 'Green', 'Green', 'Red', 'Blue', 'Green']]\\
\hline
{\footnotesize 'initialize' function generated by GPT4 for the problem above in the second stage of the MSMT approach}\begin{lstlisting}[language=Python]
def initialize():
   # Define the initial state and the goal state of the tubes, represented as 3x6 tuples
   initial_state = (('Red', 'Red', 'Blue', 'Red', 'Blue', 'Green'), 
                    ('Red', 'Green', 'Blue', 'Blue', 'Green', 'Red'), 
                    ('Blue', 'Green', 'Green', 'Red', 'Blue', 'Green'))
   goal_state = (('Red', 'Red', 'Red', 'Red', 'Red', 'Red'), 
                 ('Green', 'Green', 'Green', 'Green', 'Green', 'Green'), 
                 ('Blue', 'Blue', 'Blue', 'Blue', 'Blue', 'Blue'))    
   # Encoding other variables given in the problem statement
   num_tubes = 3
   tube_capacity = 9
   # Initialize a dictionary to store the cost of reaching each visited state
   visited_costs = {}
   visited_costs[initial_state] = 0
   # Initialize a priority queue of states not yet visited, with the initial state as the first element. The priority of each element is the cost to reach that state (g) + the estimate remaining cost (h) to reach the goal
   # Record the moves required to get to each state in a list; no moves performed to reach the initial state
   queue = [(0, 0, [], initial_state)]
   return initial_state, goal_state, num_tubes, tube_capacity, visited_costs, queue
\end{lstlisting} \\
\hline
\end{tabular}
\captionsetup{belowskip=0pt}
\caption{\footnotesize GPT4 successfully generated a correct (but non-optimal) solution for this color\_sorting problem instance using MSMT A* approach. Prompting GPT4 with all other four code and text-based prompting methods resulted in infeasible solutions.}
\end{table}

\minisection{Magic Square}

\begin{table}[H]
\centering
\begin{tabular}{|p{14.5cm}|}
\hline
\footnotesize \textbf{magic\_square instance (diff\_sorted\_id: 1)}: The game of 'Sort It' begins with 3 tubes, each filled with 4 balls of different colors. The goal is to sort the balls by color, with each tube containing balls of only one color. Only one ball can be moved at a time, taken from the top of one tube and placed on top of another. The capacity of each tube (maximum number of balls we can fit in each tube) is 6 balls. It is not allowed to place a ball in a tube that already has 6 balls. The solution should be a list of tuples, each containing, first, the index of the tube from which a ball is taken and, second, the index of the tube to which it is moved, indexing from 0. Given the initial state of the tubes, represented by the lists below (with the leftmost item being the color of the topmost ball in each tube), what is the shortest list of move tuples that will result in all the balls being correctly sorted? [['Green', 'Red', 'Green', 'Red'], ['Blue', 'Blue', 'Red', 'Green'], ['Red', 'Blue', 'Green', 'Blue']]\\
\hline
{\footnotesize A* algorithm generated by GPT4 for the problem above using A* \smodel{} approach that passed the unit tests:}\begin{lstlisting}[language=Python]
import heapq

def initialize():
    # Define the initial state of the grid, the goal sums of the specified row, column, and diagonal, and the range of unique integers that can be used to replace 'x's
    initial_state = (('35', 'x', '46'), ('x', 'x', 'x'), ('x', 'x', 'x'))
    goal_sums = {'row1': 103, 'col1': 107, 'diag': 124}
    num_range = set(range(29, 49))
    # Initialize a dictionary to store the cost of reaching each visited state
    visited_costs = {}
    visited_costs[initial_state] = 0
    # Initialize a priority queue of states not yet visited, with the initial state as the first element. The priority of each element is the cost to reach that state (g) + the estimate remaining cost (h) to reach the goal
    # Record the actions required to get to each state in a list; no actions performed to reach the initial state
    queue = [(0, 0, [], initial_state)]
    return initial_state, goal_sums, num_range, visited_costs, queue

def a_star():
    # The initialize function initializes and returns the visited_costs dictionary and the priority queue and encodes all of the variables given in the problem (ie the initial state of the grid, the goal sums, and the range of unique integers)
    initial_state, goal_sums, num_range, visited_costs, queue = initialize()
    # While there are un-visited states
    while queue:
        # Pop the state with the lowest sum of the cost so far and estimated cost to the goal from the queue
        _, g, actions, state = heapq.heappop(queue)
        # Check if the current state is the goal state
        # The goal state is where the sum of the specified row, column, and diagonal equal the goal sums, and there are no 'x's in the grid
        if all(cell != 'x' for row in state for cell in row) and sum(int(cell) for cell in state[1]) == goal_sums['row1'] and sum(int(state[i][1]) for i in range(3)) == goal_sums['col1'] and sum(int(state[i][2-i]) for i in range(3)) == goal_sums['diag']:
            return actions
        # If the state has at least 1 remaining unknown number, ie 'x', generate all possible actions from the current state, which includes replacing the next x in the grid with any of avaiable unique integers (all numbers in the range - numbers present in the current state) in the range 
        if any(cell == 'x' for row in state for cell in row):
            # Find the next 'x' in the grid
            for i in range(3):
                for j in range(3):
                    if state[i][j] == 'x':
                        # Generate the set of available numbers by subtracting the set of numbers present in the current state from the set of all numbers in the range
                        available_nums = num_range - set(int(cell) for row in state for cell in row if cell != 'x')
                        for num in available_nums:
                            # Generate the new state
                            new_state = [list(row) for row in state]
                            new_state[i][j] = str(num)
                            new_state = tuple(tuple(row) for row in new_state)
                            # The cost so far is the sum of the numbers in the grid, as the task is to minimize the sum of all of the numbers in the grid
                            new_cost = g + num
                            # If the new state is unvisited or we found a new path with a lower cost to arrive at this state, add it to the queue of un-visited states
\end{lstlisting} \\
\hline
\end{tabular}
\caption*{\footnotesize The GPT4's implementation of the A* algorithm continues on the next page.}
\end{table}

\begin{table}[H]
\centering
\begin{tabular}{|p{14cm}|}
\hline
{\fontsize{7.5pt}{8pt}\selectfont \textbf{Continuation of GPT4's implementation: }
\begin{lstlisting}[language=Python]
                            if new_state not in visited_costs or new_cost < visited_costs[new_state]:
                                visited_costs[new_state] = new_cost
                                heapq.heappush(queue, (new_cost + heuristic(new_state, goal_sums), new_cost, actions + [(i, j, num)], new_state))
                        # Stop searching for 'x's after the first 'x' is found and all possible actions from it are generated
                        break
                else:
                    continue
                break
    return None
    
def heuristic(state, goal_sums):
    # An admissible and consistent heuristic for this problem is the sum of the absolute differences between the current sums and the goal sums of the specified row, column, and diagonal
    # The heuristic relaxes the constraints that each number in the grid must be unique and that the sum of all of the numbers in the grid must be minimized; ie It presumes we can add or subtract any amount to the current sums to reach the goal sums
    # Thus the heuristic reports a lower estimate on the cost of reaching the goal state and is admissible
    # The heuristic is consistent because the estimated cost from the current state to the goal can never be greater than the sum of cost from the current node to a successor node plus the estimated cost from the successor node to the goal because the cost of reaching a successor node is the number added to the grid, which is always greater than or equal to 1, the decrease in the absolute difference between the current and goal sums
    h = 0
    if any(cell == 'x' for row in state for cell in row):
        h += abs(sum(int(cell) for cell in state[1] if cell != 'x') - goal_sums['row1'])
        h += abs(sum(int(state[i][1]) for i in range(3) if state[i][1] != 'x') - goal_sums['col1'])
        h += abs(sum(int(state[i][2-i]) for i in range(3) if state[i][2-i] != 'x') - goal_sums['diag'])
    return h

print(a_star())
\end{lstlisting}} \\
\hline
\end{tabular}
\captionsetup{belowskip=0pt}
\caption{\fontsize{8pt}{8pt}\selectfont The A* algorithm was generated by GPT4 using the MSMT A* approach. GPT4's first attempt at implementing the A* algorithm for the first instance of the magic\_square problem type passed the unit tests.}
\end{table}

\begin{table}[H]
\centering
\begin{tabular}{|p{14.5cm}|}
\hline
\fontsize{8pt}{8.5pt}\selectfont \textbf{magic\_square instance (diff\_sorted\_id: 34)}: In the magic square problem, a 3x3 grid is filled with unique integers ranging from 40 to 89. Some numbers are already given, while others are unknown and represented as 'x'. Sum of column 1 (counting from 0) must be 183, and sum of row 1 must be 194. Also, the sum of the numbers in the diagonal from the top right to the bottom left corner of the grid should equal 161. The goal is to find unique integers (ie each number can be in the final grid only once) in the given range to replace with ‘x’s in the grid below such that the sum of the specified rows, columns, and diagonal equals the given amounts and the sum of all of the numbers in the grid is as low as possible. The solution should be provided as a list of tuples in Python syntax. Each tuple should contain three numbers for each 'x' position: the row index, the column index (both starting from 0), and the value of the unique integer replaced with 'x' at that position.\textbackslash n\textbackslash nGrid:\textbackslash n [['x' 'x' 'x']\textbackslash n ['x' 'x' 'x']\textbackslash n ['40' 'x' '71']]\\
\hline
{\fontsize{8pt}{8.5pt}\selectfont 'initialize' function generated by GPT4 for the problem above in the second stage of the MSMT approach}\begin{lstlisting}[language=Python]
def initialize():
    # Define the initial state of the grid, the goal sums of the specified row, column, and diagonal, and the range of unique integers that can be used to replace 'x's
    initial_state = (('x', 'x', 'x'), ('x', 'x', 'x'), ('40', 'x', '71'))
    goal_sums = {'row1': 194, 'col1': 183, 'diag': 161}
    num_range = set(range(40, 90))
    # Initialize a dictionary to store the cost of reaching each visited state
    visited_costs = {}
    visited_costs[initial_state] = 0
    # Initialize a priority queue of states not yet visited, with the initial state as the first element. The priority of each element is the cost to reach that state (g) + the estimate remaining cost (h) to reach the goal
    # Record the actions required to get to each state in a list; no actions performed to reach the initial state
    queue = [(0, 0, [], initial_state)]
    return initial_state, goal_sums, num_range, visited_costs, queue
\end{lstlisting} \\
\hline
\end{tabular}
\captionsetup{belowskip=0pt}
\caption{\fontsize{7.pt}{8pt}\selectfont GPT4 successfully generated a correct (but non-optimal) solution for this magic\_square problem instance using MSMT A* approach. Prompting GPT4 with all other four code and text-based prompting methods resulted in infeasible solutions.}
\end{table}

\minisection{Consecutive Grid}

\begin{table}[H]
\centering
\begin{tabular}{|p{14.8cm}|}
\hline
\fontsize{8.4pt}{8.5pt}\selectfont \textbf{consecutive\_grid instance (diff\_sorted\_id: 13)}: We have a 3x3 numerical grid, with numbers ranging from 21 to 59 (21 included in the range but 59 is not included). The numbers in each row and column must be strictly increasing or decreasing. This means that either first > second > third or first < second < third in each row and column. If a grid cell is marked with an 'x', the number in that position is hidden. The objective is to replace the 'x's with unique integers from the given range, ensuring that each number only appears once in the grid. The replacements must maintain the consecutive order in each row and column. Additionally, the sum of the numbers in the topmost row plus the numbers in the rightmost column plus the numbers in the diagonal connecting the top-left corner of the grid to its bottom-right corner should be minimized. The solution should be given as a list of tuples in Python syntax. Each tuple should represent the replacement of a number with an  'x' number and contain three elements: the row index of the 'x', the column index  of the 'x' (both starting from 0), and the value of the number that replaces the 'x'. The initial state of the grid is as follows: \textbackslash n\textbackslash nGrid:\textbackslash n [['29' '45' 'x']\textbackslash n ['x' 'x' 'x']\textbackslash n ['x' '43' '34']]\\
\hline
{\fontsize{7.5pt}{9pt}\selectfont A* algorithm generated by GPT4 for the problem above using A* \smodel{} approach that passed the unit tests:}\begin{lstlisting}[language=Python]
import heapq
import numpy as np

def initialize():
    # Define the initial state of the grid, as a 2d tuple
    initial_state = (('29', '45', 'x'), ('x', 'x', 'x'), ('x', '43', '34'))
    # Define the range of numbers that can be used to replace 'x'
    num_range = set(range(21, 59))
    # Define the dimensions of the grid
    num_rows = 3
    num_cols = 3
    # Initialize a dictionary to store the cost of reaching each visited state
    visited_costs = {}
    visited_costs[initial_state] = 0
    # Initialize a priority queue of states not yet visited, with the initial state as the first element. The priority of each element is the cost to reach that state (g) + the estimate remaining cost (h) to reach the goal
    # Record the actions required to get to each state in a list; no actions performed to reach the initial state
    queue = [(0, 0, [], initial_state)]
    return initial_state, num_range, num_rows, num_cols, visited_costs, queue

def a_star():
    # The initialize function initializes and returns the visited_costs dictionary and the priority queue and encodes all of the variables given in the problem (ie the initial state of the grid, the range of numbers, and the dimensions of the grid)
    initial_state, num_range, num_rows, num_cols, visited_costs, queue = initialize()
    # While there are un-visited states
    while queue:
        # Pop the state with the lowest sum of the cost so far and estimated cost to the goal from the queue
        _, g, actions, state = heapq.heappop(queue)
        # If the current state does not have any 'x's, ie all the numbers have been filled in, and the numbers in each row and column are consecutive, return the actions taken to reach this state
        if all(all(cell != 'x' for cell in row) for row in state) and all(all(int(row[i]) < int(row[i + 1]) for i in range(len(row) - 1)) or all(int(row[i]) > int(row[i + 1]) for i in range(len(row) - 1)) for row in state) and all(all(int(state[i][col]) < int(state[i + 1][col]) for i in range(len(state) - 1)) or all(int(state[i][col]) > int(state[i + 1][col]) for i in range(len(state) - 1)) for col in range(len(state[0]))):
            return actions
        # If the state has at least 1 remaining unknown number, ie 'x', generate all possible actions from the current state, which includes replacing the next x in the grid with any of avaiable unique integers (all numbers in the range - numbers present in the current state) in the range 
        if any('x' in row for row in state):
            # Find the next 'x' in the grid
            for row_ind in range(num_rows):
                for col_ind in range(num_cols):
                    if state[row_ind][col_ind] == 'x':
                        # Generate all possible actions from the current state, which includes replacing the 'x' with any of the available unique integers in the range
                        for num in num_range - set(cell for row in state for cell in row if cell != 'x'):
                            # Generate the new state
                            new_state = [list(row[:]) for row in state]
                            new_state[row_ind][col_ind] = num
                            new_state = tuple(tuple(row) for row in new_state)
                            # The cost so far is the sum of the numbers in the topmost row, the rightmost column, and the diagonal from the top-left corner to the bottom-right corner of the grid
\end{lstlisting} \\
\hline
\end{tabular}
\caption*{\fontsize{7.5pt}{9pt}\selectfont The GPT4's implementation of the A* algorithm continues on the next page.}
\end{table}

\begin{table}[H]
\centering
\begin{tabular}{|p{14cm}|}
\hline
{\fontsize{7.5pt}{8.5pt}\selectfont \textbf{Continuation of GPT4's implementation: }\begin{lstlisting}[language=Python]
                            new_cost = sum(int(cell) for cell in new_state[0] if cell != 'x') + sum(int(new_state[i][-1]) for i in range(num_rows) if new_state[i][-1] != 'x') + sum(int(new_state[i][i]) for i in range(num_rows) if new_state[i][i] != 'x')
                            # If the new state is unvisited or we found a new path with a lower cost to reach this state, add it to the queue of not-yet-visited states
                            if new_state not in visited_costs or new_cost < visited_costs[new_state]:
                                visited_costs[new_state] = new_cost
                                heapq.heappush(queue, (new_cost + heuristic(new_state), new_cost, actions + [(row_ind, col_ind, num)], new_state))
                        # Once we have generated all possible actions from the current state, break the loop
                        break
                else:
                    continue
                break
    return None

def heuristic(state):
    # An admissible and consistent heuristic for this problem is the sum of the minimum remaining numbers that can be used to replace the 'x's in the grid
    # This heuristic is admissible because it never overestimates the cost to reach the goal, as the minimum remaining numbers are always less than or equal to the actual numbers that will replace the 'x's
    # The heuristic is consistent because the estimated cost from the current state to the goal can never be greater than the sum of cost from the current node to a successor node plus the estimated cost from the successor node to the goal because the cost of replacing an 'x' with a number is the number itself, which is always greater than or equal to the minimum remaining number
    return sum(sorted(set(range(21, 59)) - set(int(cell) for row in state for cell in row if cell != 'x'))[:sum(cell == 'x' for row in state for cell in row)])

print(a_star())
\end{lstlisting}} \\
\hline
\end{tabular}
\caption{\fontsize{8pt}{8pt}\selectfont The A* algorithm was generated by GPT4 using the MSMT A* approach. GPT4's 13-th attempt at implementing the A* algorithm for consecutive\_grid problem type passed the unit tests.}
\end{table}

\begin{table}[H]
\centering
\begin{tabular}{|p{14.5cm}|}
\hline
\fontsize{8pt}{8.5pt}\selectfont \textbf{consecutive\_grid instance (diff\_sorted\_id: 43)}: We have a 3x3 numerical grid, with numbers ranging from 7 to 60 (7 included in the range but 60 is not included). The numbers in each row and column must be strictly increasing or decreasing. This means that either first > second > third or first < second < third in each row and column. If a grid cell is marked with an 'x', the number in that position is hidden. The objective is to replace the 'x's with unique integers from the given range, ensuring that each number only appears once in the grid. The replacements must maintain the consecutive order in each row and column. Additionally, the sum of the numbers in the topmost row plus the numbers in the rightmost column plus the numbers in the diagonal connecting the top-left corner of the grid to its bottom-right corner should be minimized. The solution should be given as a list of tuples in Python syntax. Each tuple should represent the replacement of a number with an  'x' number and contain three elements: the row index of the 'x', the column index  of the 'x' (both starting from 0), and the value of the number that replaces the 'x'. The initial state of the grid is as follows: \textbackslash n\textbackslash nGrid:\textbackslash n [['13' 'x' 'x']\textbackslash n ['x' 'x' '27']\textbackslash n ['x' '26' 'x']]\\
\hline
{\footnotesize 'initialize' function generated by GPT4 for the problem above in the second stage of the MSMT approach}\begin{lstlisting}[language=Python]
def initialize():
    # Define the initial state of the grid, as a 2d tuple
    initial_state = (('13', 'x', 'x'), ('x', 'x', '27'), ('x', '26', 'x'))
    # Define the range of numbers that can be used to replace 'x'
    num_range = set(range(7, 60))
    # Define the dimensions of the grid
    num_rows = 3
    num_cols = 3
    # Initialize a dictionary to store the cost of reaching each visited state
    visited_costs = {}
    visited_costs[initial_state] = 0
    # Initialize a priority queue of states not yet visited, with the initial state as the first element. The priority of each element is the cost to reach that state (g) + the estimate remaining cost (h) to reach the goal
    # Record the actions required to get to each state in a list; no actions performed to reach the initial state
    queue = [(0, 0, [], initial_state)]
    return initial_state, num_range, num_rows, num_cols, visited_costs, queue
\end{lstlisting} \\
\hline
\end{tabular}
\captionsetup{belowskip=0pt}
\caption{\fontsize{8pt}{8pt}\selectfont GPT4 successfully generated an optimal and correct solution for this consecutive\_grid problem instance using MSMT A* approach. Prompting GPT4 with all other four code and text-based prompting methods resulted in infeasible solutions.}
\end{table}

\minisection{Traffic}

\begin{table}[H]
\centering
\begin{tabular}{|p{14.5cm}|}
\hline
\fontsize{8.5pt}{9pt}\selectfont \textbf{traffic instance (diff\_sorted\_id: 1)}: Using the provided matrix map of a city, where numbers represent travel time in minutes (all numbers are positive integers) and 'x' marks closed workshops, find the quickest route for Ben to travel from his current workshop at index (5, 7) to his destination workshop at index (2, 1), indexing from 0. Ben's car can move north, south, east, or west from a given crossroad, provided there's no x in that direction. Also, there are 3 districts in the city with district 1 covering rows 0 to 2, district 2 covering rows 3 to 4, and district 3 covering rows 5 to 8. Ben has to visit at least 1 workshop in each district on his path to the destination. The roads are bidirectional. The answer should be a list of tuples (in Python syntax) indicating the index of workshops on Ben's path. The start and end workshops must be included in the path.\textbackslash n[13 x x 3 10 20 3 x 5]\textbackslash n[x x x x 20 16 x x 17]\textbackslash n[3 14 18 8 1 20 14 x 7]\textbackslash n[13 3 6 10 7 4 6 6 1]\textbackslash n[10 12 2 x 11 x 10 8 11]\textbackslash n[x x x 11 6 18 13 20 17]\textbackslash n[x x 16 x 4 17 7 10 15]\textbackslash n[x 7 16 6 19 4 7 x x]\textbackslash n[x 11 18 x x x 3 8 x]\\
\hline
{\fontsize{8pt}{8.5pt}\selectfont A* algorithm generated by GPT4 for the problem above using A* \smodel{} approach that passed the unit tests:}\begin{lstlisting}[language=Python]
import heapq

def initialize():
   # Define the initial state and the goal state of the problem, represented as tuples
   initial_state = (5, 7)
   goal_state = (2, 1)
   # Encoding other variables given in the problem statement
   num_rows = 9
   num_cols = 9
   matrix = [[13, 'x', 'x', 3, 10, 20, 3, 'x', 5],['x', 'x', 'x', 'x', 20, 16, 'x', 'x', 17],[3, 14, 18, 8, 1, 20, 14, 'x', 7],[13, 3, 6, 10, 7, 4, 6, 6, 1],[10, 12, 2, 'x', 11, 'x', 10, 8, 11],['x', 'x', 'x', 11, 6, 18, 13, 20, 17],['x', 'x', 16, 'x', 4, 17, 7, 10, 15],['x', 7, 16, 6, 19, 4, 7, 'x', 'x'], ['x', 11, 18, 'x', 'x', 'x', 3, 8, 'x']]
   districts = {1: range(0, 3), 2: range(3, 5), 3: range(5, 9)}
   # Initialize a dictionary to store the cost of reaching each visited state
   visited_costs = {}
   visited_costs[initial_state] = 0
   # Initialize a priority queue of states not yet visited, with the initial state as the first element. The priority of each element is the cost to reach that state (g) + the estimate remaining cost (h) to reach the goal
   # Record the actions required to get to each state in a list; no actions performed to reach the initial state
   queue = [(0, 0, [], initial_state, {1: False, 2: False, 3: True})]
   return initial_state, goal_state, num_rows, num_cols, matrix, districts, visited_costs, queue
  
def a_star():
   initial_state, goal_state, num_rows, num_cols, matrix, districts, visited_costs, queue = initialize()
   while queue:
       _, g, actions, state, visited_districts = heapq.heappop(queue)
       # Check if the current state is the goal state
       if state == goal_state and all(visited_districts.values()):
           return actions
       # Generate all possible actions from the current state, which includes moving to any of the 4 neighboring cells
       for d_row, d_col in [(0, -1), (0, 1), (1, 0), (-1, 0)]:
           new_row, new_col = state[0] + d_row, state[1] + d_col
           # Check if the new state would be valid, ie if the coordinate of the cell to move to is a valid coordinate within the bounds of the matrix and the cell is not 'x'
           if 0 <= new_row < num_rows and 0 <= new_col < num_cols and matrix[new_row][new_col] != 'x':
               # Generate the new state
               new_state = (new_row, new_col)
               # The cost so far is the sum of the travel times, as our objective is to minimize the total travel time 
               new_cost = g + matrix[new_row][new_col]
               # Update the visited districts
               new_visited_districts = visited_districts.copy()
               for district, rows in districts.items():
                   if new_row in rows:
                       new_visited_districts[district] = True
               # If the new state is unvisited or we found a new path with a lower cost to reach this state, add it to the queue of not-yet-visited states
               if new_state not in visited_costs or new_cost < visited_costs[new_state]:
                   visited_costs[new_state] = new_cost
                   heapq.heappush(queue, (g + heuristic(new_state, goal_state), new_cost, actions + [new_state], new_state, new_visited_districts))
   return None
\end{lstlisting} \\
\hline
\end{tabular}
\caption*{\fontsize{8pt}{8.5pt}\selectfont The GPT4's implementation of the A* algorithm continues on the next page.}
\end{table}

\begin{table}[H]
\centering
\begin{tabular}{|p{14cm}|}
\hline
{\fontsize{8pt}{8.5pt}\selectfont \textbf{Continuation of GPT4's implementation: }\begin{lstlisting}[language=Python]
def heuristic(state, goal):
   # An admissible and consistent heuristic is the Manhattan distance (the shortest path) of the current position from the goal position
   # The heuristic relaxes the constraint that a cell can only be moved to if it's not 'x' and presumes we can move to the goal position by moving to any of the neighboring cells
   # Thus the heuristic reports a lower estimate on the cost to reach goal state and is admissible
   # The heuristic is consistent because the cost of moving to a neighboring coordinate is always greater than or equal to 1, which is exactly the decrease in the Manhattan distance, if the cell is moved toward its goal position, otherwise the estimated cost of the successor node is the same or higher, and he heuristic estimate for the goal state is 0, as the distance of the goal position from itself would be 0.
   h = abs(state[0] - goal[0]) + abs(state[1] - goal[1])
   return h
   
print(a_star())
\end{lstlisting}} \\
\hline
\end{tabular}
\caption{\fontsize{8pt}{8.5pt}\selectfont The A* algorithm was generated by GPT4 using the MSMT A* approach. GPT4's first implementation of the A* algorithm for traffic problem type passed the unit tests.}
\end{table}

\begin{table}[H]
\centering
\begin{tabular}{|p{14.5cm}|}
\hline
\fontsize{8.5pt}{9pt}\selectfont \textbf{traffic instance (diff\_sorted\_id: 104)}: Using the provided matrix map of a city, where numbers represent travel time in minutes (all numbers are positive integers) and 'x' marks closed workshops, find the quickest route for Ben to travel from his current workshop at index (7, 2) to his destination workshop at index (4, 14), indexing from 0. Ben's car can move north, south, east, or west from a given crossroad, provided there's no x in that direction. Also, there are 3 districts in the city with district 1 covering rows 0 to 4, district 2 covering rows 5 to 8, and district 3 covering rows 9 to 14. Ben has to visit at least 1 workshop in each district on his path to the destination. The roads are bidirectional. The answer should be a list of tuples (in Python syntax) indicating the index of workshops on Ben's path. The start and end workshops must be included in the path.\textbackslash n[x x 7 18 11 4 9 x 15 x x x 1 x x]\textbackslash n[x x 14 x 10 x x x 8 x x 16 1 x x]\textbackslash n[17 17 x x x x x 7 6 12 x x x 7 x]\textbackslash n[x x x x 10 x x 18 1 11 4 x 13 10 16]\textbackslash n[x x x 12 1 x x x 2 7 1 x 2 x 5]\textbackslash n[x x 8 x 12 x x 5 17 x 2 x 4 10 18]\textbackslash n[x x 4 19 1 x 18 7 x 10 3 x 19 16 19]\textbackslash n[6 12 15 16 5 9 16 18 10 15 5 x 5 14 x]\textbackslash n[x 16 8 17 12 11 16 8 9 9 7 4 5 20 3]\textbackslash n[19 17 15 x x 17 x 3 2 2 11 7 8 16 1]\textbackslash n[13 4 17 x x x x 5 x 18 16 15 19 4 x]\textbackslash n[11 x x x x 12 x x x x x x 19 x x]\textbackslash n[18 x x x 9 19 x 16 6 x 9 3 16 15 x]\textbackslash n[x x x 10 x x 4 3 x 3 x 16 18 x 12]\textbackslash n[x 13 x x x x x 10 5 x 17 x x 7 x]\\
\hline
{\fontsize{8pt}{8.5pt}\selectfont 'initialize' function generated by GPT4 for the problem above in the second stage of the MSMT approach}\begin{lstlisting}[language=Python]
def initialize():
   # Define the initial state and the goal state of the problem, represented as tuples
   initial_state = (7, 2)
   goal_state = (4, 14)
  
   # Encoding other variables given in the problem statement
   num_rows = 15
   num_cols = 15
   matrix = [[ 'x', 'x', 7, 18, 11, 4, 9, 'x', 15, 'x', 'x', 'x', 1, 'x', 'x'], [ 'x', 'x', 14, 'x', 10, 'x', 'x', 'x', 8, 'x', 'x', 16, 1, 'x', 'x'], [ 17, 17, 'x', 'x', 'x', 'x', 'x', 7, 6, 12, 'x', 'x', 'x', 7, 'x'], [ 'x', 'x', 'x', 'x', 10, 'x', 'x', 18, 1, 11, 4, 'x', 13, 10, 16], [ 'x', 'x', 'x', 12, 1, 'x', 'x', 'x', 2, 7, 1, 'x', 2, 'x', 5], [ 'x', 'x', 8, 'x', 12, 'x', 'x', 5, 17, 'x', 2, 'x', 4, 10, 18], [ 'x', 'x', 4, 19, 1, 'x', 18, 7, 'x', 10, 3, 'x', 19, 16, 19], [ 6, 12, 15, 16, 5, 9, 16, 18, 10, 15, 5, 'x', 5, 14, 'x'], [ 'x', 16, 8, 17, 12, 11, 16, 8, 9, 9, 7, 4, 5, 20, 3], [ 19, 17, 15, 'x', 'x', 17, 'x', 3, 2, 2, 11, 7, 8, 16, 1], [ 13, 4, 17, 'x', 'x', 'x', 'x', 5, 'x', 18, 16, 15, 19, 4, 'x'], [ 11, 'x', 'x', 'x', 'x', 12, 'x', 'x', 'x', 'x', 'x', 'x', 19, 'x', 'x'], [ 18, 'x', 'x', 'x', 9, 19, 'x', 16, 6, 'x', 9, 3, 16, 15, 'x'], [ 'x', 'x', 'x', 10, 'x', 'x', 4, 3, 'x', 3, 'x', 16, 18, 'x', 12],[ 'x', 13, 'x', 'x', 'x', 'x', 'x', 10, 5, 'x', 17, 'x', 'x', 7, 'x']]
   districts = {1: range(0, 5), 2: range(5, 9), 3: range(9, 15)}
   # Initialize a dictionary to store the cost of reaching each visited state
   visited_costs = {}
   visited_costs[initial_state] = 0
   # Initialize a priority queue of states not yet visited, with the initial state as the first element. The priority of each element is the cost to reach that state (g) + the estimate remaining cost (h) to reach the goal
   # Record the actions required to get to each state in a list; no actions performed to reach the initial state
   queue = [(0, 0, [initial_state], initial_state, {1: False, 2: True, 3: False})]
   return initial_state, goal_state, num_rows, num_cols, matrix, districts, visited_costs, queue
\end{lstlisting} \\
\hline
\end{tabular}
\captionsetup{belowskip=0pt}
\caption{\fontsize{8pt}{8.5pt}\selectfont GPT4 successfully generated an optimal and correct solution for this traffic problem instance using MSMT A* approach. Prompting GPT4 with A* results in a feasible but incorrect solution and all other three baseline code and text-based prompting methods resulted in infeasible solutions.}
\end{table}

\minisection{Trampoline Matrix}

\begin{table}[H]
\centering
\begin{tabular}{|p{14.5cm}|}
\hline
\fontsize{7.9pt}{8.1pt}\selectfont \textbf{trampoline\_matrix instance (diff\_sorted\_id: 4)}: Alex is at a trampoline park with a grid of mini trampolines, arranged in a square of 9x9. Some trampolines are broken and unusable. A map of the park is provided below, with 1 indicating a broken trampoline and 0 indicating a functional one. Alex can jump to any of the eight adjacent trampolines, as long as they are not broken. However, Alex must make excatly 3 diagonal jumps, no more, no less, on his path to his destination. He is currently on the trampoline at position (8, 1) (positions are counted from 0, left to right, top to bottom) and wants to reach the trampoline at position (0, 5). What is the shortest sequence of trampolines he should jump on to reach his destination (including the first and final trampolines)? The answer should be a list of tuples, in Python syntax, indicating the row and column of each trampoline Alex jumps on. \textbackslash n0 0 1 0 0 0 0 0 1\textbackslash n0 1 0 1 1 0 0 0 0\textbackslash n1 1 1 1 1 1 1 1 0\textbackslash n1 1 0 1 1 0 0 0 0\textbackslash n1 0 1 0 1 0 1 1 1\textbackslash n1 0 1 0 0 0 1 0 1\textbackslash n0 1 0 0 0 1 1 0 1\textbackslash n0 0 1 0 1 1 1 1 1\textbackslash n1 0 0 1 1 0 0 1 1\\
\hline
{\fontsize{8pt}{8.5pt}\selectfont A* algorithm generated by GPT4 for the problem above using A* \smodel{} approach that passed the unit tests:}\begin{lstlisting}[language=Python]
import heapq

def initialize():
   # Define the initial state and the goal state of the problem, represented as tuples
   initial_state = (8, 1)
   goal_state = (0, 5)
   # Encoding other variables given in the problem statement
   num_rows = 9
   num_cols = 9
   park_map = [[0, 0, 1, 0, 0, 0, 0, 0, 1],[0, 1, 0, 1, 1, 0, 0, 0, 0],[1, 1, 1, 1, 1, 1, 1, 1, 0],[1, 1, 0, 1, 1, 0, 0, 0, 0],[1, 0, 1, 0, 1, 0, 1, 1, 1],[1, 0, 1, 0, 0, 0, 1, 0, 1],[0, 1, 0, 0, 0, 1, 1, 0, 1],[0, 0, 1, 0, 1, 1, 1, 1, 1],[1, 0, 0, 1, 1, 0, 0, 1, 1]]
   diagonal_jumps = 3
   # Initialize a dictionary to store the cost of reaching each visited state
   visited_costs = {}
   visited_costs[(initial_state, diagonal_jumps)] = 0
   # Initialize a priority queue of states not yet visited, with the initial state as the first element. The priority of each element is the cost to reach that state (g) + the estimate remaining cost (h) to reach the goal
   # Record the actions required to get to each state in a list; no actions performed to reach the initial state
   queue = [(0, 0, [], (initial_state, diagonal_jumps))]
   return initial_state, goal_state, num_rows, num_cols, park_map, diagonal_jumps, visited_costs, queue
   
def a_star():
   # The initialize function initializes and returns the visited_costs dictionary and the priority queue and encodes all of the variables given in the problem (ie the initial and goal state, dimensions of the park, park map, and number of diagonal jumps)
   initial_state, goal_state, num_rows, num_cols, park_map, diagonal_jumps, visited_costs, queue = initialize()
   # While there are un-visited states
   while queue:
       # Pop the state with the lowest sum of the cost so far and estimated cost to the goal from the queue
       _, g, actions, (state, diagonal_jumps_left) = heapq.heappop(queue)
       # We can check if the current state is the goal state with a simple equality check, as the goal state is predefined
       if state == goal_state and diagonal_jumps_left == 0:
           return actions
       # Generate all valid actions from the current state, which includes jumping to any of the 8 neighboring trampolines
       # Generate the coordinates of the neighboring trampolines
       for d_row, d_col in [(0, -1), (0, 1), (1, 0), (-1, 0), (1, 1), (-1, -1), (1, -1), (-1, 1)]:
           new_row, new_col = state[0] + d_row, state[1] + d_col
           # Check if the jump is valid, ie if the coordinate of the trampoline to jump to is a valid coordinate within the bounds of the park and the trampoline is not broken
           if 0 <= new_row < num_rows and 0 <= new_col < num_cols and park_map[new_row][new_col] == 0:
               # The jump is valid, generate the new state
               new_state = (new_row, new_col)
               # The cost so far is the number of jumps made, as our objective is to minimize the number of jumps required to reach the goal state 
               new_cost = g + 1
               # If the jump is diagonal, decrement the number of diagonal jumps left
               if d_row != 0 and d_col != 0:
                   new_diagonal_jumps_left = diagonal_jumps_left - 1
               else:
                   new_diagonal_jumps_left = diagonal_jumps_left
               # If the new state is unvisited or we found a new path with a lower cost to reach this state, add it to the queue of not-yet-visited states
\end{lstlisting} \\
\hline
\end{tabular}
\caption*{\fontsize{8pt}{8.5pt}\selectfont The GPT4's implementation of the A* algorithm continues on the next page.}
\end{table}

\begin{table}[H]
\centering
\begin{tabular}{|p{14cm}|}
\hline
{\fontsize{8pt}{8.5pt}\selectfont \textbf{Continuation of GPT4's implementation: }\begin{lstlisting}[language=Python]
               if (new_state, new_diagonal_jumps_left) not in visited_costs or new_cost < visited_costs[(new_state, new_diagonal_jumps_left)]:
                   visited_costs[(new_state, new_diagonal_jumps_left)] = new_cost
                   heapq.heappush(queue, (new_cost + heuristic(new_state, goal_state), new_cost, actions + [new_state], (new_state, new_diagonal_jumps_left)))   
   return None

def heuristic(state, goal):
   # An admissible and consistent heuristic is the Manhattan distance (the shortest path) of the current position from the goal position
   # The heuristic relaxes the constraint that Alex must make exactly 3 diagonal jumps and presumes Alex can move to the goal position by jumping to any of the neighboring trampolines
   # Thus the heuristic reports a lower estimate on the cost to reach goal state and is admissible
   # The heuristic is consistent because the cost of moving to a neighboring coordinate is always 1, which is exactly the decrease in the Manhattan distance, if Alex moves toward the goal position, otherwise the estimated cost of the successor node is the same or higher, and he heuristic estimate for the goal state is 0, as the distance of the current position from the goal position would be 0 in the goal state.
   h = abs(state[0] - goal[0]) + abs(state[1] - goal[1])
   return h
   
print(a_star())
\end{lstlisting}} \\
\hline
\end{tabular}
\caption{\fontsize{8pt}{8.5pt}\selectfont The A* algorithm was generated by GPT4 using the MSMT A* approach. GPT4's fourth implementation of the A* algorithm for trampoline\_matrix problem type passed the unit tests.}
\end{table}

\begin{table}[H]
\centering
\begin{tabular}{|p{14.5cm}|}
\hline
\fontsize{7.8pt}{8pt}\selectfont \textbf{trampoline\_matrix instance (diff\_sorted\_id: 102)}: Alex is at a trampoline park with a grid of mini trampolines, arranged in a square of 15x15. Some trampolines are broken and unusable. A map of the park is provided below, with 1 indicating a broken trampoline and 0 indicating a functional one. Alex can jump to any of the eight adjacent trampolines, as long as they are not broken. However, Alex must make excatly 5 diagonal jumps, no more, no less, on his path to his destination. He is currently on the trampoline at position (14, 13) (positions are counted from 0, left to right, top to bottom) and wants to reach the trampoline at position (2, 5). What is the shortest sequence of trampolines he should jump on to reach his destination (including the first and final trampolines)? The answer should be a list of tuples, in Python syntax, indicating the row and column of each trampoline Alex jumps on. \textbackslash n1 1 0 1 1 1 0 0 0 1 0 1 0 1 0\textbackslash n0 1 0 0 1 0 0 0 0 1 0 1 1 1 0\textbackslash n1 1 1 1 0 0 0 0 0 0 1 1 0 0 1\textbackslash n1 1 1 1 0 0 1 1 0 1 1 1 1 0 0\textbackslash n1 1 0 1 0 0 1 0 0 0 0 0 0 0 0\textbackslash n1 0 1 0 1 1 1 0 1 1 0 1 0 0 1\textbackslash n1 1 0 0 0 0 0 0 1 0 0 1 0 1 1\textbackslash n0 1 1 1 1 1 0 0 1 1 0 1 0 1 1\textbackslash n0 0 1 1 1 0 0 1 1 0 0 1 0 0 1\textbackslash n1 1 1 1 1 0 1 0 0 1 1 0 0 0 0\textbackslash n1 0 0 1 0 1 1 0 0 0 1 1 1 0 0\textbackslash n0 0 0 1 0 0 1 1 0 1 0 0 0 0 0\textbackslash n0 1 1 1 1 0 0 0 1 1 1 0 0 0 0\textbackslash n1 1 1 1 1 1 0 0 0 1 1 0 0 0 0\textbackslash n1 1 1 0 0 1 1 0 1 1 1 0 1 0 1\\
\hline
{\fontsize{8pt}{8pt}\selectfont 'initialize' function generated by GPT4 for the problem above in the second stage of the MSMT approach}\begin{lstlisting}[language=Python]
def initialize():
   # Define the initial state and the goal state of the problem, represented as tuples
   initial_state = (14, 13)
   goal_state = (2, 5)
   # Encoding other variables given in the problem statement
   num_rows = 15
   num_cols = 15
   park_map = [[1, 1, 0, 1, 1, 1, 0, 0, 0, 1, 0, 1, 0, 1, 0],[0, 1, 0, 0, 1, 0, 0, 0, 0, 1, 0, 1, 1, 1, 0],[1, 1, 1, 1, 0, 0, 0, 0, 0, 0, 1, 1, 0, 0, 1],[1, 1, 1, 1, 0, 0, 1, 1, 0, 1, 1, 1, 1, 0, 0],[1, 1, 0, 1, 0, 0, 1, 0, 0, 0, 0, 0, 0, 0, 0],[1, 0, 1, 0, 1, 1, 1, 0, 1, 1, 0, 1, 0, 0, 1],[1, 1, 0, 0, 0, 0, 0, 0, 1, 0, 0, 1, 0, 1, 1],[0, 1, 1, 1, 1, 1, 0, 0, 1, 1, 0, 1, 0, 1, 1],[0, 0, 1, 1, 1, 0, 0, 1, 1, 0, 0, 1, 0, 0, 1],[1, 1, 1, 1, 1, 0, 1, 0, 0, 1, 1, 0, 0, 0, 0],[1, 0, 0, 1, 0, 1, 1, 0, 0, 0, 1, 1, 1, 0, 0],[0, 0, 0, 1, 0, 0, 1, 1, 0, 1, 0, 0, 0, 0, 0],[0, 1, 1, 1, 1, 0, 0, 0, 1, 1, 1, 0, 0, 0, 0],[1, 1, 1, 1, 1, 1, 0, 0, 0, 1, 1, 0, 0, 0, 0],[1, 1, 1, 0, 0, 1, 1, 0, 1, 1, 1, 0, 1, 0, 1]]
   diagonal_jumps = 5
   # Initialize a dictionary to store the cost of reaching each visited state
   visited_costs = {}
   visited_costs[(initial_state, diagonal_jumps)] = 0
   # Initialize a priority queue of states not yet visited, with the initial state as the first element. The priority of each element is the cost to reach that state (g) + the estimate remaining cost (h) to reach the goal
   # Record the actions required to get to each state in a list; no actions performed to reach the initial state
   queue = [(0, 0, [initial_state], (initial_state, diagonal_jumps))]
   return initial_state, goal_state, num_rows, num_cols, park_map, diagonal_jumps, visited_costs, queue
\end{lstlisting} \\
\hline
\end{tabular}
\captionsetup{belowskip=0pt}
\caption{\fontsize{8pt}{8pt}\selectfont GPT4 successfully generated a correct (but non-optimal) solution for this trampoline\_matrix problem instance using MSMT A* approach. Prompting GPT4 with A* results in a feasible but incorrect solution and all other three baseline code and text-based prompting methods resulted in infeasible solutions.}
\end{table}

\minisection{City Directed Graph}

\begin{table}[H]
\centering
\begin{tabular}{|p{14.5cm}|}
\hline
\footnotesize \textbf{city\_directed\_graph instance (diff\_sorted\_id: 5)}: We have a map of cities, each represented by a letter, and they are connected by one-way roads. The adjacency matrix below shows the connections between the cities. Each row and column represents a city, and a '1' signifies a direct road from the city of the row to the city of the column. The travel time between any two directly connected cities is the same. Currently, we are located in city 'H'. Our task is to visit city Z and city K excatly twice. Determine the quickest route that allows us to visit both these destination cities, ensuring that we stop at the two destinations twice on our path. The sequence in which we visit the destination cities is not important. However, apart from K and Z, we can only visit each city once on our path. Provide the solution as a list of the city names on our path, including the start, in Python syntax.\textbackslash n\textbackslash n  K B Z S X H O E U V \textbackslash nK 0 0 1 1 0 0 1 1 0 1 \textbackslash nB 0 0 0 0 0 0 0 1 0 1 \textbackslash nZ 0 1 0 1 0 1 1 0 0 1 \textbackslash nS 0 1 1 0 1 0 0 0 0 0 \textbackslash nX 1 1 0 0 0 0 0 0 0 1 \textbackslash nH 0 0 0 1 0 0 1 0 1 0 \textbackslash nO 0 0 1 1 1 0 0 0 0 0 \textbackslash nE 1 0 0 0 0 0 1 0 0 1 \textbackslash nU 0 0 0 0 0 0 1 0 0 0 \textbackslash nV 0 0 0 0 1 0 0 0 0 0 \textbackslash n\\
\hline
{\footnotesize A* algorithm generated by GPT4 for the problem above using A* \smodel{} approach that passed the unit tests:}\begin{lstlisting}[language=Python]
import heapq

def initialize():
   # Define the initial state and the goal state of the puzzle, represented as 3x3 tuples
   initial_state = 'H'
   goal_state = ['K', 'K', 'Z', 'Z']   
   # Encoding other variables given in the problem statement
   adjacency_matrix = [[0, 0, 1, 1, 0, 0, 1, 1, 0, 1],[0, 0, 0, 0, 0, 0, 0, 1, 0, 1],[0, 1, 0, 1, 0, 1, 1, 0, 0, 1],[0, 1, 1, 0, 1, 0, 0, 0, 0, 0],[1, 1, 0, 0, 0, 0, 0, 0, 0, 1],[0, 0, 0, 1, 0, 0, 1, 0, 1, 0],[0, 0, 1, 1, 1, 0, 0, 0, 0, 0],[1, 0, 0, 0, 0, 0, 1, 0, 0, 1],[0, 0, 0, 0, 0, 0, 1, 0, 0, 0],[0, 0, 0, 0, 1, 0, 0, 0, 0, 0]]
   cities = ['K', 'B', 'Z', 'S', 'X', 'H', 'O', 'E', 'U', 'V']
   # Initialize a dictionary to store the cost of reaching each visited state
   visited_costs = {}
   visited_costs[(initial_state, tuple(goal_state))] = 0
   # Initialize a priority queue of states not yet visited, with the initial state as the first element. The priority of each element is the cost to reach that state (g) + the estimate remaining cost (h) to reach the goal
   # Record the actions required to get to each state in a list; no actions performed to reach the initial state
   queue = [(0, 0, [], (initial_state, tuple(goal_state)))]  
   return initial_state, goal_state, adjacency_matrix, cities, visited_costs, queue
   
def a_star():
   # The initialize function initializes and returns the visited_costs dictionary and the priority queue and encodes all of the variables given in the problem (ie the initial and goal board and dimensions of the puzzle board)
   initial_state, goal_state, adjacency_matrix, cities, visited_costs, queue = initialize()
   # While there are un-visited states
   while queue:
       # Pop the state with the lowest sum of the cost so far and estimated cost to the goal from the queue
       _, g, actions, state = heapq.heappop(queue)
       current_city, remaining_cities = state
       # We can check if the current state is the goal state with a simple equality check, as the goal state is predefined
       if not remaining_cities:
           return actions
       # Generate all valid actions from the current state, which includes moving to any of the cities connected to the current city
       for i in range(len(cities)):
           if adjacency_matrix[cities.index(current_city)][i] == 1:
               next_city = cities[i]
               # Check if the next city is a destination city and we have not visited it twice already
               if next_city in remaining_cities:
                   new_remaining_cities = list(remaining_cities)
                   new_remaining_cities.remove(next_city)
                   new_state = (next_city, tuple(new_remaining_cities))
                   new_cost = g + 1
                   if new_state not in visited_costs or new_cost < visited_costs[new_state]:
                       visited_costs[new_state] = new_cost
                       heapq.heappush(queue, (new_cost + heuristic(new_state, goal_state), new_cost, actions + [next_city], new_state))
\end{lstlisting} \\
\hline
\end{tabular}
\caption*{\footnotesize The GPT4's implementation of the A* algorithm continues on the next page.}
\end{table}

\begin{table}[H]
\centering
\begin{tabular}{|p{14cm}|}
\hline
{\fontsize{8pt}{8.5pt}\selectfont \textbf{Continuation of GPT4's implementation: }\begin{lstlisting}[language=Python]
               # If the next city is not a destination city and we have not visited it before, we can move to it
               elif next_city not in actions:
                   new_state = (next_city, remaining_cities)
                   new_cost = g + 1
                   if new_state not in visited_costs or new_cost < visited_costs[new_state]:
                       visited_costs[new_state] = new_cost
                       heapq.heappush(queue, (new_cost + heuristic(new_state, goal_state), new_cost, actions + [next_city], new_state))
   return None

def heuristic(state, goal_state):
   # An admissible and consistent heuristic is the number of remaining cities to visit
   # The heuristic relaxes the constraint that we can only move to cities connected to the current city and presumes we can move to any of the remaining cities directly from the current city
   # Thus the heuristic reports a lower estimate on the cost to reach goal state and is admissible
   # The heuristic is consistent because the cost of moving to a city is always 1, which is exactly the decrease in the number of remaining cities to visit, if the city is a destination city, otherwise the estimated cost of the successor node is the same or higher, and he heuristic estimate for the goal state is 0, as there are no remaining cities to visit in the goal state.
   return len(state[1])

print(a_star())
\end{lstlisting}} \\
\hline
\end{tabular}
\caption{\fontsize{8.3pt}{8.5pt}\selectfont The A* algorithm was generated by GPT4 using the MSMT A* approach. GPT4's fifth implementation of the A* algorithm for city\_directed\_graph problem type passed the unit tests.}
\end{table}

\begin{table}[H]
\centering
\begin{tabular}{|p{14.5cm}|}
\hline
\fontsize{8.5pt}{8.8pt}\selectfont \textbf{city\_directed\_graph instance (diff\_sorted\_id: 84)}: We have a map of cities, each represented by a letter, and they are connected by one-way roads. The adjacency matrix below shows the connections between the cities. Each row and column represents a city, and a '1' signifies a direct road from the city of the row to the city of the column. The travel time between any two directly connected cities is the same. Currently, we are located in city 'T'. Our task is to visit city H and city K excatly twice. Determine the quickest route that allows us to visit both these destination cities, ensuring that we stop at the two destinations twice on our path. The sequence in which we visit the destination cities is not important. However, apart from K and H, we can only visit each city once on our path. Provide the solution as a list of the city names on our path, including the start, in Python syntax.\textbackslash n\textbackslash n  F O L P Y R Q B T V A K S H \textbackslash nF 0 0 0 1 1 0 0 1 0 0 0 0 0 0 \textbackslash nO 0 0 1 0 1 0 1 0 1 0 0 1 0 0 \textbackslash nL 1 0 0 0 0 1 1 0 0 0 1 1 1 0 \textbackslash nP 0 0 1 0 1 0 0 1 0 1 1 0 0 0 \textbackslash nY 0 0 0 0 0 0 0 1 0 0 1 0 1 0 \textbackslash nR 0 0 0 0 1 0 1 1 0 0 0 0 1 0 \textbackslash nQ 0 0 0 1 0 0 0 0 0 0 0 1 0 1 \textbackslash nB 0 1 1 0 0 0 1 0 0 1 0 0 1 1 \textbackslash nT 1 0 0 0 0 0 1 0 0 1 0 0 0 0 \textbackslash nV 0 0 0 0 0 0 0 0 1 0 0 1 1 1 \textbackslash nA 1 1 0 0 0 0 0 1 1 1 0 0 0 1 \textbackslash nK 1 0 0 1 1 0 0 0 1 0 1 0 0 0 \textbackslash nS 0 1 0 0 0 0 1 0 1 1 1 1 0 0 \textbackslash nH 1 0 0 1 0 1 0 0 0 0 0 0 1 0 \textbackslash n\\
\hline
{\fontsize{8pt}{8pt}\selectfont 'initialize' function generated by GPT4 for the problem above in the second stage of the MSMT approach}\begin{lstlisting}[language=Python]
def initialize():
   # Define the initial state and the goal state of the puzzle
   initial_state = ('T', ('K', 'K', 'H', 'H'))   
   goal_state = []   
   # Encoding other variables given in the problem statement
   adjacency_matrix = [[0, 0, 0, 1, 1, 0, 0, 1, 0, 0, 0, 0, 0, 0], [0, 0, 1, 0, 1, 0, 1, 0, 1, 0, 0, 1, 0, 0], [1, 0, 0, 0, 0, 1, 1, 0, 0, 0, 1, 1, 1, 0], [0, 0, 1, 0, 1, 0, 0, 1, 0, 1, 1, 0, 0, 0], [0, 0, 0, 0, 0, 0, 0, 1, 0, 0, 1, 0, 1, 0], [0, 0, 0, 0, 1, 0, 1, 1, 0, 0, 0, 0, 1, 0], [0, 0, 0, 1, 0, 0, 0, 0, 0, 0, 0, 1, 0, 1], [0, 1, 1, 0, 0, 0, 1, 0, 0, 1, 0, 0, 1, 1], [1, 0, 0, 0, 0, 0, 1, 0, 0, 1, 0, 0, 0, 0], [0, 0, 0, 0, 0, 0, 0, 0, 1, 0, 0, 1, 1, 1], [1, 1, 0, 0, 0, 0, 0, 1, 1, 1, 0, 0, 0, 1], [1, 0, 0, 1, 1, 0, 0, 0, 1, 0, 1, 0, 0, 0], [0, 1, 0, 0, 0, 0, 1, 0, 1, 1, 1, 1, 0, 0], [1, 0, 0, 1, 0, 1, 0, 0, 0, 0, 0, 0, 1, 0]]
   cities = ['F', 'O', 'L', 'P', 'Y', 'R', 'Q', 'B', 'T', 'V', 'A', 'K', 'S', 'H']
   # Initialize a dictionary to store the cost of reaching each visited state
   visited_costs = {}
   visited_costs[initial_state] = 0
   # Initialize a priority queue of states not yet visited, with the initial state as the first element. The priority of each element is the cost to reach that state (g) + the estimate remaining cost (h) to reach the goal
   # Record the actions required to get to each state in a list; no actions performed to reach the initial state
   queue = [(0, 0, [initial_state[0]], initial_state)]
   return initial_state, goal_state, adjacency_matrix, cities, visited_costs, queue
\end{lstlisting} \\
\hline
\end{tabular}
\captionsetup{belowskip=0pt}
\caption{\fontsize{8.3pt}{8pt}\selectfont GPT4 successfully generated an optimal solution for this city\_directed\_graph problem instance using MSMT A* approach. Prompting GPT4 with all other four code and text-based methods resulted in infeasible solutions.}
\end{table}

\clearpage

\section{Prompts} \label{full_prompts}

In this section, we provide the complete in-context examples and instructions given to the LLMs in each of the five prompting methods used in our experiments. Additionally, we present GPT4's generated response for a pathfinding problem using each of these five prompts.

\minisection{0\_shot text}

\begin{table}[H]
\centering
\begin{tabular}{|p{14.5cm}|}
\hline
{ \textbf{0\_shot text prompt}: Solve the problem below step-by-step, using only text. The solution should be presented right after the phrase /answer\_list/. Ensure there is no text following the reported solution. The solution should be formatted as a Python list, and each element in the list should be of the data type specified in the problem.}\\
\hline
\footnotesize \textbf{problem statement (city\_directed\_graph instance 8)}: We have a map of cities, each represented by a letter, and they are connected by one-way roads. The adjacency matrix below shows the connections between the cities. Each row and column represents a city, and a '1' signifies a direct road from the city of the row to the city of the column. The travel time between any two directly connected cities is the same. Currently, we are located in city 'U'. Our task is to visit city A and city X excatly twice. Determine the quickest route that allows us to visit both these destination cities, ensuring that we stop at the two destinations twice on our path. The sequence in which we visit the destination cities is not important. However, apart from X and A, we can only visit each city once on our path. Provide the solution as a list of the city names on our path, including the start, in Python syntax.\textbackslash n\textbackslash n  Y O Q X K R A U M L \textbackslash nY 0 1 0 0 0 1 1 1 1 1 \textbackslash nO 0 0 0 0 0 0 1 0 0 0 \textbackslash nQ 1 0 0 1 0 1 0 0 1 0 \textbackslash nX 1 0 1 0 0 1 0 0 1 1 \textbackslash nK 0 1 0 1 0 0 1 0 0 0 \textbackslash nR 0 0 0 0 1 0 0 0 0 0 \textbackslash nA 0 0 0 1 0 1 0 0 1 1 \textbackslash nU 0 0 0 0 0 0 0 0 0 1 \textbackslash nM 0 0 1 0 0 0 0 0 0 0 \textbackslash nL 0 1 0 0 0 0 0 0 0 0 \textbackslash n\\
\hline
{\footnotesize \textbf{GPT4's generation}: {\ttfamily

To solve this problem, we need to find the quickest route that allows us to visit both city A and city X twice, starting from city U. We can only visit each city once, except for city A and city X. 

From the adjacency matrix, we can see that city U is only directly connected to city L. So, we start our journey from city U to city L.

From city L, we can only go to city O. So, our next stop is city O.

From city O, we can only go to city A. So, our next stop is city A. This is our first visit to city A.

From city A, we can go to either city X, city R, or city Y. But since we need to visit city X twice, we choose to go to city X next. This is our first visit to city X.

From city X, we can go to either city Y, city Q, or city A. But since we need to visit city A twice, we choose to go to city A next. This is our second visit to city A.

From city A, we can go to either city X, city R, or city Y. But since we need to visit city X twice, we choose to go to city X next. This is our second visit to city X.

So, the quickest route that allows us to visit both city A and city X twice, starting from city U, is U -> L -> O -> A -> X -> A -> X.

/answer\_list/ ['U', 'L', 'O', 'A', 'X', 'A', 'X']}} \\
\hline
\textbf{LLM-solution extracted}: ['U', 'L', 'O', 'A', 'X', 'A', 'X']

\textbf{LLM-solution is not feasible}\\
\hline
\end{tabular}
\caption{\footnotesize GPT4's solution for a city\_directed\_graph problem using 0\_shot text prompting method.}
\end{table}

\minisection{4\_shot CoT text}

\begin{table}[H]
\centering
\begin{tabular}{|p{14.5cm}|}
\hline
{\fontsize{9pt}{8pt}\selectfont \textbf{4\_shot CoT text prompt}:The objective is to solve the problems below step-by-step, using only text. The solution should be presented right after the phrase /answer\_list/. Ensure there is no text following the reported solution. The solution should be formatted as a Python list, and each element in the list should be of the data type specified in the problem.}

Examples:

\\
In the 8-puzzle game, you are given a grid with numbered square tiles arranged randomly and one tile missing. The goal is to arrange the tiles in descending order by sliding them into the empty space. The tiles can move in 4 directions: left, right, up, and down.  Given the initial state of the puzzle below, where the empty spot is represented as “\_”, provide the shortest list of tiles that need to be swapped with the empty spot to achieve the goal state. The goal state is when all tiles are in descending order, with the largest number in the top left corner, and the empty spot is in the bottom right corner. The solution should be a list of numbers in Python format, where each number represents the number on the tile that the empty spot is swapped with at each turn. Initial state of the puzzle: [[55, 43, 17], [97, 35, 9], [12, 25, '\_']]

\\

The goal state of the puzzle is [[97, 55, 43], [35, 25, 17], [12, 9, '\_']].

To achieve this, we need to follow these steps:

1. Swap 9 with '\_': [[55, 43, 17], [97, 35, '\_'], [12, 25, 9]]

2. Swap 17 with '\_': [[55, 43, '\_'], [97, 35, 17], [12, 25, 9]]

3. Swap 43 with '\_': [[55, '\_', 43], [97, 35, 17], [12, 25, 9]]

4. Swap 55 with '\_': [['\_', 55, 43], [97, 35, 17], [12, 25, 9]]

5. Swap 97 with '\_': [[97, 55, 43], ['\_', 35, 17], [12, 25, 9]]

6. Swap 35 with '\_': [[97, 55, 43], [35, '\_', 17], [12, 25, 9]]

7. Swap 25 with '\_': [[97, 55, 43], [35, 25, 17], [12, '\_', 9]]

8. Swap 9 with '\_': [[97, 55, 43], [35, 25, 17], [12, 9, '\_']]

So, the list of tiles that need to be swapped with the empty spot to achieve the goal state is [9, 17, 43, 55, 97, 35, 25, 9].

The answer is /answer\_list/[9, 17, 43, 55, 97, 35, 25, 9]

---------

The game of 'Sort It' begins with 3 tubes, each filled with balls of different colors. The goal is to sort the balls by color, with each tube containing balls of only one color. Only one ball can be moved at a time, taken from the top of one tube and placed on top of another. All of the tubes have the same capacity of 6 balls. It is not allowed to place a ball in a tube that is already at capacity. The solution should be a list of tuples, each containing, first, the index of the tube from which a ball is taken and, second, the index of the tube to which it is moved, indexing from 0. Given the initial state of the tubes, represented by the lists below (with the leftmost item being the color of the topmost ball in each tube), what is the shortest list of move tuples that will result in all the balls being correctly sorted? [['Green', 'Red', 'Green', 'Red'], ['Blue', 'Blue', 'Red', 'Green'], ['Red', 'Blue', 'Green', 'Blue']]
\\

The initial state of the tubes is:

Tube 0: ['Green', 'Red', 'Green', 'Red']

Tube 1: ['Blue', 'Blue', 'Red', 'Green']

Tube 2: ['Red', 'Blue', 'Green', 'Blue']

The goal is to sort the balls by color, with each tube containing balls of only one color. The moves are as follows:

1. Move 'Green' from Tube 0 to Tube 2: (0, 2)
The state of the tubes now is:

Tube 0: ['Red', 'Green', 'Red'] Tube 1: ['Blue', 'Blue', 'Red', 'Green'] Tube 2: ['Green', 'Red', 'Blue', 'Green', 'Blue']

2. Move 'Red' from Tube 0 to Tube 1: (0, 1)

Tube 0: ['Green', 'Red'] Tube 1: ['Red', 'Blue', 'Blue', 'Red', 'Green'] Tube 2: ['Green', 'Red', 'Blue', 'Green', 'Blue']

3. Move 'Green' from Tube 0 to Tube 2: (0, 2) 

Tube 0: ['Red'] Tube 1: ['Red', 'Blue', 'Blue', 'Red', 'Green'] Tube 2: ['Green', 'Green', 'Red', 'Blue', 'Green', 'Blue']

\\
\hline
\end{tabular}
\caption*{\fontsize{9pt}{8pt}\selectfont The 4\_shot CoT text prompt continues on the next page.}
\end{table}

\begin{table}[H]
\centering
\begin{tabular}{|p{14.5cm}|}
\hline
{\fontsize{9pt}{8pt}\selectfont \textbf{Continuation of the 4\_shot CoT text:} 

4. Move 'Red' from Tube 1 to Tube 0: (1, 0)

Tube 0: ['Red', 'Red'] Tube 1: ['Blue', 'Blue', 'Red', 'Green'] Tube 2: ['Green', 'Green', 'Red', 'Blue', 'Green', 'Blue']

5. Move 'Green' from Tube 2 to Tube 1: (2, 1)

Tube 0: ['Red', 'Red'] Tube 1: ['Green', 'Blue', 'Blue', 'Red', 'Green'] Tube 2: ['Green', 'Red', 'Blue', 'Green', 'Blue']}

6. Move 'Green' from Tube 2 to Tube 1: (2, 1)

Tube 0: ['Red', 'Red']
Tube 1: ['Green', 'Green', 'Blue', 'Blue', 'Red', 'Green']
Tube 2: ['Red', 'Blue', 'Green', 'Blue']

7. Move 'Red' from Tube 2 to Tube 1: (2, 0)

Tube 0: ['Red', 'Red', 'Red']
Tube 1: ['Green', 'Green', 'Blue', 'Blue', 'Red', 'Green']
Tube 2: ['Blue', 'Green', 'Blue']

8. Move 'Green' from Tube 1 to Tube 0: (1, 0)

Tube 0: ['Green', 'Red', 'Red', 'Red']
Tube 1: ['Green', 'Blue', 'Blue', 'Red', 'Green']
Tube 2: ['Blue', 'Green', 'Blue']

9. Move 'Green' from Tube 1 to Tube 0: (1, 0)

Tube 0: ['Green', 'Green', 'Red', 'Red', 'Red']
Tube 1: ['Blue', 'Blue', 'Red', 'Green']
Tube 2: ['Blue', 'Green', 'Blue']

10. Move 'Blue' from Tube 2 to Tube 1: (2, 1)

Tube 0: ['Green', 'Green', 'Red', 'Red', 'Red']
Tube 1: ['Blue', 'Blue', 'Blue', 'Red', 'Green']
Tube 2: ['Green', 'Blue']

11. Move 'Green' from Tube 2 to Tube 0: (2, 0)

Tube 0: ['Green', 'Green', 'Green', 'Red', 'Red', 'Red']
Tube 1: ['Blue', 'Blue', 'Blue', 'Red', 'Green']
Tube 2: ['Blue']

12. Move 'Blue' from Tube 1 to Tube 2: (1, 2)

Tube 0: ['Green', 'Green', 'Green', 'Red', 'Red', 'Red']
Tube 1: ['Blue', 'Blue', 'Red', 'Green']
Tube 2: ['Blue', 'Blue']

13. Move 'Blue' from Tube 1 to Tube 2: (1, 2)

Tube 0: ['Green', 'Green', 'Green', 'Red', 'Red', 'Red']
Tube 1: ['Blue', 'Red', 'Green']
Tube 2: ['Blue', 'Blue', 'Blue',]

14. Move 'Blue' from Tube 1 to Tube 2: (1, 2)

Tube 0: ['Green', 'Green', 'Green', 'Red', 'Red', 'Red']
Tube 1: ['Red', 'Green']
Tube 2: ['Blue', 'Blue', 'Blue', 'Blue']

15. Move 'Red' from Tube 1 to Tube 2: (1, 2)

Tube 0: ['Green', 'Green', 'Green', 'Red', 'Red', 'Red']
Tube 1: ['Green']
Tube 2: ['Red', 'Blue', 'Blue', 'Blue', 'Blue']

16. Move 'Green' from Tube 0 to Tube 1: (0, 1)

Tube 0: ['Green', 'Green', 'Red', 'Red', 'Red']
Tube 1: ['Green', 'Green']
Tube 2: ['Red', 'Blue', 'Blue', 'Blue', 'Blue']

17. Move 'Green' from Tube 0 to Tube 1: (0, 1)

Tube 0: ['Green', 'Red', 'Red', 'Red']
Tube 1: ['Green', 'Green', 'Green']
Tube 2: ['Red', 'Blue', 'Blue', 'Blue', 'Blue']

18. Move 'Green' from Tube 0 to Tube 1: (0, 1)

Tube 0: ['Red', 'Red', 'Red']
Tube 1: ['Green', 'Green', 'Green', 'Green']
Tube 2: ['Red', 'Blue', 'Blue', 'Blue', 'Blue']

19. Move 'Red' from Tube 2 to Tube 0: (2, 0)

Tube 0: ['Red', 'Red', 'Red', 'Red']
Tube 1: ['Green', 'Green', 'Green', 'Green']
Tube 2: ['Blue', 'Blue', 'Blue', 'Blue']

The final state of the tubes is:

Tube 0: ['Red', 'Red', 'Red', 'Red']

Tube 1: ['Green', 'Green', 'Green', 'Green']

Tube 2: ['Blue', 'Blue', 'Blue', 'Blue']

So, the shortest list of move tuples that will result in all the balls being correctly sorted is [(0, 1), (0, 2), (1, 0), (2, 1), (2, 1), (2, 0), (1, 0), (1, 0), (2, 1), (2, 0), (1, 2), (1, 2), (1, 2), (1, 2), (0, 1), (0, 1), (0, 1), (2, 0)].

The answer is /answer\_list/[(0, 1), (0, 2), (1, 0), (2, 1), (2, 1), (2, 0), (1, 0), (1, 0), (2, 1), (2, 0), (1, 2), (1, 2), (1, 2), (1, 2), (0, 1), (0, 1), (0, 1), (2, 0)]

 \\
\hline
\end{tabular}
\caption*{\fontsize{9pt}{8pt}\selectfont The 4\_shot CoT text prompt continues on the next page.}
\end{table}

\begin{table}[H]
\centering
\begin{tabular}{|p{14.5cm}|}
\hline
{\fontsize{10pt}{9pt}\selectfont \textbf{Continuation of the 4\_shot CoT text:} 

---------

Given 6 labeled water jugs with capacities 37, 133, 38, 72, 41, 23, 122 liters, we aim to fill 3 unlabeled buckets, numbered 1 to 3 and arranged in a line in ascending order, with 195, 224, 268 liters of water respectively. The amount of water in each unlabeled bucket can not at any point in time exceed the amount of water in the bucket placed before it. Jugs can only be filled to the top and emptied completely, and the unlabeled buckets cannot be overfilled. An action, represented as a tuple ('+', X, Y) or ('-', X, Y), involves adding to or removing water from the unlabeled bucket numbered Y, using the jug with capacity X. Determine the shortest sequence of actions needed to fill the buckets as specified, and present the solution as a list of action tuples in Python syntax.}

\\

The problem can be solved using a greedy algorithm. The idea is to start filling the last bucket that has not been filled yet in order to not violate the constraint on the relative amount of water in the buckets based on their order. Here are the steps to solve the problem:

1. Start with the last bucket (bucket 3) and find the largest jug that can be used to fill it without overfilling it or making it difficult to fill the remaining capacity using the given jugs. In this case, the largest jug that can be used is the 122-liter jug. So, the first action is ('+', 122, 3).

2. Repeat the process for the same bucket (bucket 3) until it is filled with 268 liters of water. Each time, find the largest jug that can be used without overfilling the bucket. The actions are ('+', 72, 3), ('+', 37, 3), ('+', 37, 3).

3. Move to the next bucket (bucket 2) and repeat the process until the bucket is filled with 224 liters of water. The actions are ('+', 122, 2),('+', 41, 2),('+', 38, 2), ('+', 23, 2).

4. Move to the last bucket (bucket 1) and repeat the process until the bucket is filled with 195 liters of water. The greedy actions are ('+', 122, 1),  ('+', 72, 1).

5. However, the last bucket (bucket 1) is filled with 194 liters, 1 liter less than required. Since there is no 1 liter jug, the 38 liter jug can be used to add water, then 37 liters is removed using the 37 liter jug, adding a net of 1 liter to the bucket. We can not use the 38 liter jug to add water after filling the bucket with 194 liters of water because it will overflow. Thus the order of actions to fill the last bucket is: ('+', 122, 1), ('+', 72, 1), ('-', 37, 1), ('+', 38, 1).

The final sequence of actions is [('+', 122, 3), ('+', 72, 3), ('+', 37, 3), ('+', 37, 3), ('+', 122, 2),('+', 41, 2),('+', 38, 2), ('+', 23, 2), ('+', 122, 1),  ('+', 38, 1), ('-', 37, 1), ('+', 72, 1)].

The answer is /answer\_list/[('+', 122, 3), ('+', 72, 3), ('+', 37, 3), ('+', 37, 3), ('+', 122, 2),('+', 41, 2),('+', 38, 2), ('+', 23, 2), ('+', 122, 1), ('+', 72, 1), ('-', 37, 1), ('+', 38, 1)]

---------

"In the magic square problem, a 4x4 grid is filled with unique integers ranging from 29 to 54. Some numbers are already given, while others are unknown and represented as 'x'. The sums of columns 1 to 2 (counting from 0) must be 148, 196 respectively, and the sums of rows 1 to 2 must be 187, 149 respectively. Also, the sum of the numbers in the diagonal from the top left to the bottom right corner of the grid should equal 166. The goal is to find unique integers in the given range to replace with ‘x’s in the grid below such that the sum of the specified rows, columns, and diagonal equals the given amounts and the sum of all of the numbers in the grid is as low as possible. The solution should be provided as a list of tuples in Python syntax. Each tuple should contain three numbers: the row index, the column index (both starting from 0), and the value of the unknown number at that position.\textbackslash n\textbackslash nGrid:\textbackslash n [['47' 'x' 'x' '32']\textbackslash n ['x' 'x' 'x' '49']\textbackslash n ['x' '31' '50' 'x']\textbackslash n ['x' 'x' '52' '30']]

\\

Step 1: The sum of the diagonal is given with only 1 element missing (in index (1, 1)). The sum of the diagonal is 166 and the known values in the diagonal are 47, 50, and 30. So, the missing value is 166 - 47 - 50 - 30 = 39. So, we fill the position (1, 1) with 39.

Grid after Step 1:

 [['47' 'x' 'x' '32']

 ['x' '39' 'x' '49']
 
 ['x' '31' '50' 'x']
 
 ['x' 'x' '52' '30']]

Step 2: The sum of the first row is 187 and the known values in the first row are 39 and 49. So, the sum of the missing values is 187 - 49 - 39 = 99. We fill the missing values with the unique integers that sum to 99, which are 46 and 53. So, we fill the positions (1, 0) and (1, 2) with 46 and 53 respectively.

Grid after Step 2:

 [['47' 'x' 'x' '32']
 
 ['46' '39' '53' '49']
 
 ['x' '31' '50' 'x']
 
 ['x' 'x' '52' '30']]

 \\
\hline
\end{tabular}
\caption*{\fontsize{9pt}{8pt}\selectfont The 4\_shot CoT text prompt continues on the next page.}
\end{table}

\begin{table}[H]
\centering
\begin{tabular}{|p{14.5cm}|}
\hline
{\fontsize{10pt}{9pt}\selectfont \textbf{Continuation of the 4\_shot CoT text:} 

Step 3: The sum of the second column is given with only 1 element missing (in index (0, 2)). The sum of the second column is 196 and the known values in the diagonal are 53, 50, and 52. So, the missing value is 196 - 53 - 50 - 52 = 41. So, we fill the position (0, 2) with 41.

Grid after Step 3:

 [['47' 'x' '41' '32']
 
 ['46' '39' '53' '49']
 
 ['x' '31' '50' 'x']
 
 ['x' 'x' '52' '30']]

Step 4: The sum of the second row is 149 and the known values in the second row are 31 and 50. So, the sum of the missing values is 149 - 31 - 50 = 68. We fill the missing values with the unique integers that sum to 68, which are 33 and 35. So, we fill the positions (2, 0) and (2, 3) with 33 and 35 respectively.

Grid after Step 4:

 [['47' 'x' '41' '32']
 
 ['46' '39' '53' '49']
 
 ['33' '31' '50' '35']
 
 ['x' 'x' '52' '30']]

Step 5: The sum of the first column is 148 and the known values in the first column are 39 and 31. So, the sum of the missing values is 148 - 31 - 39 = 78. We fill the missing values with the unique integers that sum to 78, which are 34 and 44. So, we fill the positions (0, 1) and (3, 1) with 34 and 44 respectively.

Grid after Step 5:

 [['47' '34' '41' '32']
 
 ['46' '39' '53' '49']
 
 ['33' '31' '50' '35']
 
 ['x' '44' '52' '30']]

Step 6: There is 1 unknown number left in the grid at position (3,0). We must fill this cell with the smallest unique integer which is 29.

Grid after Step 6:

 [['47' '34' '41' '32']
 
 ['46' '39' '53' '49']
 
 ['33' '31' '50' '35']
 
 ['29' '44' '52' '30']]

So, the solution is [(0, 1, 34), (0, 2, 41), (1, 0, 46), (1, 1, 39), (1, 2, 53), (2, 0, 33), (2, 3, 35), (3, 0, 29), (3, 1, 44)].

The answer is /answer\_list/[(0, 1, 34), (0, 2, 41), (1, 0, 46), (1, 1, 39), (1, 2, 53), (2, 0, 33), (2, 3, 35), (3, 0, 29), (3, 1, 44)]}

---------

Solve the problem below in the same format, step-by-step, using only text.\\
\hline
\footnotesize \textbf{problem statement (city\_directed\_graph instance 8)}: We have a map of cities, each represented by a letter, and they are connected by one-way roads. The adjacency matrix below shows the connections between the cities. Each row and column represents a city, and a '1' signifies a direct road from the city of the row to the city of the column. The travel time between any two directly connected cities is the same. Currently, we are located in city 'U'. Our task is to visit city A and city X excatly twice. Determine the quickest route that allows us to visit both these destination cities, ensuring that we stop at the two destinations twice on our path. The sequence in which we visit the destination cities is not important. However, apart from X and A, we can only visit each city once on our path. Provide the solution as a list of the city names on our path, including the start, in Python syntax.\textbackslash n\textbackslash n  Y O Q X K R A U M L \textbackslash nY 0 1 0 0 0 1 1 1 1 1 \textbackslash nO 0 0 0 0 0 0 1 0 0 0 \textbackslash nQ 1 0 0 1 0 1 0 0 1 0 \textbackslash nX 1 0 1 0 0 1 0 0 1 1 \textbackslash nK 0 1 0 1 0 0 1 0 0 0 \textbackslash nR 0 0 0 0 1 0 0 0 0 0 \textbackslash nA 0 0 0 1 0 1 0 0 1 1 \textbackslash nU 0 0 0 0 0 0 0 0 0 1 \textbackslash nM 0 0 1 0 0 0 0 0 0 0 \textbackslash nL 0 1 0 0 0 0 0 0 0 0 \textbackslash n\\
\hline
\footnotesize \textbf{GPT4's generation}: {\ttfamily

The adjacency matrix represents a directed graph where each city is a node and each road is a directed edge. The task is to find a path that starts at city 'U', visits cities 'A' and 'X' exactly twice, and visits all other cities at most once. 

We can solve this problem using depth-first search (DFS) with backtracking. DFS is a graph traversal algorithm that explores as far as possible along each branch before backtracking. 

Here are the steps to solve the problem:

1. Start at city 'U'. 

2. From city 'U', the only city we can go to is city 'L'. So, we go to city 'L'. 

3. From city 'L', we can go to city 'O'. So, we go to city 'O'. 

4. From city 'O', we can go to city 'A'. So, we go to city 'A'. This is the first visit to city 'A'. }

 \\
\hline
\end{tabular}
\caption*{\fontsize{9pt}{8pt}\selectfont GPT4's generation for this problem using 4\_shot CoT text prompting continues on the next page.}
\end{table}

\begin{table}[H]
\centering
\begin{tabular}{|p{14.5cm}|}
\hline
{\fontsize{10pt}{9pt}\selectfont \textbf{Continuation of GPT4's generation:} 
{\ttfamily

5. From city 'A', we can go to cities 'X', 'R', 'M', and 'L'. But we have already visited city 'L', and we want to visit city 'X' last, so we go to city 'R'. 

6. From city 'R', we can go to city 'K'. So, we go to city 'K'. 

7. From city 'K', we can go to cities 'O', 'X', and 'A'. But we have already visited city 'O', and we want to visit city 'X' last, so we go to city 'A'. This is the second visit to city 'A'. 

8. From city 'A', we can go to cities 'X', 'R', 'M', and 'L'. But we have already visited cities 'R', 'M', and 'L', so we go to city 'X'. This is the first visit to city 'X'. 

9. From city 'X', we can go to cities 'Y', 'Q', 'R', 'M', and 'A'. But we have already visited cities 'R', 'M', and 'A', so we go to city 'Y'. 

10. From city 'Y', we can go to cities 'O', 'R', 'A', 'U', 'M', and 'L'. But we have already visited cities 'O', 'R', 'A', 'U', 'M', and 'L', so we go to city 'Q'. 

11. From city 'Q', we can go to cities 'Y', 'X', 'R', 'M', and 'L'. But we have already visited cities 'Y', 'R', 'M', and 'L', so we go to city 'X'. This is the second visit to city 'X'. 

So, the quickest route that allows us to visit both cities 'A' and 'X' twice, ensuring that we stop at the two destinations twice on our path, is ['U', 'L', 'O', 'A', 'R', 'K', 'A', 'X', 'Y', 'Q', 'X'].}}

The answer is /answer\_list/['U', 'L', 'O', 'A', 'R', 'K', 'A', 'X', 'Y', 'Q', 'X'] \\
\hline
\textbf{LLM-solution extracted}: ['U', 'L', 'O', 'A', 'R', 'K', 'A', 'X', 'Y', 'Q', 'X']

\textbf{LLM-solution is not feasible}\\
\hline
\end{tabular}
\caption{\footnotesize GPT4's solution for a city\_directed\_graph problem using 4\_shot CoT text prompting method. As shown above, in the CoT part of in-context examples, we use ASCII characters to represent the intermediate states of the problem after each action. Also as city\_directed\_graph is a pathfinding problem, the 4\_shot CoT text prompt is constructed using one solved instance from each of the four other categories in SearchBench.}
\end{table}

\minisection{0\_shot code}

\begin{table}[H]
\centering
\begin{tabular}{|p{14.5cm}|}
\hline
{ \textbf{0\_shot code prompt}: Write a python script that solves the problem below. Explain the steps taken in comments:}\\
\hline
\footnotesize \textbf{problem statement (city\_directed\_graph instance 8)}: We have a map of cities, each represented by a letter, and they are connected by one-way roads. The adjacency matrix below shows the connections between the cities. Each row and column represents a city, and a '1' signifies a direct road from the city of the row to the city of the column. The travel time between any two directly connected cities is the same. Currently, we are located in city 'U'. Our task is to visit city A and city X excatly twice. Determine the quickest route that allows us to visit both these destination cities, ensuring that we stop at the two destinations twice on our path. The sequence in which we visit the destination cities is not important. However, apart from X and A, we can only visit each city once on our path. Provide the solution as a list of the city names on our path, including the start, in Python syntax.\textbackslash n\textbackslash n  Y O Q X K R A U M L \textbackslash nY 0 1 0 0 0 1 1 1 1 1 \textbackslash nO 0 0 0 0 0 0 1 0 0 0 \textbackslash nQ 1 0 0 1 0 1 0 0 1 0 \textbackslash nX 1 0 1 0 0 1 0 0 1 1 \textbackslash nK 0 1 0 1 0 0 1 0 0 0 \textbackslash nR 0 0 0 0 1 0 0 0 0 0 \textbackslash nA 0 0 0 1 0 1 0 0 1 1 \textbackslash nU 0 0 0 0 0 0 0 0 0 1 \textbackslash nM 0 0 1 0 0 0 0 0 0 0 \textbackslash nL 0 1 0 0 0 0 0 0 0 0 \textbackslash n\\
\hline
\footnotesize \textbf{GPT4's generation}: {\ttfamily
This problem can be solved using a depth-first search (DFS) algorithm. The DFS algorithm is an algorithm for traversing or searching tree or graph data structures. The algorithm starts at the root (selecting some arbitrary node as the root in the case of a graph) and explores as far as possible along each branch before backtracking.

Here is a Python script that solves the problem:}

```python
\begin{lstlisting}[language=Python]
# Define the adjacency matrix
adjacency_matrix = {
    'Y': ['O', 'R', 'A', 'U', 'M', 'L'],
    'O': ['A'],
    'Q': ['Y', 'X', 'R', 'M'],
    'X': ['Y', 'Q', 'R', 'M', 'L'],
    'K': ['O', 'X', 'A'],
    'R': ['K'],
    'A': ['X', 'R', 'M', 'L'],
    'U': ['L'],
    'M': ['Q'],
    'L': ['O']
}

# Define the start city and the destination cities
start_city = 'U'
destination_cities = ['A', 'X']

# Define a function to find all paths
def find_all_paths(start, end, path=[]):
    path = path + [start]
    if start == end:
        return [path]
    if start not in adjacency_matrix:
        return []
    paths = []
    for node in adjacency_matrix[start]:
        if node not in path:
            newpaths = find_all_paths(node, end, path)
            for newpath in newpaths:
                paths.append(newpath)
    return paths

# Find all paths from the start city to the destination cities
paths = []
for city in destination_cities:
    paths.extend(find_all_paths(start_city, city))

# Filter the paths to only include those that visit both destination cities twice
valid_paths = [path for path in paths if path.count('A') == 2 and path.count('X') == 2]

# Print the shortest valid path
print(min(valid_paths, key=len))
\end{lstlisting}
```
 \\
\hline
\end{tabular}
\caption*{\fontsize{9pt}{8pt}\selectfont GPT4's generation for this problem using 0\_shot code prompting continues on the next page.}
\end{table}

\begin{table}[H]
\centering
\begin{tabular}{|p{14.5cm}|}
\hline
{\fontsize{9pt}{9pt}\selectfont \textbf{Continuation of GPT4's generation:} 
{\ttfamily

This script first defines the adjacency matrix as a dictionary, where each key is a city and each value is a list of cities that can be reached directly from that city. It then defines a function to find all paths from a start city to an end city. This function uses a depth-first search algorithm to explore all possible paths. The script then finds all paths from the start city to each of the destination cities, and filters these paths to only include those that visit both destination cities twice. Finally, it prints the shortest valid path.}}
\\
\hline
\textbf{LLM-code run result}: LLM  program returned the following error: Traceback (most recent call last):

  File "./city\_directed\_graph\_results/gpt\_4\_0shot\_full\_code/gpt\_4\_0shot\_code\_python\_scripts/problem\_8.py", line 44, in <module>
  
    print(min(valid\_paths, key=len))
ValueError: min() arg is an empty sequence

\textbf{Solution is not feasible}\\
\hline
\end{tabular}
\caption{\footnotesize GPT4's solution for a city\_directed\_graph problem using 0\_shot text prompting method.}
\end{table}

\minisection{4\_shot A*}

\begin{table}[H]
\centering
\begin{tabular}{|p{14.5cm}|}
\hline
{\fontsize{9pt}{8pt}\selectfont \textbf{4\_shot A* prompt}: The goal is to solve the problems given by implementing the A* search algorithm in python.}
\\

Examples:

\\
In the 8-puzzle game, you are given a grid with numbered square tiles arranged randomly and one tile missing. The goal is to arrange the tiles in descending order by sliding them into the empty space. The tiles can move in 4 directions: left, right, up, and down.  Given the initial state of the puzzle below, where the empty spot is represented as “\_”, provide the shortest list of tiles that need to be swapped with the empty spot to achieve the goal state. The goal state is when all tiles are in descending order, with the largest number in the top left corner, and the empty spot is in the bottom right corner. The solution should be a list of numbers in Python format, where each number represents the number on the tile that the empty spot is swapped with at each turn. Initial state of the puzzle: [[55, 43, 17], [97, 35, 9], [12, 25, '\_']]

```python
\begin{lstlisting}[language=Python]

import heapq

def a_star():
   # Define the initial state and the goal state of the puzzle, represented as 3x3 tuples
   initial_state = ((55, 43, 17), (97, 35, 9), (12, 25, '_'))
   goal_state = ((97, 55, 43), (35, 25, 17), (12, 9, '_'))   
   # Encoding other variables given in the problem statement
   num_rows = 3
   num_cols = 3
   # Initialize a dictionary to store the cost of reaching each visited state
   visited_costs = {}
   visited_costs[initial_state] = 0
   # Initialize a priority queue of states not yet visited, with the initial state as the first element. The priority of each element is the cost to reach that state (g) + the estimate remaining cost (h) to reach the goal
   # Record the swaps required to get to each state in a list; no swaps performed to reach the initial state
   queue = [(0, 0, [], initial_state)]
   # While there are un-visited states
   while queue:
       # Pop the state with the lowest sum of the cost so far and estimated cost to the goal from the queue
       _, g, actions, state = heapq.heappop(queue)
       # We can check if the current state is the goal state with a simple equality check, as the goal state is predefined
       if state == goal_state:
           return actions
       # Generate all valid actions from the current state, which includes swapping any of the tiles neighboring the empty spot, with the empty spot
       # Generate the coordinates of the tiles neighboring "_"
       empty_row, empty_col = [(i, j) for i in range(num_rows) for j in range(num_cols) if state[i][j] == '_'][0]
       for d_row, d_col in [(0, -1), (0, 1), (1, 0), (-1, 0)]:
           swap_row, swap_col = empty_row + d_row, empty_col + d_col
           # Check if the swap is valid, ie if the coordinate of the tile to be swapped is a valid coordinate within the bounds of the board
           if 0 <= swap_row < num_rows and 0 <= swap_col < num_cols:
               # The actions is valid, generate the new state
               new_state = [list(row[:]) for row in state]
               number_to_be_swapped = new_state[swap_row][swap_col]
               # Do the swap
               new_state[empty_row][empty_col], new_state[swap_row][swap_col] = new_state[swap_row][swap_col], new_state[empty_row][empty_col]
               new_state = tuple(tuple(row) for row in new_state)
               # The cost so far is the number of swaps made, as our objective is to minimize the number of swaps required to reach the goal state 
               new_cost = g + 1
               # If the new state is unvisited or we found a new path with a lower cost to reach this state, add it to the queue of not-yet-visited states
               if new_state not in visited_costs or new_cost < visited_costs[new_state]:
                   visited_costs[new_state] = new_cost
                   heapq.heappush(queue, (new_cost + heuristic(new_state, goal_state), new_cost, actions + [number_to_be_swapped], new_state))       
   return None

\end{lstlisting} \\
\hline
\end{tabular}
\caption*{\fontsize{8pt}{8.5pt}\selectfont The A* prompt continues on the next page.}
\end{table}

\begin{table}[H]
\centering
\begin{tabular}{|p{14cm}|}
\hline
{\fontsize{8pt}{8.5pt}\selectfont \textbf{Continuation of A* prompt: }\begin{lstlisting}[language=Python]

def heuristic(state, goal):
   # An admissible and consistent heuristic is the sum of the Manhattan distances (the shortest path) of each tile from its goal position
   # The heuristic relaxes the constraint that a tile can only be swapped with the empty spot and presumes we can move the tiles to their goal position by swapping them with any of the other tiles
   # Thus the heuristic reports a lower estimate on the cost to reach goal state and is admissible
   # The heuristic is consistent because the cost of moving a tile to a neighboring coordinate is always 1, which is exactly the decrease in the Manhattan distance, if the tile is moved toward its goal position, otherwise the estimated cost of the successor node is the same or higher, and he heuristic estimate for the goal state is 0, as the distance of each tile from its goal position would be 0 in the goal state.
   h = 0
   for i in range(len(state)):
       for j in range(len(state[i])):
           # Can't compare integers with "_" when finding the goal position of each tile, thus ignore the "_" tile
           if state[i][j] != '_':
               # Get goal position of each tile
               goal_row, goal_col = [(x, y) for x in range(len(goal)) for y in range(len(goal[x])) if goal[x][y] == state[i][j]][0]
               # Add the the Manhattan distance of the current and goal coordinates of the tile to the estimate
               h += abs(i - goal_row) + abs(j - goal_col)
   return h

print(a_star())
```
\end{lstlisting}

---------

The game of 'Sort It' begins with 3 tubes, each filled with 4 balls of different colors. The goal is to sort the balls by color, with each tube containing balls of only one color. Only one ball can be moved at a time, taken from the top of one tube and placed on top of another. The capacity of each tube (maximum number of balls we can fit in each tube) is 6 balls. It is not allowed to place a ball in a tube that already has 6 balls. The solution should be a list of tuples, each containing, first, the index of the tube from which a ball is taken and, second, the index of the tube to which it is moved, indexing from 0. Given the initial state of the tubes, represented by the lists below (with the leftmost item being the color of the topmost ball in each tube), what is the shortest list of move tuples that will result in all the balls being correctly sorted? [['Green', 'Red', 'Green', 'Red'], ['Blue', 'Blue', 'Red', 'Green'], ['Red', 'Blue', 'Green', 'Blue']]}

\\
```python

\begin{lstlisting}[language=Python]
import heapq
from collections import Counter

def a_star():
   # Define the initial state of the tubes, as a 2d tuple of color of the balls in tubes 0 to 2
   initial_state = (('Green', 'Red', 'Green', 'Red'), ('Blue', 'Blue', 'Red', 'Green'), ('Red', 'Blue', 'Green', 'Blue'))
   # Encoding other variables given in the problem statement
   num_tubes = 3
   capacity = 6
   # Initialize a dictionary to store the cost of reaching each visited state
   visited_costs = {}
   visited_costs[initial_state] = 0
   # Initialize a priority queue of states not yet visited, with the initial state as the first element. The priority of each element is the cost to reach that state (g) + the estimate remaining cost (h) to reach the goal
   # Record the actions required to get to each state in a list; no actions performed to reach the initial state
   queue = [(0, 0, [], initial_state)]
   # While there are un-visited states
   while queue:
       # Pop the state with the lowest sum of the cost so far and estimated cost to the goal from the queue
       _, g, actions, state = heapq.heappop(queue)

\end{lstlisting} \\
\hline
\end{tabular}
\caption*{\fontsize{8pt}{8.5pt}\selectfont The A* prompt continues on the next page.}
\end{table}

\begin{table}[H]
\centering
\begin{tabular}{|p{14cm}|}
\hline
{\fontsize{8pt}{8.5pt}\selectfont \textbf{Continuation of A* prompt: }\begin{lstlisting}[language=Python]

       # Check of the current state is the goal state
       # The goal state is where each tube only contains balls of 1 single color
       if all(len(set(tube)) <= 1 for tube in state):
           return actions
       # Generate all possible actions from the current state, which includes moving a ball from any of the 3 tubes to another tube
       for from_tube_ind in range(num_tubes):
           for to_tube_ind in range(num_tubes):
               # Check if the new state would be valid, ie from_tube and to_tube must not be the same tube
               # And from_tube must at least have 1 ball to move and the to_tube cannot be at capacity
               if from_tube_ind != to_tube_ind and state[from_tube_ind] and len(state[to_tube_ind]) < capacity:
                   # Generate the new state
                   new_state = [list(tube[:]) for tube in state]
                   # The ball to move is the topmost ball in the from_tube, at index 0
                   ball_to_move = new_state[from_tube_ind].pop(0)
                   # Add the ball to the top of the to_tube
                   new_state[to_tube_ind].insert(0, ball_to_move)
                   new_state = tuple(tuple(tube) for tube in new_state)
                   # The cost so far is the number of moves made, as the task is to minimize the number of moves required
                   new_cost = g + 1
                   # If the new state is unvisited or we found a new path with a lower cost to arrive at this state, add it to the queue of un-visited states
                   if new_state not in visited_costs or new_cost < visited_costs[new_state]:
                       visited_costs[new_state] = new_cost
                       heapq.heappush(queue, (new_cost + heuristic(new_state), g + 1, actions + [(from_tube_ind, to_tube_ind)], new_state))
   return None

def heuristic(tubes):
   # An admissible and consistent heuristic for this problem is the count of balls that are not the same color as the most frequent color in their tube
   # This heuristic relaxes the constraint that only the ball at the top of the tube can be moved
   # It is admissible because it never overestimates the cost to reach the goal, as each mismatched ball must be moved at least once
   # It's consistent because moving a ball from one tube to another reduces the heuristic cost of the successor node by a max of 1 (if the moved ball's color matches the most common color in the new tube but not in the old one), which is equal to the cost of reaching the successor node
   # Thus h(s) is always less than or equal to c(s, n)(equal to 1) + h(n)
   h = 0
   for tube in tubes:
       if tube:
           # If there are ties in the frequency of colors, the most_commonm_color must be match the color of the balls lower that are in the tube, as moving lower balls is costlier          
           reversed_tube = tube[:]
           reversed_tube = reversed_tube[::-1]
           # Get the most common color
           most_common_color = Counter(reversed_tube).most_common(1)[0][0]
           for ball in tube:
               if ball != most_common_color:
                   h += 1
   return h

print(a_star())

\end{lstlisting}
```

---------

Given 6 labeled water jugs with capacities 37, 133, 38, 72, 41, 23, 122 liters, we aim to fill 3 unlabeled buckets, numbered 1 to 3 and arranged in a line in ascending order, with 195, 224, 268 liters of water respectively. The amount of water in each unlabeled bucket can not at any point in time exceed the amount of water in the bucket placed before it. Jugs can only be filled to the top and emptied completely, and the unlabeled buckets cannot be overfilled. An action, represented as a tuple ('+', X, Y) or ('-', X, Y), involves adding to or removing water from the unlabeled bucket numbered Y, using the jug with capacity X. Determine the shortest sequence of actions needed to fill the buckets as specified, and present the solution as a list of action tuples in Python syntax.} \\
\hline
\end{tabular}
\caption*{\fontsize{8pt}{8.5pt}\selectfont The A* prompt continues on the next page.}
\end{table}

\begin{table}[H]
\centering
\begin{tabular}{|p{14cm}|}
\hline
{\fontsize{8pt}{8.5pt}\selectfont \textbf{Continuation of A* prompt: }
```python

\begin{lstlisting}[language=Python]
from heapq import heappush, heappop

def a_star():
   # Define the capacities of the jugs, the goal state, and initial state, with states having an immutable data type
   jugs = [37, 133, 38, 72, 41, 23, 122]
   goal_state = (195, 224, 268)
   initial_state = (0, 0, 0)
   num_buckets = 3
   visited_costs = {}
   visited_costs[initial_state] = 0
   queue = []
   # No actions taken in the initial state
   heappush(queue, (0, 0, [], initial_state))
   while queue:
       _, g, actions, state = heappop(queue)
       # If the amount of water in the buckets in the current state equal the goal amounts, return the actions taken
       if state == goal_state:
           return actions
       # Generate all possible actions from the current state, which includes adding or subtracting water using any of the 6 jugs to any of the 3 buckets
       # Iterating through capacities of jugs and indices of buckets as the action tuples must include the operation ('+' or '-'), capacity of the jug used, and the index of the bucket affected
       for jug in jugs:
           for bucket_ind in range(num_buckets):
               # Check if adding water using the current jug results in a valid state, ie the addition must not result in overflowing any of the buckets
               if (state[bucket_ind] + jug <= goal_state[bucket_ind]):
                   temp_state = list(state)[:]
                   temp_state[bucket_ind] += jug
                   # And the new state must maintain the constraint on the relative amount of water in the buckets based on their order
                   if all(temp_state[i] <= temp_state[i + 1] for i in range(len(temp_state) - 1)):
                       # Generate the new state
                       new_state = tuple(temp_state)
                       # The cost so far is the number of actions taken, as the task is to minimize the number of actions required  to fill the buckets with the designated amount of water
                       new_cost = g + 1
                       if new_state not in visited_costs or new_cost < visited_costs[new_state]:
                           visited_costs[new_state] = new_cost
                           h = heuristic(state, goal_state, jugs)
                           # In the problem statement the buckets are indexed starting from 1, thus must add 1 to the bucket_ind
                           heappush(queue, (new_cost + h, new_cost,  actions + [('+', jug, bucket_ind+1)], new_state))
               # Check if removing water from the bucket results in a valid state. The buckest cannot have a negative amount of water
               if state[bucket_ind] - jug >= 0:
                   temp_state = list(state)[:]
                   temp_state[bucket_ind] -= jug
                   # The constraint on the relative amount of water in the buckets based on their order must hold after this action
                   if all(temp_state[i] <= temp_state[i + 1] for i in range(len(temp_state) - 1)):
                       new_state = tuple(temp_state)
                       new_cost = g + 1
                       if new_state not in visited_costs or new_cost < visited_costs[new_state]:
                           visited_costs[new_state] = new_cost
                           h = heuristic(state, goal_state, jugs)
                           heappush(queue, (new_cost + h, new_cost, actions + [('-', jug, bucket_ind+1)], new_state))
   return None

def heuristic(buckets_state, buckets_goal, jugs):
   # The heuristic function can be a simulation of filling buckets greedily, using the next largest jug repeatedly as long as the amount of water in the bucket does not exceed the goal amount
\end{lstlisting} }\\
\hline
\end{tabular}
\caption*{\fontsize{8pt}{8.5pt}\selectfont The A* prompt continues on the next page.}
\end{table}

\begin{table}[H]
\centering
\begin{tabular}{|p{14cm}|}
\hline
{\fontsize{8pt}{8.5pt}\selectfont \textbf{Continuation of A* prompt: }\begin{lstlisting}[language=Python]
   # This heuristic is admissible because it is greedy, always opting for the action that fills the buckets the most, ensuring it never overestimates the cost to reach the goal
   # The heuristic is consistent as the estimated cost of the next node is higher if water is removed from a bucket, or equal or less by at most 1 (equal to the cost of reaching the successor node, ie one action) as the maximum amount of water that can be added to the bucket is by using the largest jug that won't cause an overflow, which is exactly the jug used to fill the bucket in the heuristic. Thus h(n) can never be greater than c(n, n')(equal to 1) + h(n')
   h = 0
   # Sort the jugs by decreasing capacity
   jugs = sorted(jugs, reverse=True)
   # Iterate through the buckets
   for i in range(len(buckets_state)):
       bucket_fill = buckets_state[i]
       goal = buckets_goal[i]
       # Fill the bucket using the next largest jug as long as the bucket does not overflows
       for jug in jugs:
           while bucket_fill + jug < goal:
               bucket_fill += jug
               # Increment the estimated cost to the goal by 1 actions
               h += 1
   return h
   
print(a_star())

\end{lstlisting}
```

--------------

In the magic square problem, a 4x4 grid is filled with unique integers ranging from 29 to 54. Some numbers are already given, while others are unknown and represented as 'x'. The sums of columns must be None, 148, 196, None for columns 0 to 3 respectively, and the sums of rows must be None, 187, 149, None for rows 0 to 3 respectively, where None means that we do not have any constraints on the sum of the numbers in the row or column at that index. Also, the sum of the numbers in the diagonal from the top left to the bottom right corner of the grid should equal 166. The goal is to find unique integers in the given range to replace with ‘x’s in the grid below such that the sum of the specified rows, columns, and diagonal equals the given amounts and the sum of all of the numbers in the grid is as low as possible. The solution should be provided as a list of tuples in Python syntax. Each tuple should contain three numbers: the row index, the column index (both starting from 0), and the value of the unknown number at that position.\textbackslash n\textbackslash nGrid:\textbackslash n [[47 x x 32]\textbackslash n [x x x 49]\textbackslash n [x 31 50 x]\textbackslash n [x x 52 30]]

```python
\begin{lstlisting}[language=Python]
import heapq
import math
import numpy as np

def a_star():
   # Define the initial state of the grid as a 2d tuple
   initial_state = (('47', 'x', 'x', '32'),
                   ('x', 'x', 'x', '49'),
                   ('x', '31', '50', 'x'),
                   ('x', 'x', '52', '30'))
   num_rows = 4
   num_cols = 4
   row_sums = [None, 187, 149, None]
   col_sums = [None, 148, 196, None]
   diagonal_sum = 166
   # Create the set of the valid numbers that could be in the grid
   numbers = set(range(29, 54))
   visited_costs = {}
   visited_costs[initial_state] = 0
   queue = [(0, 0, [], initial_state)]
   while queue:
       _, g, actions, state = heapq.heappop(queue)
       # The goal state must not have any unknown numbers, thus we need to get (the coordinates of) the unknown numbers to check whether the current state is the goal state
       x_coords = [(i, j) for i, row in enumerate(state) for j, element in enumerate(row) if element == 'x']
       if not x_coords:
\end{lstlisting}} \\
\hline
\end{tabular}
\caption*{\fontsize{8pt}{8.5pt}\selectfont The A* prompt continues on the next page.}
\end{table}

\begin{table}[H]
\centering
\begin{tabular}{|p{14cm}|}
\hline
{\fontsize{8pt}{8.5pt}\selectfont \textbf{Continuation of A* prompt: }\begin{lstlisting}[language=Python]
           # Convert the cells of the state to ints to calculate and compare the sum of the specific positions in the current state with the given goal sums
           state_array = np.array([[int(element) for element in row] for row in state])
           if (np.all([i == j for i, j in zip(np.sum(state_array, axis=0), col_sums) if j]) and
               np.all([i == j for i, j in zip(np.sum(state_array, axis=1), row_sums) if j]) and
               np.trace(state_array) == diagonal_sum):
               return actions
       # If the state has at least 1 remaining unknown number, generate all possible actions from the current state, which includes replacing the next x in the grid with any of unique integers in the range
       else:
           first_x_coord = x_coords[0]
           # The number must be unique and not be present in any other cells of the grid
           used_numbers  = set(int(cell) for row in state for cell in row if cell != 'x')
           for number in numbers:
               # Check if the new state, containing the new number, would be valid; ie the number must be unique and the sum of specified positions must not exceed the goal sums with the addition of the new number
               sum_x_row_new_state = sum(int(cell) for cell in state[first_x_coord[0]] if cell != 'x') + number
               sum_x_col_new_state = sum(int(state[k][first_x_coord[1]]) for k in range(num_rows) if state[k][first_x_coord[1]] != 'x') + number
               sum_diag_new_state = sum(int(state[k][k]) for k in range(num_rows) if state[k][k] != 'x') + number
               if (number not in used_numbers and
                   # If the x is in one of the rows with a given sum, then the sum of the new row, with addition of the number, must not exceed the target sum
                   (row_sums[first_x_coord[0]] is None or sum_x_row_new_state <= row_sums[first_x_coord[0]]) and
                   # Similarly, if the x position is in a column or the diagonal with a goal sum
                   (col_sums[first_x_coord[1]] is None or sum_x_col_new_state <= col_sums[first_x_coord[1]]) and
                   (first_x_coord[0] != first_x_coord[1] or sum_diag_new_state <= diagonal_sum)):    
                   # Generate the new state
                   new_state = [list(row[:]) for row in state]
                   new_state[first_x_coord[0]][first_x_coord[1]] = str(number)
                   new_state = tuple(tuple(row) for row in new_state)
                   # The additional cost of this state is the value of the number replaced with x as we are trying to minimize the sum of the numbers in the grid
                   new_cost = g + number
                   if new_state not in visited_costs or new_cost < visited_costs[new_state]:
                       visited_costs[new_state] = new_cost
                       # Relaxing the constraints on the columns sums and the diagonal sum to calculate the heuristic
                       h = heuristic(new_state, row_sums, numbers)
                       heapq.heappush(queue, (new_cost + h, new_cost, actions + [(first_x_coord[0], first_x_coord[1], number)], new_state))
   return None

def heuristic(state, row_sums, numbers):
   # Relax the columns and diagonal sum constraints to compute an admissible and consistent heuristic
   # This heuristic considers the sum of differences between the given and current row sums, for rows with a specified sum value that have at least one unknown number, filling other x with the smallest unique number
   # The heuristic assumes xs can be filled with the smallest unique number satisfying row sum constraints, even if column or diagonal constraints are violated, thus reporting a lower cost estimate to reach the goal state, making it admissible
   # The heuristic is consistent, ie non-decreasing along the path to the goal state, because the cost of replacing an x in the grid with a number is the number itself, which is always greater than or equal to the number used to fill that position in the heuristic
   # Also the cost of the goal state is 0, as the sum of the rows equals their goal sums and there are no unknown numbers to fill in the grid
\end{lstlisting} }\\
\hline
\end{tabular}
\caption*{\fontsize{8pt}{8.5pt}\selectfont The A* prompt continues on the next page.}
\end{table}

\begin{table}[H]
\centering
\begin{tabular}{|p{14cm}|}
\hline
{\fontsize{8pt}{8.5pt}\selectfont \textbf{Continuation of A* prompt: }\begin{lstlisting}[language=Python]
   # Get numbers not used in the state currently
   used_numbers = set(int(cell) for row in state for cell in row if cell != 'x')
   available_numbers = sorted(list(numbers - used_numbers))
   h = 0
   for i in range(len(state)):
       if row_sums[i] is not None:
           row_full = True
           sum_row = sum(int(cell) for cell in state[i] if cell != 'x')
           # Check to see if the row with a target some has any more xs
           for cell in state[i]:
               if cell == 'x':
                   row_full = False   
           if not row_full:  
               h += row_sums[i] - sum_row
           # Otherwise the sum of a row must equal the target sum or there's no path to reach the goal state from this state. Return math.inf
           else:
               if sum_row != row_sums[i]:
                   return math.inf
       # For other rows, greedily fill the x positions with the smallest unique number in the range
       else:
           for cell in state[i]:
               if cell == 'x' and available_numbers:
                   h += available_numbers.pop(0)
   return h

print(a_star())
\end{lstlisting}
```

---------

Solve the below problem in the same format by encoding the problem states as a graph and implementing an A* search algorithm. Explain the rational behind each line of code using comments. The program must print the answer in the instructed format as a python list.}\\
\hline
\footnotesize \textbf{problem statement (city\_directed\_graph instance 8)}: We have a map of cities, each represented by a letter, and they are connected by one-way roads. The adjacency matrix below shows the connections between the cities. Each row and column represents a city, and a '1' signifies a direct road from the city of the row to the city of the column. The travel time between any two directly connected cities is the same. Currently, we are located in city 'U'. Our task is to visit city A and city X excatly twice. Determine the quickest route that allows us to visit both these destination cities, ensuring that we stop at the two destinations twice on our path. The sequence in which we visit the destination cities is not important. However, apart from X and A, we can only visit each city once on our path. Provide the solution as a list of the city names on our path, including the start, in Python syntax.\textbackslash n\textbackslash n  Y O Q X K R A U M L \textbackslash nY 0 1 0 0 0 1 1 1 1 1 \textbackslash nO 0 0 0 0 0 0 1 0 0 0 \textbackslash nQ 1 0 0 1 0 1 0 0 1 0 \textbackslash nX 1 0 1 0 0 1 0 0 1 1 \textbackslash nK 0 1 0 1 0 0 1 0 0 0 \textbackslash nR 0 0 0 0 1 0 0 0 0 0 \textbackslash nA 0 0 0 1 0 1 0 0 1 1 \textbackslash nU 0 0 0 0 0 0 0 0 0 1 \textbackslash nM 0 0 1 0 0 0 0 0 0 0 \textbackslash nL 0 1 0 0 0 0 0 0 0 0 \textbackslash n\\
\hline
{\footnotesize \textbf{GPT4's generation}:

```python
\begin{lstlisting}[language=Python]
import heapq

def a_star():
    # Define the adjacency matrix of the graph
    adjacency_matrix = {
        'Y': ['O', 'R', 'A', 'U', 'M', 'L'],
        'O': ['A'],
        'Q': ['Y', 'X', 'R', 'M'],
        'X': ['Y', 'Q', 'R', 'M', 'L'],
        'K': ['O', 'X', 'A'],
        'R': ['K'],
        'A': ['X', 'R', 'M', 'L'],
        'U': ['L'],
        'M': ['Q'],
        'L': ['O']}
    # Define the initial state and the goal state
    initial_state = ('U', frozenset(), 0)
    goal_state = 2
    # Initialize a dictionary to store the cost of reaching each visited state
    visited_costs = {}
    visited_costs[initial_state] = 0
    # Initialize a priority queue of states not yet visited, with the initial state as the first element
    queue = [(0, 0, [], initial_state)]
\end{lstlisting} }\\
\hline
\end{tabular}
\caption*{\fontsize{8pt}{8.5pt}\selectfont GPT4's generation continues on the next page.}
\end{table}

\begin{table}[H]
\centering
\begin{tabular}{|p{14cm}|}
\hline
{\fontsize{8pt}{8.5pt}\selectfont \textbf{Continuation of GPT4's generation: }\begin{lstlisting}[language=Python]
    # While there are un-visited states
    while queue:
        # Pop the state with the lowest sum of the cost so far and estimated cost to the goal from the queue
        _, g, actions, state = heapq.heappop(queue)
        # Check if the current state is the goal state
        if state[2] == goal_state:
            return actions
        # Generate all possible actions from the current state, which includes moving to any of the neighboring cities
        for neighbor in adjacency_matrix[state[0]]:
            # Check if the new state would be valid, ie if the city has not been visited before or is one of the destination cities
            if neighbor not in state[1] or neighbor in ['A', 'X']:
                # Generate the new state
                new_state = (neighbor, state[1].union({neighbor}), state[2] + (1 if neighbor in ['A', 'X'] else 0))
                # The cost so far is the number of cities visited, as the task is to minimize the number of cities visited
                new_cost = g + 1
                # If the new state is unvisited or we found a new path with a lower cost to reach this state, add it to the queue of not-yet-visited states
                if new_state not in visited_costs or new_cost < visited_costs[new_state]:
                    visited_costs[new_state] = new_cost
                    heapq.heappush(queue, (new_cost + heuristic(new_state, goal_state), new_cost, actions + [neighbor], new_state))
    return None

def heuristic(state, goal):
    # An admissible and consistent heuristic is the number of destination cities not yet visited twice
    # The heuristic is admissible because it never overestimates the cost to reach the goal state, as each destination city must be visited twice
    # The heuristic is consistent because the cost of moving to a neighboring city is always 1, which is exactly the decrease in the heuristic cost, if the city is a destination city and has not been visited twice before
    return goal - state[2]

print(a_star())
\end{lstlisting}
```}
\\
\hline
\textbf{LLM-code run result}: ['L', 'O', 'A', 'X']

\textbf{Solution is feasible! Solution is not correct.}\\
\hline
\end{tabular}
\caption{\footnotesize GPT4's solution for a city\_directed\_graph problem using A* prompting method.}
\end{table}

\minisection{MSMT A* second stage}

\begin{table}[H]
\centering
\begin{tabular}{|p{14.5cm}|}
\hline
{\fontsize{9pt}{8pt}\selectfont \textbf{MSMT A* second stage prompt}: The goal is to implement the initialize function for the given A* algorithm. The initialize function must initialize the visited\_costs dictionary and the priority queue and encode all of the information about the initial state and other variables given in the problem (ie a given matrix, eligible actions, goal coordinate, initial state of the board, etc).}
\\

Examples:

\\
In the 8-puzzle game, you are given a grid with numbered square tiles arranged randomly and one tile missing. The goal is to arrange the tiles in descending order by sliding them into the empty space. The tiles can move in 4 directions: left, right, up, and down.  Given the initial state of the puzzle below, where the empty spot is represented as “\_”, provide the shortest list of tiles that need to be swapped with the empty spot to achieve the goal state. The goal state is when all tiles are in descending order, with the largest number in the top left corner, and the empty spot is in the bottom right corner. The solution should be a list of numbers in Python format, where each number represents the number on the tile that the empty spot is swapped with at each turn. Initial state of the puzzle: [[55, 43, 17], [97, 35, 9], [12, 25, '\_']]

```python
\begin{lstlisting}[language=Python]
import heapq  

def a_star():
   # The initialize function initializes and returns the visited_costs dictionary and the priority queue and encodes all of the variables given in the problem (ie the initial and goal board and dimensions of the puzzle board)
   initial_state, goal_state, num_rows, num_cols, visited_costs, queue = initialize()
   # While there are un-visited states
   while queue:
       # Pop the state with the lowest sum of the cost so far and estimated cost to the goal from the queue
       _, g, actions, state = heapq.heappop(queue)
       # We can check if the current state is the goal state with a simple equality check, as the goal state is predefined
       if state == goal_state:
           return actions
       # Generate all valid actions from the current state, which includes swapping any of the tiles neighboring the empty spot, with the empty spot
       # Generate the coordinates of the tiles neighboring "_"
       empty_row, empty_col = [(i, j) for i in range(num_rows) for j in range(num_cols) if state[i][j] == '_'][0]
       for d_row, d_col in [(0, -1), (0, 1), (1, 0), (-1, 0)]:
           swap_row, swap_col = empty_row + d_row, empty_col + d_col
           # Check if the swap is valid, ie if the coordinate of the tile to be swapped is a valid coordinate within the bounds of the board
           if 0 <= swap_row < num_rows and 0 <= swap_col < num_cols:
               # The actions is valid, generate the new state
               new_state = [list(row[:]) for row in state]
               number_to_be_swapped = new_state[swap_row][swap_col]
               # Do the swap
               new_state[empty_row][empty_col], new_state[swap_row][swap_col] = new_state[swap_row][swap_col], new_state[empty_row][empty_col]
               new_state = tuple(tuple(row) for row in new_state)
               # The cost so far is the number of swaps made, as our objective is to minimize the number of swaps required to reach the goal state 
               new_cost = g + 1            
               # If the new state is unvisited or we found a new path with a lower cost to reach this state, add it to the queue of not-yet-visited states
               if new_state not in visited_costs or new_cost < visited_costs[new_state]:
                   visited_costs[new_state] = new_cost
                   heapq.heappush(queue, (new_cost + heuristic(new_state, goal_state), new_cost, actions + [number_to_be_swapped], new_state))       
   return None
\end{lstlisting} \\
\hline
\end{tabular}
\caption*{\fontsize{8pt}{8.5pt}\selectfont The MSMT A* prompt continues on the next page.}
\end{table}

\begin{table}[H]
\centering
\begin{tabular}{|p{14cm}|}
\hline
{\fontsize{8pt}{8.5pt}\selectfont \textbf{Continuation of MSMT A* prompt: }\begin{lstlisting}[language=Python]
def heuristic(state, goal):
   # An admissible and consistent heuristic is the sum of the Manhattan distances (the shortest path) of each tile from its goal position
   # The heuristic relaxes the constraint that a tile can only be swapped with the empty spot and presumes we can move the tiles to their goal position by swapping them with any of the other tiles
   # Thus the heuristic reports a lower estimate on the cost to reach goal state and is admissible
   # The heuristic is consistent because the cost of moving a tile to a neighboring coordinate is always 1, which is exactly the decrease in the Manhattan distance, if the tile is moved toward its goal position, otherwise the estimated cost of the successor node is the same or higher, and he heuristic estimate for the goal state is 0, as the distance of each tile from its goal position would be 0 in the goal state.
   h = 0
   for i in range(len(state)):
       for j in range(len(state[i])):
           # Can't compare integers with "_" when finding the goal position of each tile, thus ignore the "_" tile
           if state[i][j] != '_':
               # Get goal position of each tile
               goal_row, goal_col = [(x, y) for x in range(len(goal)) for y in range(len(goal[x])) if goal[x][y] == state[i][j]][0]
               # Add the the Manhattan distance of the current and goal coordinates of the tile to the estimate
               h += abs(i - goal_row) + abs(j - goal_col)
   return h

print(a_star())
\end{lstlisting}} \\

The target initialize function:

```python
\begin{lstlisting}[language=Python]
def initialize():
   # Define the initial state and the goal state of the puzzle, represented as 3x3 tuples
   initial_state = ((55, 43, 17), (97, 35, 9), (12, 25, '_'))
   goal_state = ((97, 55, 43), (35, 25, 17), (12, 9, '_'))   
   # Encoding other variables given in the problem statement
   num_rows = 3
   num_cols = 3
   # Initialize a dictionary to store the cost of reaching each visited state
   visited_costs = {}
   visited_costs[initial_state] = 0
   # Initialize a priority queue of states not yet visited, with the initial state as the first element. The priority of each element is the cost to reach that state (g) + the estimate remaining cost (h) to reach the goal
   # Record the swaps required to get to each state in a list; no swaps performed to reach the initial state
   queue = [(0, 0, [], initial_state)]
   return initial_state, goal_state, num_rows, num_cols, visited_costs, queue
\end{lstlisting}
```

------------

\\

The game of 'Sort It' begins with 3 tubes, each filled with 4 balls of different colors. The goal is to sort the balls by color, with each tube containing balls of only one color. Only one ball can be moved at a time, taken from the top of one tube and placed on top of another. The capacity of each tube (maximum number of balls we can fit in each tube) is 6 balls. It is not allowed to place a ball in a tube that already has 6 balls. The solution should be a list of tuples, each containing, first, the index of the tube from which a ball is taken and, second, the index of the tube to which it is moved, indexing from 0. Given the initial state of the tubes, represented by the lists below (with the leftmost item being the color of the topmost ball in each tube), what is the shortest list of move tuples that will result in all the balls being correctly sorted? [['Green', 'Red', 'Green', 'Red'], ['Blue', 'Blue', 'Red', 'Green'], ['Red', 'Blue', 'Green', 'Blue']]
\\
\hline
\end{tabular}
\caption*{\fontsize{8pt}{8.5pt}\selectfont The A* prompt continues on the next page.}
\end{table}

\begin{table}[H]
\centering
\begin{tabular}{|p{14cm}|}
\hline
{\fontsize{8pt}{8.5pt}\selectfont \textbf{Continuation of A* prompt}: 

```python

\begin{lstlisting}[language=Python]
import heapq
from collections import Counter

def a_star():  
   # The initialize function initializes and returns the visited_costs dictionary and the priority queue and encodes all of the variables given in the problem (ie the initial state of the tubes, number of tubes, and capacity of each tube)
   initial_state, num_tubes, capacity, visited_costs, queue = initialize()
   # While there are un-visited states
   while queue:
       # Pop the state with the lowest sum of the cost so far and estimated cost to the goal from the queue
       _, g, actions, state = heapq.heappop(queue)
       # Check of the current state is the goal state
       # The goal state is where each tube only contains balls of 1 single color
       if all(len(set(tube)) <= 1 for tube in state):
           return actions
       # Generate all possible actions from the current state, which includes moving a ball from any of the 3 tubes to another tube
       for from_tube_ind in range(num_tubes):
           for to_tube_ind in range(num_tubes):
               # Check if the new state would be valid, ie from_tube and to_tube must not be the same tube
               # And from_tube must at least have 1 ball to move and the to_tube cannot be at capacity
               if from_tube_ind != to_tube_ind and state[from_tube_ind] and len(state[to_tube_ind]) < capacity:
                   # Generate the new state
                   new_state = [list(tube[:]) for tube in state]
                   # The ball to move is the topmost ball in the from_tube, at index 0
                   ball_to_move = new_state[from_tube_ind].pop(0)
                   # Add the ball to the top of the to_tube
                   new_state[to_tube_ind].insert(0, ball_to_move)
                   new_state = tuple(tuple(tube) for tube in new_state)
                   # The cost so far is the number of moves made, as the task is to minimize the number of moves required
                   new_cost = g + 1
                   # If the new state is unvisited or we found a new path with a lower cost to arrive at this state, add it to the queue of un-visited states
                   if new_state not in visited_costs or new_cost < visited_costs[new_state]:
                       visited_costs[new_state] = new_cost
                       heapq.heappush(queue, (new_cost + heuristic(new_state), g + 1, actions + [(from_tube_ind, to_tube_ind)], new_state))
   return None
   
def heuristic(tubes):
   # An admissible and consistent heuristic for this problem is the count of balls that are not the same color as the most frequent color in their tube
   # This heuristic relaxes the constraint that only the ball at the top of the tube can be moved
   # It is admissible because it never overestimates the cost to reach the goal, as each mismatched ball must be moved at least once
   # It's consistent because moving a ball from one tube to another reduces the heuristic cost of the successor node by a max of 1 (if the moved ball's color matches the most common color in the new tube but not in the old one), which is equal to the cost of reaching the successor node
   # Thus h(s) is always less than or equal to c(s, n)(equal to 1) + h(n)
   h = 0
   for tube in tubes:
       if tube:
           # If there are ties in the frequency of colors, the most_commonm_color must be match the color of the balls lower that are in the tube, as moving lower balls is costlier          
           reversed_tube = tube[:]
           reversed_tube = reversed_tube[::-1]
           # Get the most common color
           most_common_color = Counter(reversed_tube).most_common(1)[0][0]
           for ball in tube:
               if ball != most_common_color:
                   h += 1
   return h
print(a_star())
\end{lstlisting}
```}\\
\hline
\end{tabular}
\caption*{\fontsize{8pt}{8.5pt}\selectfont The MSMT A* prompt continues on the next page.}
\end{table}

\begin{table}[H]
\centering
\begin{tabular}{|p{14cm}|}
\hline
{\fontsize{8pt}{8.5pt}\selectfont \textbf{Continuation of MSMT A* prompt}: 

The target initialize function:

```python
\begin{lstlisting}[language=Python]
def initialize():
   # Define the initial state of the tubes, as a 2d tuple of color of the balls in tubes 0 to 2
   initial_state = (('Green', 'Red', 'Green', 'Red'), ('Blue', 'Blue', 'Red', 'Green'), ('Red', 'Blue', 'Green', 'Blue'))
   # Encoding other variables given in the problem statement
   num_tubes = 3
   capacity = 6
   # Initialize a dictionary to store the cost of reaching each visited state
   visited_costs = {}
   visited_costs[initial_state] = 0


   # Initialize a priority queue of states not yet visited, with the initial state as the first element. The priority of each element is the cost to reach that state (g) + the estimate remaining cost (h) to reach the goal
   # Record the actions required to get to each state in a list; no actions performed to reach the initial state
   queue = [(0, 0, [], initial_state)]
  
   return initial_state, num_tubes, capacity, visited_costs, queue
```
\end{lstlisting}

---------}

\\

Given 6 labeled water jugs with capacities 37, 133, 38, 72, 41, 23, 122 liters, we aim to fill 3 unlabeled buckets, numbered 1 to 3 and arranged in a line in ascending order, with 195, 224, 268 liters of water respectively. The amount of water in each unlabeled bucket can not at any point in time exceed the amount of water in the bucket placed before it. Jugs can only be filled to the top and emptied completely, and the unlabeled buckets cannot be overfilled. An action, represented as a tuple ('+', X, Y) or ('-', X, Y), involves adding to or removing water from the unlabeled bucket numbered Y, using the jug with capacity X. Determine the shortest sequence of actions needed to fill the buckets as specified, and present the solution as a list of action tuples in Python syntax. \\

```python

\begin{lstlisting}[language=Python]
from heapq import heappush, heappop
  
def a_star():
   jugs, goal_state, initial_state, num_buckets, visited_costs, queue = initialize()
   while queue:
       _, g, actions, state = heappop(queue)
       # If the amount of water in the buckets in the current state equal the goal amounts, return the actions taken
       if state == goal_state:
           return actions
       # Generate all possible actions from the current state, which includes adding or subtracting water using any of the 6 jugs to any of the 3 buckets
       # Iterating through capacities of jugs and index of buckets as the action tuples must include the operation ('+' or '-'), capacity of the jug used, and the index of the bucket affected
       for jug in jugs:
           for bucket_ind in range(num_buckets):
               # Check if adding water using the current jug results in a valid state, ie the addition must not result in overflowing any of the buckets
               if (state[bucket_ind] + jug <= goal_state[bucket_ind]):
                   temp_state = list(state)[:]
                   temp_state[bucket_ind] += jug
                   # And the new state must maintain the constraint on the relative amount of water in the buckets based on their order
                   if all(temp_state[i] <= temp_state[i + 1] for i in range(len(temp_state) - 1)):
                       # Generate the new state
                       new_state = tuple(temp_state)
                       # The cost so far is the number of actions taken, as the task is to minimize the number of actions required  to fill the buckets with the designated amount of water
                       new_cost = g + 1
\end{lstlisting} \\
\hline
\end{tabular}
\caption*{\fontsize{8pt}{8.5pt}\selectfont The MSMT A* prompt continues on the next page.}
\end{table}

\begin{table}[H]
\centering
\begin{tabular}{|p{14.5cm}|}
\hline
{\fontsize{8pt}{8.5pt}\selectfont \textbf{Continuation of MSMT A* prompt: }\begin{lstlisting}[language=Python]
                       if new_state not in visited_costs or new_cost < visited_costs[new_state]:
                           visited_costs[new_state] = new_cost
                           h = heuristic(state, goal_state, jugs)
                           # In the problem statement the buckets are indexed starting from 1, thus must add 1 to the bucket_ind
                           heappush(queue, (new_cost + h, new_cost,  actions + [('+', jug, bucket_ind+1)], new_state))
               # Check if removing water from the bucket results in a valid state. The buckest cannot have a negative amount of water
               if state[bucket_ind] - jug >= 0:
                   temp_state = list(state)[:]
                   temp_state[bucket_ind] -= jug
                   # The constraint on the relative amount of water in the buckets based on their order must hold after this action
                   if all(temp_state[i] <= temp_state[i + 1] for i in range(len(temp_state) - 1)):
                       new_state = tuple(temp_state)
                       new_cost = g + 1
                       if new_state not in visited_costs or new_cost < visited_costs[new_state]:
                           visited_costs[new_state] = new_cost
                           h = heuristic(state, goal_state, jugs)
                           heappush(queue, (new_cost + h, new_cost, actions + [('-', jug, bucket_ind+1)], new_state))
   return None

def heuristic(buckets_state, buckets_goal, jugs):
   # The heuristic function can be a simulation of filling buckets greedily, using the next largest jug repeatedly as long as the amount of water in the bucket does not exceed the goal amount
   # This heuristic is admissible because it is greedy, always opting for the action that fills the buckets the most, ensuring it never overestimates the cost to reach the goal
   # The heuristic is consistent as the estimated cost of the next node is higher if water is removed from a bucket, or equal or less by at most 1 (equal to the cost of reaching the successor node, ie one action) as the maximum amount of water that can be added to the bucket is by using the largest jug that won't cause an overflow, which is exactly the jug used to fill the bucket in the heuristic. Thus h(n) can never be greater than c(n, n')(equal to 1) + h(n')
   h = 0
   # Sort the jugs by decreasing capacity
   jugs = sorted(jugs, reverse=True)
   # Iterate through the buckets
   for i in range(len(buckets_state)):
       bucket_fill = buckets_state[i]
       goal = buckets_goal[i]
       # Fill the bucket using the next largest jug as long as the bucket does not overflows
       for jug in jugs:
           while bucket_fill + jug < goal:
               bucket_fill += jug
               # Increment the estimated cost to the goal by 1 actions
               h += 1
   return h

print(a_star())
\end{lstlisting}
```

The target initialize function:

```python

\begin{lstlisting}[language=Python]
def initialize():
   # Define the capacities of the jugs, the goal state, and initial state, with states having an immutable data type
   jugs = [37, 133, 38, 72, 41, 23, 122]
   goal_state = (195, 224, 268)
   initial_state = (0, 0, 0)
   num_buckets = 3
   visited_costs = {}
   visited_costs[initial_state] = 0
   queue = []
   # No actions taken in the initial state
   heappush(queue, (0, 0, [], initial_state))
   return jugs, goal_state, initial_state, num_buckets, visited_costs, queue
\end{lstlisting}
```}\\
\hline
\end{tabular}
\caption*{\fontsize{8pt}{8.5pt}\selectfont The MSMT A* prompt continues on the next page.}
\end{table}

\begin{table}[H]
\centering
\begin{tabular}{|p{14.5cm}|}
\hline
{\fontsize{8pt}{8.5pt}\selectfont \textbf{Continuation of MSMT A* prompt: }

--------------

In the magic square problem, a 4x4 grid is filled with unique integers ranging from 29 to 54. Some numbers are already given, while others are unknown and represented as 'x'. The sums of columns must be None, 148, 196, None for columns 0 to 3 respectively, and the sums of rows must be None, 187, 149, None for rows 0 to 3 respectively, where None means that we do not have any constraints on the sum of the numbers in the row or column at that index. Also, the sum of the numbers in the diagonal from the top left to the bottom right corner of the grid should equal 166. The goal is to find unique integers in the given range to replace with ‘x’s in the grid below such that the sum of the specified rows, columns, and diagonal equals the given amounts and the sum of all of the numbers in the grid is as low as possible. The solution should be provided as a list of tuples in Python syntax. Each tuple should contain three numbers: the row index, the column index (both starting from 0), and the value of the unknown number at that position.\textbackslash n\textbackslash nGrid:\textbackslash n [[47 x x 32]\textbackslash n [x x x 49]\textbackslash n [x 31 50 x]\textbackslash n [x x 52 30]]

```python
\begin{lstlisting}[language=Python]
import heapq
import math
import numpy as np
  
def a_star():
   initial_state, num_rows, num_cols, row_sums, col_sums, diagonal_sum, numbers, visited_costs, queue = initialize()
   while queue:
       _, g, actions, state = heapq.heappop(queue)
       # The goal state must not have any unknown numbers, thus we need to get (the coordinates of) the unknown numbers to check whether the current state is the goal state
       x_coords = [(i, j) for i, row in enumerate(state) for j, element in enumerate(row) if element == 'x']
       if not x_coords:
           # Convert the cells of the state to ints to calculate and compare the sum of the specific positions in the current state with the given goal sums
           state_array = np.array([[int(element) for element in row] for row in state])
           if (np.all([i == j for i, j in zip(np.sum(state_array, axis=0), col_sums) if j]) and
               np.all([i == j for i, j in zip(np.sum(state_array, axis=1), row_sums) if j]) and
               np.trace(state_array) == diagonal_sum):
               return actions
       # If the state has at least 1 remaining unknown number, generate all possible actions from the current state, which includes replacing the next x in the grid with any of unique integers in the range
       else:
           first_x_coord = x_coords[0]
           # The number must be unique and not be present in any other cells of the grid
           used_numbers  = set(int(cell) for row in state for cell in row if cell != 'x')
           for number in numbers:
               # Check if the new state, containing the new number, would be valid; ie the number must be unique and the sum of specified positions must not exceed the goal sums with the addition of the new number
               sum_x_row_new_state = sum(int(cell) for cell in state[first_x_coord[0]] if cell != 'x') + number
               sum_x_col_new_state = sum(int(state[k][first_x_coord[1]]) for k in range(num_rows) if state[k][first_x_coord[1]] != 'x') + number
               sum_diag_new_state = sum(int(state[k][k]) for k in range(num_rows) if state[k][k] != 'x') + number
               if (number not in used_numbers and
                   # If the x is in one of the rows with a given sum, then the sum of the new row, with addition of the number, must not exceed the target sum
                   (row_sums[first_x_coord[0]] is None or sum_x_row_new_state <= row_sums[first_x_coord[0]]) and
                   # Similarly, if the x position is in a column or the diagonal with a goal sum
                   (col_sums[first_x_coord[1]] is None or sum_x_col_new_state <= col_sums[first_x_coord[1]]) and
                   (first_x_coord[0] != first_x_coord[1] or sum_diag_new_state <= diagonal_sum)):  
                   # Generate the new state
                   new_state = [list(row[:]) for row in state]
                   new_state[first_x_coord[0]][first_x_coord[1]] = str(number)
                   new_state = tuple(tuple(row) for row in new_state)
                   # The additional cost of this state is the value of the number replaced with x as we are trying to minimize the sum of the numbers in the grid
                   new_cost = g + number     
\end{lstlisting} }\\
\hline
\end{tabular}
\caption*{\fontsize{8pt}{8.5pt}\selectfont The MSMT A* prompt continues on the next page.}
\end{table}

\begin{table}[H]
\centering
\begin{tabular}{|p{14cm}|}
\hline
{\fontsize{8pt}{8.5pt}\selectfont \textbf{Continuation of MSMT A* prompt: }\begin{lstlisting}[language=Python]
                   if new_state not in visited_costs or new_cost < visited_costs[new_state]:
                       visited_costs[new_state] = new_cost
                       # Relaxing the constraints on the columns sums and the diagonal sum to calculate the heuristic
                       h = heuristic(new_state, row_sums, numbers)
                       heapq.heappush(queue, (new_cost + h, new_cost, actions + [(first_x_coord[0], first_x_coord[1], number)], new_state))
   return None

def heuristic(state, row_sums, numbers):
   # Relax the columns and diagonal sum constraints to compute an admissible and consistent heuristic
   # This heuristic considers the sum of differences between the given and current row sums, for rows with a specified sum value that have at least one unknown number, filling other x with the smallest unique number
   # The heuristic assumes xs can be filled with the smallest unique number satisfying row sum constraints, even if column or diagonal constraints are violated, thus reporting a lower cost estimate to reach the goal state, making it admissible
   # The heuristic is consistent, ie non-decreasing along the path to the goal state, because the cost of replacing an x in the grid with a number is the number itself, which is always greater than or equal to the number used to fill that position in the heuristic
   # Also the cost of the goal state is 0, as the sum of the rows equals their goal sums and there are no unknown numbers to fill in the grid
   # Get numbers not used in the state currently
   used_numbers = set(int(cell) for row in state for cell in row if cell != 'x')
   available_numbers = sorted(list(numbers - used_numbers))
   h = 0
   for i in range(len(state)):
       if row_sums[i] is not None:
           row_full = True
           sum_row = sum(int(cell) for cell in state[i] if cell != 'x')
           # Check to see if the row with a target some has any more xs
           for cell in state[i]:
               if cell == 'x':
                   row_full = False    
           if not row_full:  
               h += row_sums[i] - sum_row
           # Otherwise the sum of a row must equal the target sum or there's no path to reach the goal state from this state. Return math.inf
           else:
               if sum_row != row_sums[i]:
                   return math.inf
       # For other rows, greedily fill the x positions with the smallest unique number in the range
       else:
           for cell in state[i]:
               if cell == 'x' and available_numbers:
                   h += available_numbers.pop(0)
   return h

print(a_star())
\end{lstlisting} 
```

The target initialize function:

```python
\begin{lstlisting}[language=Python]
def initialize():
   # Define the initial state of the grid as a 2d tuple
   initial_state = (('47', 'x', 'x', '32'),
                   ('x', 'x', 'x', '49'),
                   ('x', '31', '50', 'x'),
                   ('x', 'x', '52', '30'))
   num_rows = 4
   num_cols = 4
   row_sums = [None, 187, 149, None]
   col_sums = [None, 148, 196, None]
   diagonal_sum = 166
   # Create the set of the valid numbers that could be in the grid
   numbers = set(range(29, 54))
   visited_costs = {}
   visited_costs[initial_state] = 0
   queue = [(0, 0, [], initial_state)] 
   return initial_state, num_rows, num_cols, row_sums, col_sums, diagonal_sum, numbers, visited_costs, queue
\end{lstlisting} 
```}\\
\hline
\end{tabular}
\caption*{\fontsize{8pt}{8.5pt}\selectfont The MSMT A*  continues on the next page.}
\end{table}

\begin{table}[H]
\centering
\begin{tabular}{|p{14.5cm}|}
\hline
{\fontsize{8pt}{8.5pt}\selectfont \textbf{Continuation of MSMT A* prompt: }}\\
\hline
\footnotesize \textbf{problem statement (city\_directed\_graph instance 8)}: We have a map of cities, each represented by a letter, and they are connected by one-way roads. The adjacency matrix below shows the connections between the cities. Each row and column represents a city, and a '1' signifies a direct road from the city of the row to the city of the column. The travel time between any two directly connected cities is the same. Currently, we are located in city 'U'. Our task is to visit city A and city X excatly twice. Determine the quickest route that allows us to visit both these destination cities, ensuring that we stop at the two destinations twice on our path. The sequence in which we visit the destination cities is not important. However, apart from X and A, we can only visit each city once on our path. Provide the solution as a list of the city names on our path, including the start, in Python syntax.\textbackslash n\textbackslash n  Y O Q X K R A U M L \textbackslash nY 0 1 0 0 0 1 1 1 1 1 \textbackslash nO 0 0 0 0 0 0 1 0 0 0 \textbackslash nQ 1 0 0 1 0 1 0 0 1 0 \textbackslash nX 1 0 1 0 0 1 0 0 1 1 \textbackslash nK 0 1 0 1 0 0 1 0 0 0 \textbackslash nR 0 0 0 0 1 0 0 0 0 0 \textbackslash nA 0 0 0 1 0 1 0 0 1 1 \textbackslash nU 0 0 0 0 0 0 0 0 0 1 \textbackslash nM 0 0 1 0 0 0 0 0 0 0 \textbackslash nL 0 1 0 0 0 0 0 0 0 0 \textbackslash n\\
\hline

{\fontsize{8pt}{8.5pt}\selectfont \textbf{The A* algorithm generated by GPT4 for the city\_directed\_graph problem type during the first stage of MSMT that passed the unit tests:}
```python

\begin{lstlisting}[language=Python]
import heapq

def a_star():
   # The initialize function initializes and returns the visited_costs dictionary and the priority queue and encodes all of the variables given in the problem (ie the initial and goal board and dimensions of the puzzle board)
   initial_state, goal_state, adjacency_matrix, cities, visited_costs, queue = initialize()
   # While there are un-visited states
   while queue:
       # Pop the state with the lowest sum of the cost so far and estimated cost to the goal from the queue
       _, g, actions, state = heapq.heappop(queue)
       current_city, remaining_cities = state
       # We can check if the current state is the goal state with a simple equality check, as the goal state is predefined
       if not remaining_cities:
           return actions
       # Generate all valid actions from the current state, which includes moving to any of the cities connected to the current city
       for i in range(len(cities)):
           if adjacency_matrix[cities.index(current_city)][i] == 1:
               next_city = cities[i]
               # Check if the next city is a destination city and we have not visited it twice already
               if next_city in remaining_cities:
                   new_remaining_cities = list(remaining_cities)
                   new_remaining_cities.remove(next_city)
                   new_state = (next_city, tuple(new_remaining_cities))
                   new_cost = g + 1
                   if new_state not in visited_costs or new_cost < visited_costs[new_state]:
                       visited_costs[new_state] = new_cost
                       heapq.heappush(queue, (new_cost + heuristic(new_state, goal_state), new_cost, actions + [next_city], new_state))
               # If the next city is not a destination city and we have not visited it before, we can move to it
               elif next_city not in actions:
                   new_state = (next_city, remaining_cities)
                   new_cost = g + 1
                   if new_state not in visited_costs or new_cost < visited_costs[new_state]:
                       visited_costs[new_state] = new_cost
                       heapq.heappush(queue, (new_cost + heuristic(new_state, goal_state), new_cost, actions + [next_city], new_state))
   return None

def heuristic(state, goal_state):
   # An admissible and consistent heuristic is the number of remaining cities to visit
   # The heuristic relaxes the constraint that we can only move to cities connected to the current city and presumes we can move to any of the remaining cities directly from the current city
   # Thus the heuristic reports a lower estimate on the cost to reach goal state and is admissible
   # The heuristic is consistent because the cost of moving to a city is always 1, which is exactly the decrease in the number of remaining cities to visit, if the city is a destination city, otherwise the estimated cost of the successor node is the same or higher, and he heuristic estimate for the goal state is 0, as there are no remaining cities to visit in the goal state.
   return len(state[1])
   
print(a_star())
\end{lstlisting}
```}
\\
\hline
\end{tabular}
\caption*{\fontsize{8pt}{8.5pt}\selectfont The MSMT A*  continues on the next page.}
\end{table}

\begin{table}[H]
\centering
\begin{tabular}{|p{14.5cm}|}
\hline
{\fontsize{8pt}{8.5pt}\selectfont \textbf{Continuation of MSMT A* prompt: }

 An example implementation of the target initialize function. Crucial: the value of variables must be updated according to the problem statement, the implementation above is with hypothetical values: 
 
```python

\begin{lstlisting}[language=Python]
def initialize():
   # Define the initial state and the goal state of the puzzle, represented as 3x3 tuples
   initial_state = 'H'
   goal_state = ['K', 'K', 'Z', 'Z']   
   # Encoding other variables given in the problem statement
   adjacency_matrix = [[0, 0, 1, 1, 0, 0, 1, 1, 0, 1],
                       [0, 0, 0, 0, 0, 0, 0, 1, 0, 1],
                       [0, 1, 0, 1, 0, 1, 1, 0, 0, 1],
                       [0, 1, 1, 0, 1, 0, 0, 0, 0, 0],
                       [1, 1, 0, 0, 0, 0, 0, 0, 0, 1],
                       [0, 0, 0, 1, 0, 0, 1, 0, 1, 0],
                       [0, 0, 1, 1, 1, 0, 0, 0, 0, 0],
                       [1, 0, 0, 0, 0, 0, 1, 0, 0, 1],
                       [0, 0, 0, 0, 0, 0, 1, 0, 0, 0],
                       [0, 0, 0, 0, 1, 0, 0, 0, 0, 0]]
   cities = ['K', 'B', 'Z', 'S', 'X', 'H', 'O', 'E', 'U', 'V']
   # Initialize a dictionary to store the cost of reaching each visited state
   visited_costs = {}
   visited_costs[(initial_state, tuple(goal_state))] = 0
   # Initialize a priority queue of states not yet visited, with the initial state as the first element. The priority of each element is the cost to reach that state (g) + the estimate remaining cost (h) to reach the goal
   # Record the actions required to get to each state in a list; no actions performed to reach the initial state
   queue = [(0, 0, [], (initial_state, tuple(goal_state)))]
   return initial_state, goal_state, adjacency_matrix, cities, visited_costs, queue
\end{lstlisting}
```

Initialize the visited\_costs dictionary and the priority queue and encode all the variables given of only the last problem in the initialize function as the examples provided above. The value of variables must match the information encoded in the problem statement.

The target initialize function:}\\
\hline
{\footnotesize \textbf{GPT4's generation}:

```python
\begin{lstlisting}[language=Python]
def initialize():
   # Define the initial state and the goal state of the puzzle, represented as 3x3 tuples
   initial_state = ('U', ('A', 'A', 'X', 'X')) 
   goal_state = []   
   # Encoding other variables given in the problem statement
   adjacency_matrix = [[0, 1, 0, 0, 0, 1, 1, 1, 1, 1],
                       [0, 0, 0, 0, 0, 0, 1, 0, 0, 0],
                       [1, 0, 0, 1, 0, 1, 0, 0, 1, 0],
                       [1, 0, 1, 0, 0, 1, 0, 0, 1, 1],
                       [0, 1, 0, 1, 0, 0, 1, 0, 0, 0],
                       [0, 0, 0, 0, 1, 0, 0, 0, 0, 0],
                       [0, 0, 0, 1, 0, 1, 0, 0, 1, 1],
                       [0, 0, 0, 0, 0, 0, 0, 0, 0, 1],
                       [0, 0, 1, 0, 0, 0, 0, 0, 0, 0],
                       [0, 1, 0, 0, 0, 0, 0, 0, 0, 0]]
   cities = ['Y', 'O', 'Q', 'X', 'K', 'R', 'A', 'U', 'M', 'L']
   # Initialize a dictionary to store the cost of reaching each visited state
   visited_costs = {}
   visited_costs[initial_state] = 0
   # Initialize a priority queue of states not yet visited, with the initial state as the first element. The priority of each element is the cost to reach that state (g) + the estimate remaining cost (h) to reach the goal
   # Record the actions required to get to each state in a list; no actions performed to reach the initial state
   queue = [(0, 0, [initial_state[0]], initial_state)]
   return initial_state, goal_state, adjacency_matrix, cities, visited_costs, queue
\end{lstlisting} }\\

\hline
\textbf{LLM-code run result}: ['U', 'L', 'O', 'A', 'X', 'Y', 'A', 'X']

\textbf{Solution is feasible! Solution is correct! Solution is optimal! Computation time of the LLM code is 97\% of the our A* code computation time.}\\
\hline
\end{tabular}
\caption{\footnotesize GPT4's solution for a city\_directed\_graph problem using MSMT A* prompting method.}
\end{table}

\clearpage
\section{Hosting, Licensing, and Maintenance}

We accept responsibility for any violations of rights that might have occurred in the curation of this dataset. We affirm that the dataset is composed solely of search problems and does not include any sensitive information. The data and code associated with SearchBench are licensed under the Creative Commons (CC BY-SA) license, ensuring open access and usability for the research community.

To ensure the long-term availability and preservation of the SearchBench dataset, we have hosted it on both Hugging Face and GitHub. Moreover, we will provide full access to  the code for prompting and inference methods, as well as automated pipelines for generating and evaluating an arbitrary number of instances though these platforms, after the double blind review period. We are committed to maintaining the dataset on these platforms with continued open access. Additionally, we anticipate releasing future versions of this dataset with increased scalability.

\end{document}